\documentclass{article}

\PassOptionsToPackage{numbers}{natbib}

\usepackage[final]{neurips_2021}

\usepackage{enumitem}
\usepackage[table]{xcolor}
\definecolor{Gray}{gray}{0.9}
\definecolor{midgreen}{rgb}{0.1,0.5,0.1}
\definecolor{darkgray}{gray}{0.25}
\definecolor{lightblue}{rgb}{0.25,0.25,0.8}
\definecolor{mydarkblue}{rgb}{0,0.08,0.45}
\usepackage[colorlinks,
            linkcolor=mydarkblue,
            anchorcolor=blue,
            urlcolor=mydarkblue,
            citecolor=mydarkblue]{hyperref}
\usepackage{amsmath}
\usepackage{amsthm}
\usepackage{amssymb}
\usepackage{mathtools}
\usepackage{bm}

\usepackage[utf8]{inputenc} 
\usepackage[T1]{fontenc}    
\usepackage{hyperref}       
\usepackage{url}            
\usepackage{booktabs}       
\usepackage{amsfonts}       
\usepackage{nicefrac}       
\usepackage{microtype}      

\usepackage{thmtools}
\usepackage{thm-restate}

\usepackage{algorithm}
\usepackage[noend]{algorithmic}

\usepackage{subcaption}

\usepackage[capitalize,nameinlink]{cleverref} 
\newcommand{\email}[1]{\href{mailto:#1}{\color{black} \texttt{#1}}}

\newtheorem{theorem}{Theorem}
\newtheorem{lemma}{Lemma}
\newtheorem{corr}{Corollary}

\theoremstyle{definition}
\newtheorem{rmk}{Remark}
\newtheorem{defn}{Definition}

\newcommand{\NTK}{\mathrm{NTK} }
\newcommand{\NNGP}{\mathrm{NNGP}}
\newcommand{\CS}{{\sc CountSketch} }
\newcommand{\TS}{{\sc TensorSketch} }
\newcommand{\NTKS}{{\sc NTKSketch} }
\newcommand{\NTKRF}{{\sc NTKRF} }

\newcommand{\GradRF}{{\sc GradRF} }

\newcommand{\CNTKS}{{\sc CNTKSketch} }
\newcommand{\EE}{\mathbb{E}}

\newcommand{\TSRHT}{{\sc TensorSRHT} }

\newcommand{\SRHT}{{\sc SRHT} }

\newcommand{\PolyS}{{\sc PolySketch} }

\newcommand{\CIFARTEN}{$\mathtt{CIFAR}$-10}

\newcommand{\VOC}{$\mathtt{VOC07}$}
\newcommand{\CUB}{$\mathtt{CUB}$-200}
\newcommand{\CALTECH}{$\mathtt{Caltech}$-101}

\newcommand{\FLOWER}{$\mathtt{Flower}$-102}
\newcommand{\FOOD}{$\mathtt{Food}$-101}
\newcommand{\DOG}{$\mathtt{Dog}$-120}
\newcommand{\MSD}{$\mathrm{MillionSongs}$}

\newcommand{\Protein}{$\mathrm{Protein}$}

\newcommand{\WorkLoads}{$\mathrm{WorkLoads}$}
\newcommand{\CT}{$\mathrm{CT}$}

\newcommand{\rank}{\mathrm{rank}}

\newcommand{\norm}[1]{\ensuremath{\left\| #1 \right\|}}

\newcommand{\inner}[1]{\left \langle {#1} \right \rangle}
\newcommand{\bigo}{\mathcal{O}}
\newcommand{\abs}[1]{\left |#1\right|}
\newcommand{\FFT}{\mathtt{FFT}}
\newcommand{\iFFT}{\mathtt{FFT}^{-1}}

\def\tr{\mathtt{tr}}

\def\0{{\bm 0}}

\def\c{{\bm c}}

\def\v{{\bm v}}
\def\w{{\bm w}}
\def\x{{\bm x}}
\def\y{{\bm y}}

\def\A{{\bm A}}
\def\B{{\bm B}}
\def\C{{\bm C}}

\def\I{{\bm I}}

\def\K{{\bm K}}

\def\Q{{\bm Q}}
\def\S{{\bm S}}
\def\U{{\bm U}}
\def\V{{\bm V}}
\def\W{{\bm W}}
\def\X{{\bm X}}
\def\Y{{\bm Y}}
\def\Z{{\bm Z}}
\def\T{{\bm T}}
\def\G{{\bm G}}

\def\BSigma{\boldsymbol{\Sigma}}
\def\BLambda{\boldsymbol{\Lambda}}
\def\BGamma{\boldsymbol{\Gamma}}
\def\BPhi{\boldsymbol{\Phi}}
\def\BPsi{\boldsymbol{\Psi}}

\def\Btheta{\boldsymbol{\theta}}

\def\E{{\mathbb{E}}}
\def\R{{\mathbb{R}}}

\global\long\def\RR{\mathbb{R}}

\usepackage{tikz}

\usetikzlibrary{decorations.pathmorphing}
\usetikzlibrary{decorations.markings}
\usetikzlibrary{shapes.gates.logic.US,trees,positioning,arrows}

\tikzset{
	arn/.style = {circle, white, draw=black, fill=gray!30, inner sep = 10.5},
	arn_t/.style = {circle, white, draw=black, very thick, fill=gray!30, inner sep = 11.0},
	arn_l/.style = {circle, white, draw=black, very thick, fill=black, inner sep = 2},
	photon/.style={draw=black, very thick, dashed},
	electron/.style={draw=black, very thick},
	tr/.style={buffer gate US,thick,draw,fill=gray!60,rotate=90,	anchor=east,minimum width=2.25cm},
	br/.style={buffer gate US,thick,draw,fill=gray!60,rotate=90,	anchor=east,minimum width=4.5cm},
	brr/.style={buffer gate US,draw,fill=gray!60,rotate=90,	anchor=east,minimum width=4.5cm, opacity = 0.6},
	trr/.style={buffer gate US,thick,draw,fill=gray!60,rotate=90,	anchor=east,minimum width=2.25cm, opacity = 0.6},
	trrr/.style={buffer gate US,draw,fill=white!60,rotate=90,	anchor=east,minimum width=2.25cm, opacity = 0.5}
}

\title{Scaling Neural Tangent Kernels\\ via Sketching and Random Features}

\usepackage[symbol]{footmisc}

\author{
	Amir Zandieh\footnotemark \\ Max-Planck-Institut für Informatik \\ \email{azandieh@mpi-inf.mpg.de}
	\And
	Insu Han\footnotemark[1] \\ Yale University \\ \email{insu.han@yale.edu}
	\And
	Haim Avron \\ Tel Aviv University \\ \email{haimav@tauex.tau.ac.il}
	\AND
	Neta Shoham \\ Tel Aviv University \\ \email{shohamne@gmail.com}
	\And
	Chaewon Kim \\ KAIST \\ \email{chaewonk@kaist.ac.kr} 
	\And
	Jinwoo Shin \\ KAIST \\ \email{jinwoos@kaist.ac.kr} 
}

\begin{document}
\footnotetext[1]{Equal contribution.}
\maketitle
\begin{abstract}
The Neural Tangent Kernel (NTK) characterizes the behavior of infinitely-wide neural networks trained under least squares loss by gradient descent. Recent works also report that NTK regression can outperform finitely-wide neural networks trained on small-scale datasets.
However, the computational complexity of kernel methods has limited its use in large-scale learning tasks.
To accelerate learning with NTK,
we design a near input-sparsity time approximation algorithm for NTK, by sketching the polynomial expansions of arc-cosine kernels: our sketch for the convolutional counterpart of NTK (CNTK) can transform any image using a linear runtime in the number of pixels.
Furthermore, we prove a spectral approximation guarantee for the NTK matrix,
by combining random features (based on leverage score sampling) 
of the arc-cosine kernels 
with a sketching algorithm.
We benchmark our methods on various large-scale regression and classification tasks and show that a linear regressor trained on our CNTK features matches the accuracy of exact CNTK on CIFAR-10 dataset while achieving 150$\times$ speedup.

\end{abstract}

\section{Introduction} \label{sec:intro}

Recent results have shown that over-parameterized Deep Neural Networks (DNNs), generalize surprisingly well. 
In an effort to understand this phenomena, researchers have studied ultra-wide DNNs and shown that in the infinite width limit, a fully connected DNN trained by gradient descent under least-squares loss is equivalent to kernel regression with respect to the Neural Tangent Kernel (NTK)~\cite{arora2019exact, chizat2019lazy, jacot2018neural, lee2020generalized}. This connection has shed light on DNNs' ability to generalize~\cite{cao2019generalization,neyshabur2014search} and optimize (train) their parameters efficiently~\cite{allen2019convergence,arora2019fine,du2018gradient}. More recently, \citet{arora2019exact} proved an analogous equivalence between convolutional DNNs with infinite number of channels and Convolutional NTK (CNTK).
Beyond the aforementioned theoretical purposes, several papers have explored the algorithmic use of this kernel. ~\citet{arora2019harnessing} and ~\citet{geifman2020similarity} showed that NTK based kernel models can outperform trained DNNs (of finite width). Additionally, CNTK kernel regression sets an impressive performance record on CIFAR-10 for kernel methods without trainable kernels~\cite{arora2019exact}. The NTK has also been used in experimental design~\cite{shoham2020experimental} and predicting training time~\cite{zancato2020predicting}.


However, the NTK-based approaches encounter the computational bottlenecks of kernel learning. In particular, for a dataset of $n$ images $x_1,x_2, \ldots x_n \in \RR^{d \times d}$, only writing down the CNTK kernel matrix requires $\Omega\left(d^4 \cdot n^2\right)$ operations \cite{arora2019exact}. Running regression or PCA on the resulting kernel matrix takes additional cubic time in $n$, which is infeasible in large-scale setups. 



There is a rich literature on kernel approximations for large-scale learning. 
One of the most popular approaches is the {\em random features method} which works by randomly sampling the feature space of the kernel function, originally due to the seminal work of~\citet{rahimi2007random}. 
Another popular approach which is developed in linear sketching literature~\cite{woodruff2014sketching}, works by designing sketches that can be efficiently applied to the feature space of a kernel without needing to explicitly form the high dimensional feature space. This approach has been successful at designing efficient subspace embeddings for the polynomial kernel~\cite{avron2014subspace,ahle2020oblivious}.
In this paper, we propose solutions for scaling the NTK and CNTK by building on both of these kernel approximations techniques and designing efficient feature maps that approximate the NTK/CNTK evaluation. 
Consequently, we can simply transform the input dataset to these feature spaces, and then apply fast linear learning methods to approximate the answer of the corresponding nonlinear kernel method efficiently.
The performance of such approximate methods is similar or sometimes better than the exact kernel methods due to implicit regularization effects of the approximation algorithms~\cite{rahimi2007random,rudi2016generalization,jacot2020implicit}.

\subsection{Overview of Our Contributions}
\begin{itemize}[wide, labelwidth=!, labelindent=5pt]

\item One of our results is an efficient random features construction for the NTK. Our starting point is the explicit NTK feature map suggested by \citet{bietti2019inductive} based on tensor product of the feature maps of arc-cosine kernels. We obtain our random features, by sampling the feature space of arc-cosine kernels~\cite{cho2009kernel}.
However, the na\"ive construction of the features would incur an exponential cost in the depth of the NTK, due to the tensor product of features generated in consecutive layers. We remedy this issue, by utilizing an efficient sketching algorithm for tensor products known as \TSRHT~\cite{ahle2020oblivious} which can effectively approximate the tensor products of vectors while preserving their inner products. 
We provide a rigorous error analysis of the proposed scheme in Theorem~\ref{thm:ntk-random-features-error}.

\item Our next results are sketching methods for both NTK and CNTK using a runtime that is linearly proportional to the sparsity of the input dataset (or number of pixels of images). Our methods rely on the arc-cosine kernels' feature space defined by their Taylor expansion. By careful truncation of their Taylor series, we approximate the arc-cosine kernels with bounded-degree polynomial kernels. 
Because the feature space of a polynomial kernel is the tensor product of its input space, its dimensionality is exponential in the degree of the kernel. Fortunately,~\citet{ahle2020oblivious} have developed a linear sketch known as \PolyS that can reduce the dimensionality of high-degree tensor products very efficiently, therefore, we can sketch the resulting polynomial kernels using this technique. We then combine the transformed features from consecutive layers by further sketching their tensor products. In case of CNTK, we have an extra operation which sketches the direct sum of the features of neighbouring pixels at each layer that precisely corresponds to the convolution operation in CNNs. We carefully analyze the errors introduced by polynomial approximations and various sketching steps in our algorithms and also bound their runtimes in Theorems~\ref{mainthm-ntk}~and~\ref{maintheorem-cntk}.

\item Furthermore, we improve the arc-cosine random features to spectrally approximate the entire kernel matrix, which is advocated in recent literature for ensuring high approximation quality in downstream tasks~\cite{avron2017random, musco2017recursive}. 
 Our construction is based on leverage score sampling, which entertains better convergence bounds~\cite{avron2017random, lee2020generalized, li2021towards}. 
 However, computing this distribution is as expensive as solving the kernel methods exactly.
We propose a simple distribution that tightly upper bounds the leverage scores of arc-cosine kernels and for further efficiency, use Gibbs sampling to generate random features from our proposed distribution.
We provide our spectral approximation guarantee in Theorem~\ref{thm:ntk_spectral}.

\item Finally, we empirically benchmark our proposed methods on various classification/regression tasks and demonstrate that our methods perform similar to or better than exact kernel method with NTK and CNTK while running extremely faster. In particular, we classify CIFAR-10 dataset 150$\times$ faster than exact CNTK and at the same time achieve higher test accuracy.
\end{itemize}

\subsection{Related Works}
There has been a long line of work on the correspondence between DNN and kernel machines \cite{lee2018deep,matthews2018gaussian,novak2018bayesian,garriga2018deep,yang2019scaling}.
Furthermore, there has been many efforts in understanding a variety of NTK properties including optimization~\cite{lee2019wide,allen2019convergence, du2018gradient, zou2018stochastic}, generalization~\cite{cao2019generalization}, loss surface~\cite{mei2018mean}, etc. 

\citet{novak2018bayesian} tried accelerating CNTK computations via Monte Carlo methods by taking the gradient of a randomly initialized CNN with respect to its weights. Although they do not theoretically bound the number of required features, the fully-connected version of this method is analyzed in \cite{arora2019exact}. Particularly, for the gradient features to approximate the NTK up to $\varepsilon$, the network width needs to be $\Omega(\frac{L^{6}}{\varepsilon^4} \log \frac{L}{\delta})$, thus, transforming a single vector $x\in \mathbb{R}^{d}$ requires $\Omega(\frac{L^{13}}{\varepsilon^8}  \log^2 \frac{L}{\delta} + \frac{L^{6}}{\varepsilon^4}  \log \frac{L}{\delta} \cdot {\rm nnz}(x))$ operations, which is slower than our Theorem~\ref{mainthm-ntk} by a factor of $L^3/\varepsilon^2$.
Furthermore, \cite{arora2019exact} shows that the performance of these random gradients is worse than exact CNTK by a large margin, in practice. 
More recently, \cite{lee2020generalized} proposed leverage score sampling for the NTK, however, their work is primarily theoretical and suggests no practical way of sampling the features.
Another line of work on NTK approximation is an explicit feature map construction via tensor product proposed by \citet{bietti2019inductive}. These explicit features can have infinite dimension in general and even if one uses a finite-dimensional approximation to the features, the computational gain of random features will be lost due to expensive tensor product operations.

A popular line of work on kernel approximation problem is based on the Fourier features method~\cite{rahimi2007random}, which works well for shift-invariant kernels and with some modifications can embed the Gaussian kernel near optimally~\cite{avron2017random}. Other random feature constructions have been suggested for a variety of kernels, e.g., arc-cosine kernels~\cite{cho2009kernel}, polynomial kernels~\cite{pennington2015spherical}.
In linear sketching literature,~\citet{avron2014subspace} proposed a subspace embedding for the polynomial kernel which was recently extended to general dot product kernels~\cite{han2020polynomial}. The runtime of this method, while nearly linear in sparsity of the input dataset, scales exponentially in kernel's degree. Recently,~\citet{ahle2020oblivious} improved this exponential dependence to polynomial which enabled them to sketch high-degree polynomial kernels and led to near-optimal embeddings for Gaussian kernel. In fact, this sketching technology constitutes one of the main ingredients of our proposed methods. 
Additionally, combining sketching with leverage score sampling can improve the runtime of the polynomial kernel embeddings~\cite{woodruff2020near}.

\subsection{Preliminaries: \PolyS and \TSRHT Transforms} \label{sec:preliminaries}
\paragraph{Notations.} 
We use $[n]:=\{1,\dots,n\}$.  We denote the tensor (a.k.a. Kronecker) product by $\otimes$ and the element-wise (a.k.a. Hadamard) product of two vectors or matrices by $\odot$. 
Although tensor products are multidimensional objects, we often associate $x\otimes y$ with a single dimensional vector $(x_1y_1,x_2y_1, \ldots x_my_1, x_1y_2, \ldots x_my_2, \ldots x_my_n)$. For shorthand, we use the notation $x^{\otimes p}$ to denote $\underbrace{x\otimes \ldots \otimes x}_{p \text{ terms}}$, the $p$-fold self-tensoring of $x$. Another related operation that we use is the \emph{direct sum} of vectors: $x\oplus y:= \left[ x^\top, y^\top \right]^\top$. We need notation for \emph{sub-tensors} of a tensor. For instance, for a $3$-dimensional tensor $\Y \in \RR^{m \times n \times d}$ and every $l \in [d]$, we denote by $\Y_{(:,:,l)}$ the $m \times n$ matrix that is defined as $\left[\Y_{(:,:,l)}\right]_{i,j} := \Y_{i,j,l}$ for $i \in[m], j \in [n]$.  
For square matrices $\A$ and $\B$, we write $\A \preceq \B$ if $\B - \A$ is positive semi-definite. 
We also denote $\mathrm{ReLU}(x) = \max(x,0)$ and consider this element-wise operation when the input is a matrix. We use $\mathrm{nnz}(x)$ to denote the number of nonzero entries in $x$.
Given a positive semidefinite matrix $\K$ and $\lambda >0$, the statistical dimension of $\K$ with $\lambda$ is defined as $s_{\lambda}(\K) := \mathtt{tr}(\K (\K + \lambda \I)^{-1})$.
For two functions $f$ and $g$ we denote their twofold composition by $f\circ g$, defined as $f\circ g(\alpha):= f(g(\alpha))$.

The \TSRHT is a norm-preserving dimensionality reduction that can be applied to the tensor product of two vectors very quickly~\cite{ahle2020oblivious}. 
This transformation is a generalization of the Subsampled Randomized Hadamard Transform (SRHT)~\cite{ailon2009fast} and can be computed in near linear time using the FFT algorithm.
The \PolyS extends the idea behind \TSRHT to high-degree tensor products by recursively sketching pairs of vectors in a binary tree structure. This sketch preserves the norm of vectors in $\RR^{d^p}$ with high probability and can be applied to tensor product vectors very quickly.
The following Lemma, summarizes Theorems 1.2 and 1.3 of \cite{ahle2020oblivious} and is proved in Appendix~\ref{appendix-sketch-prelims}.

\begin{restatable}[{\sc PolySketch}]{lemma}{sodaresults}\label{soda-result}
    For every integers $p,d\ge 1$ and every $\varepsilon, \delta>0$, there exists a distribution on random matrices $\Q^p \in \RR^{m \times d^p}$, called degree $p$ \PolyS such that {\bf (1)} for some $m=\bigo\left(\frac{p}{\varepsilon^{2}} \log^3 \frac{1}{\varepsilon\delta} \right)$ and any $y \in \RR^{d^p}$, $\Pr\left[ \|\Q^p y\|_2^2 \in (1\pm \varepsilon)\|y\|_2^2 \right] \ge 1 - \delta$; {\bf (2)} for any $x \in \RR^d$, if $e_1\in\RR^d$ is the standard basis vector along the first coordinate, the total time to compute~$\Q^p (x^{\otimes (p-j)} \otimes {e}_1^{\otimes j})$~for all $j=0,1,\dots,p$ is $\bigo\left( p m \log^2 m + \min\left\{\frac{p^{3/2}}{\varepsilon}\log\frac{1}{\delta}~{\rm nnz}(x), pd\log d\right\} \right)$; {\bf (3)} for any collection of vectors $v_1,\dots,v_p \in \RR^d$, the time to compute $\Q^p \left(v_1 \otimes \dots \otimes v_p\right)$ is bounded by $\bigo\left( p m \log m + \frac{p^{3/2}}{\varepsilon} d \log \frac{1}{\delta} \right)$; {\bf (4)} for any $\lambda>0$ and any matrix $\A \in \RR^{d^p \times n}$, where the statistical dimension of $\A^\top \A$ is $s_\lambda$, there exists some $m = \bigo\left( \frac{p^4 s_\lambda}{\varepsilon^2} \log^3 \frac{n}{\varepsilon\delta} \right)$ such that,
    \begin{align}
        \Pr \left[ (1-\varepsilon) \left(\A^\top \A + \lambda \I\right) \preceq (\Q^p \A)^\top (\Q^p \A)  + \lambda \I \preceq (1+\varepsilon)\left( \A^\top \A + \lambda \I \right) \right] \ge 1 - \delta.
    \end{align}
\end{restatable}

\begin{figure}[!t]
    \centering
    \begin{subfigure}{0.4\textwidth}
        \includegraphics[width=\textwidth]{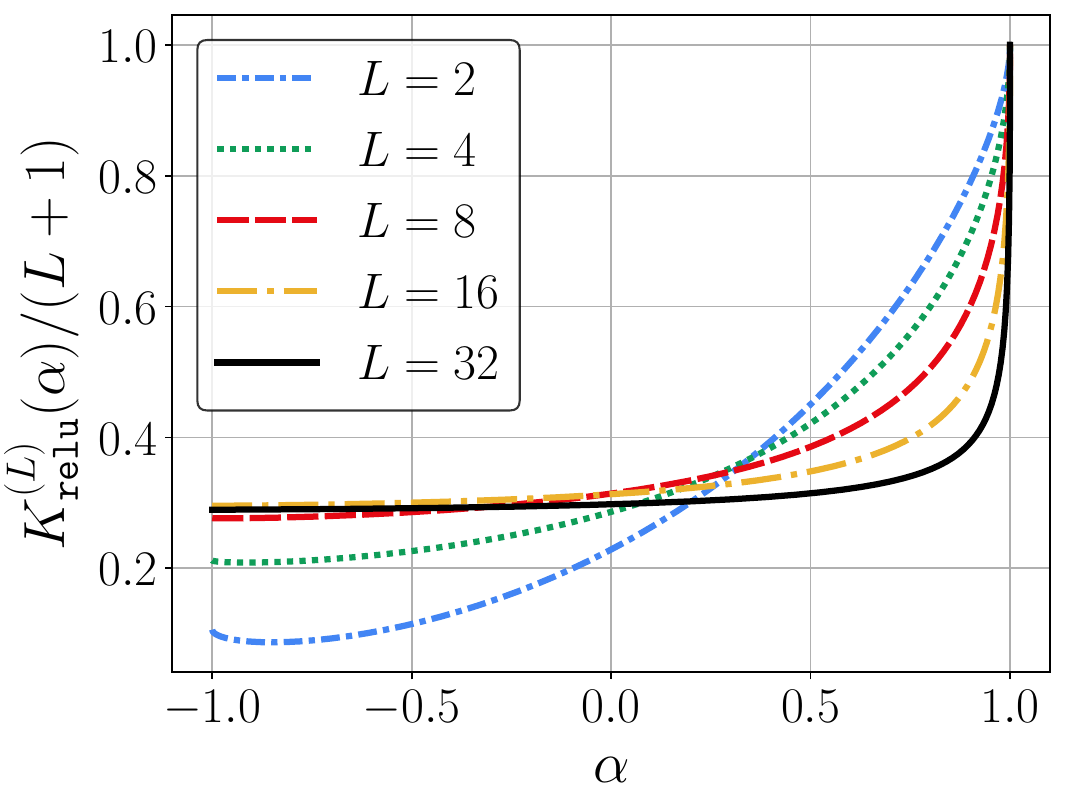}
	\end{subfigure}
	\hskip 20pt
	\begin{subfigure}{0.4\textwidth}
	    \includegraphics[width=\textwidth]{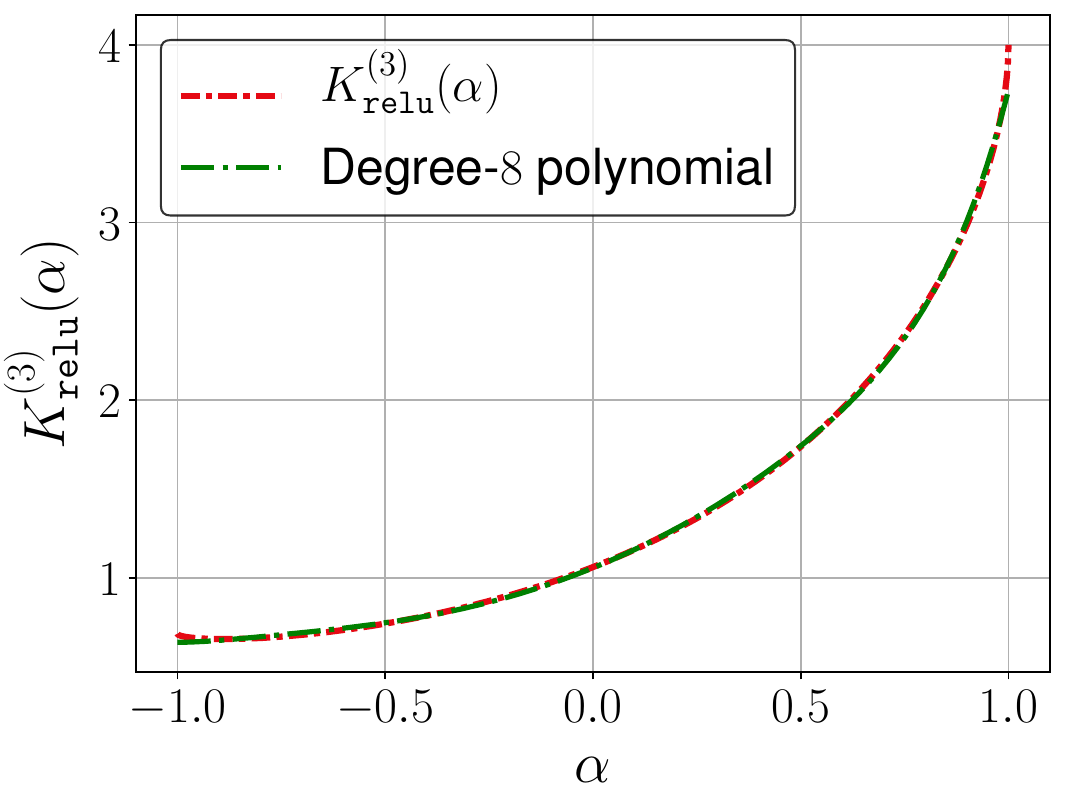} 
	\end{subfigure}
	\vskip -10pt
	\caption{(Left) Normalized ReLU-NTK function $K_{\tt relu}^{(L)}(\cdot)$ for $L=\{2,4,8,16,32\}$ and (Right) a degree-$8$ polynomial approximation of ReLU-NTK with $L=3$.} \label{fig:relu-ntk}
	\vspace{-6pt}
\end{figure}

\section{ReLU Neural Tangent Kernel} \label{sec:mainresults}
\citet{arora2019exact} showed how to exactly compute the NTK of a $L$-layer fully-connected network, denoted by $\Theta_{\tt ntk}^{(L)}(y,z)$, for any pair of vectors $y,z \in \RR^{d}$ using a dynamic program in $\bigo(d + L)$ time. However, it is hard to gain insight into the structure of this kernel using that the dynamic program expression which involves recursive applications of nontrivial expectations.
Fortunately, for the important case of ReLU activation this kernel takes an extremely nice and highly structured form. 
The NTK in this case can be fully characterized by a univariate function $K_{\tt relu}^{(L)}:[-1,1] \to \RR$ that we refer to as \emph{ReLU-NTK}, which is the composition of the arc-cosine kernels~\cite{cho2009kernel} and was recently derived in \cite{bietti2019inductive}. Exploiting this special structure is the key to designing efficient sketching methods and random features for this kernel.

\begin{defn}[ReLU-NTK function]\label{def:relu-ntk}
	For every integer $L>0$, the $L$-layer ReLU-NTK function $K_{\tt relu}^{(L)}:[-1,1] \to \RR$ is defined via following procedure, for every $\alpha \in [-1,1]$:
	\begin{enumerate}[wide, labelwidth=!, labelindent=0pt]
		\item Let $\kappa_0(\alpha)$ and $\kappa_1(\alpha)$ be $0^{th}$ and $1^{st}$ order arc-cosine kernels~\cite{cho2009kernel} defined as follows,
		\begin{align}\label{relu-activ-cov}
				\kappa_0(\alpha) := \frac{1}{\pi} \left( \pi - \arccos\left( \alpha \right) \right), \text{~~~and~~~} \kappa_1(\alpha) := \frac{1}{\pi} \left(\sqrt{1 - \alpha^2} + \alpha \cdot \left( \pi - \arccos\left( \alpha \right) \right) \right).
		\end{align}
		\item Let $\Sigma_{\tt relu}^{(0)}(\alpha) := \alpha$ and for $\ell = 1,2, \ldots L$, define $\Sigma_{\tt relu}^{(\ell)}(\alpha)$ and $\dot{\Sigma}_{\tt relu}^{(\ell)}(\alpha)$ as follows, 
	    \begin{align}\label{eq:dp-covar-relu}
	            \Sigma_{\tt relu}^{(\ell)}(\alpha) := \underbrace{\kappa_1 \circ \kappa_1 \circ \dots \circ \kappa_1}_{\ell \text{-fold self composition}}  (\alpha),\text{~~~and~~~} \dot{\Sigma}_{\tt relu}^{(\ell)}(\alpha) := \kappa_0 \left( \Sigma_{\tt relu}^{(\ell-1)}(\alpha) \right).
	        \end{align}
	    \item Let $K_{\tt relu}^{(0)}(\alpha) := \Sigma_{\tt relu}^{(0)}(\alpha) = \alpha$ and for $\ell=1,2, \ldots L$, define $K_{\tt relu}^{(\ell)}(\alpha)$ recursively as follows,
		    \begin{align}\label{eq:dp-ntk-relu}
			    K_{\tt relu}^{(\ell)}(\alpha) := K_{\tt relu}^{(\ell-1)}(\alpha)\cdot \dot{\Sigma}_{\tt relu}^{(\ell)}(\alpha) + \Sigma_{\tt relu}^{(\ell)}(\alpha).
		    \end{align}
	\end{enumerate}
\end{defn}
The connection between ReLU-NTK function $K_{\tt relu}^{(L)}$ and the NTK kernel $\Theta_{\tt ntk}^{(L)}$ is formalized bellow,
\begin{align}\label{K_relu_NTK_correspondence}
    \Theta_{\tt ntk}^{(L)}(y,z) \equiv \|y\|_2 \|z\|_2 \cdot K_{\tt relu}^{(L)}\left( \frac{\langle y, z \rangle}{\|y\|_2 \|z\|_2} \right), \text{~~~for any } y,z \in \RR^d.
\end{align}
This shows that the NTK is a \emph{normalized dot-product kernel} which can be fully characterized by $K_{\tt relu}^{(L)}:[-1,1] \to \RR$, plotted in \cref{fig:relu-ntk}.
As shown in \cref{fig:relu-ntk}, this function is smooth and can be tightly approximated with a low-degree polynomial. It is evident that for larger values of $L$, $K_{\tt relu}^{(L)}(\cdot)$ converges to a \emph{knee shape}, i.e., it has a nearly constant value of roughly $0.3 (L+1)$ on the interval $[-1,1 - \bigo(L^{-1})]$, and on the interval $[1 - \bigo(L^{-1}), 1]$ its value sharply increases to $L+1$ at $\alpha=1$.
\section{Sketching and Random Features for NTK}
The main results of this section are efficient oblivious sketching as well as random features for the fully-connected NTK. 
As shown in \cref{def:relu-ntk} and \cref{K_relu_NTK_correspondence}, the NTK $\Theta_{\tt ntk}^{(L)}$, is constructed by recursive composition of arc-cosine kernels $\kappa_1(\cdot)$ and $\kappa_0(\cdot)$. So, to design efficient sketches for the NTK we crucially need efficient methods for approximating these functions. Generally, there are two main approaches to approximating these functions; one is random features sampling and the other is truncated Taylor series expansion coupled with fast sketching.
We design algorithms by exploiting both of these techniques.

\vspace{-2pt}
\subsection{NTK Sketch}
\vspace{-1pt}
Our main tool is approximating the arc-cosine kernels with low-degree polynomials, and then applying \PolyS to the resulting polynomial kernels. The features for multi-layer NTK are the recursive tensor product of arc-cosine sketches at consecutive layers, which in turn can be sketched efficiently using {\sc PolySketch}. We present our oblivious sketch in \cref{alg-def-ntk-sketch}.

\begin{algorithm}[t]
\caption{\NTKS for fully-connected ReLU networks} \label{alg-def-ntk-sketch}
\begin{algorithmic}[1]
	\STATE {\bf input}: vector $x \in \RR^{d}$, network depth $L$, error and failure parameters $\varepsilon, \delta > 0$
	\STATE Choose integers $s = \widetilde{\bigo}\left(\frac{L^2}{\varepsilon^2}\right)$, $n_1=\widetilde{\bigo}\left(\frac{L^4}{\varepsilon^4}\right)$, $r = \widetilde{\bigo}\left(\frac{L^6}{\varepsilon^4}\right)$, $m=\widetilde{\bigo}\left( \frac{L^8}{\varepsilon^{\frac{16}{3}}}\right)$, and $s^*=\bigo\left( \frac{1}{\varepsilon^2} \log\frac{1}{\delta}\right)$ appropriately\footnotemark
	
	\STATE For $p = \left\lceil 2L^2/\varepsilon^{\frac{4}{3}} \right\rceil$ and $p' = \left\lceil 9L^2/\varepsilon^{2} \right\rceil$, polynomials $P_{\tt relu}^{(p)}(\cdot)$ and $\dot{P}_{\tt relu}^{(p')}(\cdot)$ are defined as,
		\begin{equation}\label{eq:poly-approx-krelu}
			\begin{split}
				P_{\tt relu}^{(p)}(\alpha) \equiv \sum_{j=0}^{2p+2} c_j \cdot \alpha^j &: = \frac{1}{\pi} + \frac{\alpha}{2}+ \frac{1}{\pi} \sum_{i=0}^p \frac{(2i)! \cdot \alpha^{2i+2}}{2^{2i}  (i!)^2  (2i+1) (2i+2)},\\
				\dot{P}_{\tt relu}^{(p')}(\alpha) \equiv \sum_{j=0}^{2p'+1} b_j \cdot \alpha^j &: = \frac{1}{2} + \frac{1}{\pi} \sum_{i=0}^{p'} \frac{(2i)!}{2^{2i}  (i!)^2  (2i+1)} \cdot \alpha^{2i+1}.
			\end{split}
		\end{equation}
		
	\STATE $\phi^{(0)}(x) \gets \|x\|_2^{-1} \cdot \Q^{1} \cdot  x$, where $\Q^{1} \in \RR^{r \times d}$ is a degree-$1$ \PolyS as per \cref{soda-result} \label{eq:map-covar-zero}

	\STATE $\psi^{(0)}(x) \gets {\bm V} \cdot \phi^{(0)}(x)$, where ${\bm V} \in \RR^{s \times r}$ is an instance of SRHT~\citep{ailon2009fast} \label{eq:map-relu-zero}
	
	\FOR{ $\ell=1$ to $L$}
		\STATE 
		Let $\Q^{2p+2} \in \RR^{m \times r^{2p+2}}$ be a degree-$2p+2$ {\sc PolySketch}.
		Also, let $\T \in \RR^{r \times (2p+3)\cdot m}$ be an instance of SRHT. 
		For every $l=0,1, \ldots, 2p+2$, compute:
		\begin{equation}\label{eq:map-covar}
			Z^{(\ell)}_{l}(x) \gets \Q^{2p+2}  \left(\left[ \phi^{(\ell-1)}(x) \right]^{\otimes l} \otimes {e}_1^{\otimes 2p+2-l}\right),~\phi^{(\ell)}(x) \gets \T  \cdot \bigoplus_{l=0}^{2p+2} \sqrt{c_l}  Z^{(\ell)}_l(x) 
		\end{equation}
		
		\STATE 
		Let $\Q^{2p'+1} \in \RR^{n_1 \times r^{2p'+1}}$ be a degree-$2p'+1$ {\sc PolySketch}. 
		Also, let $\W \in \RR^{s \times (2p'+2)\cdot n_1}$ be an instance of SRHT.
		For every $l=0,1,\ldots, 2p'+1$, compute:
		\begin{equation}\label{eq:map-derivative-covar}
		Y^{(\ell)}_{l}(x) \gets \Q^{2p'+1} \left(\left[ \phi^{(\ell-1)}(x) \right]^{\otimes l} \otimes {e}_1^{\otimes 2p'+1-l}\right),~\dot{\phi}^{(\ell)}(x) \gets \W \cdot  \bigoplus_{l=0}^{2p'+1} \sqrt{b_l}  Y^{(\ell)}_l(x)
		\end{equation}

        \STATE Let $\Q^{2} \in \RR^{s \times s^2}$ be a degree-$2$ {\sc PolySketch}. Also, let ${\bm R} \in \RR^{s \times (s+r)}$ be an SRHT.
		Compute:
		\begin{equation}\label{eq:map-relu}
			\psi^{(\ell)}(x) \gets {\bm R} \cdot \left(\Q^2 \left(\psi^{(\ell-1)}(x) \otimes \dot{\phi}^{(\ell)}(x)\right) \oplus \phi^{(\ell)}(x)\right).
		\end{equation}
		
	\ENDFOR
	\vspace{-7pt}
	
	\STATE Let $\G \in \RR^{s^* \times s}$ be a matrix of i.i.d. entries with distribution $\mathcal{N}(0,\frac{1}{s^*})$. Compute: 
		\begin{equation}\label{Psi-ntk-def}
			\Psi_{\tt ntk}^{(L)}(x) \gets {\|x\|_2} \cdot \G \cdot \psi^{(L)}(x).
	\end{equation}
	
	\STATE {\bf return} ${\Psi}_{\tt ntk}^{(L)}(x)$
\end{algorithmic}
\end{algorithm}
\footnotetext{$\widetilde{\bigo}(\cdot)$ suppresses $\text{poly}(\log\frac{L}{\varepsilon\delta})$ factors.}

Now we present our main theorem on \NTKS algorithm as follows.
\begin{restatable}{theorem}{mainthmntk}\label{mainthm-ntk}
	For every integers $d\ge 1$ and $L\ge 2$, and any $\varepsilon, \delta>0$, let $\Theta_{\tt ntk}^{(L)}:\RR^d \times \RR^d \to \RR$ be the $L$-layer NTK with ReLU activation as per \cref{def:relu-ntk} and \cref{K_relu_NTK_correspondence}.
	Then there exists a randomized map $\Psi_{\tt ntk}^{(L)}: \RR^d \to \RR^{s^*}$ for some $s^* = \bigo\left( \frac{1}{\varepsilon^2}  \log \frac{1}{\delta} \right)$ such that the following invariants hold,
	\begin{enumerate}[wide, labelwidth=!, labelindent=0pt]
	\item For any vectors $y,z \in \RR^d$: $\Pr \left[ \left| \left< \Psi_{\tt ntk}^{(L)}(y) , \Psi_{\tt ntk}^{(L)}(z) \right> - \Theta_{\tt ntk}^{(L)}(y,z) \right| \le \varepsilon\cdot \Theta_{\tt ntk}^{(L)}(y,z) \right] \ge 1-\delta$.
	\item For every vecor $x \in \RR^d$, the time to compute $\Psi_{\tt ntk}^{(L)}(x)$ is $\bigo\left( \frac{L^{11}}{\varepsilon^{6.7}} \log^3 \frac{L}{\varepsilon\delta} + \frac{L^3}{\varepsilon^2} \log \frac{L}{\varepsilon\delta} \cdot {\rm nnz}(x) \right)$.
	\end{enumerate}
\end{restatable}
For a proof, see \cref{appendix-ntk-sketch}.
One can observe that the runtime of our \NTKS is faster than the gradient features of an ultra-wide random DNN, studied by \citet{arora2019exact},
by a factor of $L^3/\varepsilon^2$.

\subsection{NTK Random Features}
The main difference between our random features construction and \NTKS is the use of random features for approximating arc-cosine kernels $\kappa_0$ and $\kappa_1$~in \cref{relu-activ-cov}. For any $x \in \RR^d$, we denote
\small
\begin{align}  
\Phi_0(x) := \sqrt{\frac{2}{m_0}} \ \mathrm{Step} \left( \left[ w_1, \ldots, w_{m_0} \right]^\top x\right),~~~ 
\Phi_1(x) := \sqrt{\frac{2}{m_1}} \ \mathrm{ReLU} \left( \left[ w^\prime_1, \ldots, w_{m_1}^\prime \right]^\top x\right),\label{eq:arccosine_random_features}
\end{align}
\normalsize
where $w_1, \dots, w_{m_0}, w^\prime_{1}, \dots, w^\prime_{m_1}\in \R^d$ are i.i.d. samples from 
$\mathcal{N}({0}, \I_d)$. \citet{cho2009kernel} showed that $\E[\inner{\Phi_0(y), \Phi_0(z)}] = \kappa_0\left(\frac{\langle y, z \rangle}{\|y\|_2 \|z\|_2}\right)$ and $\E[\inner{\Phi_1(y), \Phi_1(z)}] = \|y\|_2 \|z\|_2 \cdot \kappa_1\left(\frac{\langle y, z \rangle}{\|y\|_2 \|z\|_2}\right)$.
The feature map for multi-layer NTK can be obtained by recursive tensoring of random feature maps for arc-cosine kernels at each layer of the network.
However, one major drawback of such explicit tensoring is that the number of features, and thus the runtime, will be exponential in depth $L$. 
In order to make the feature map more compact, we utilize a degree-2 \PolyS $\Q^2$ to reduce the dimension of the tensor products at each layer and get rid of exponential dependence on $L$. 
We present the performance guarantee of our random features, defined in \cref{alg:ntk_random_features}, in \cref{thm:ntk-random-features-error}.

\begin{restatable}{theorem}{randomfeatureserror} \label{thm:ntk-random-features-error}
Given $y, z \in \mathbb{R}^d$ and $L \geq 2$, let $\Theta_{\tt ntk}^{(L)}$ the $L$-layer fully-connected ReLU NTK. For $\varepsilon, \delta>0 $, there exist
$m_0 = \bigo \left(\frac{L^2}{\varepsilon^2}  \log \frac{L}{\delta} \right), 
m_1 = \bigo \left( \frac{L^6}{\varepsilon^4} \log \frac{L}{\delta}\right), m_{s} = \bigo \left(\frac{L^2}{\varepsilon^2} \log^3 \frac{L}{\varepsilon\delta} \right)$,
such that, 
\begin{align}
\Pr\left[ \abs{\inner{{\Psi}_{\tt rf}^{(L)}(y), \Psi_{\tt rf}^{(L)}(z)} - \Theta_{\tt ntk}^{(L)}(y,z)} \le \varepsilon \cdot \Theta_{\tt ntk}^{(L)}(y,z) \right] \geq 1-\delta,
\end{align}
where $\Psi_{\tt rf}^{(L)}(y), \Psi_{\tt rf}^{(L)}(z) \in \R^{ m_1 + m_{\mathtt{s}}}$ are the outputs of \cref{alg:ntk_random_features}, using the same randomness.
\end{restatable}

The proof of \cref{thm:ntk-random-features-error} is provided in~\cref{sec:proof-ntk-random-features-error}. 
\citet{arora2019exact} proved that the gradient of randomly initialized ReLU network with finite width can approximate the NTK, but their feature dimension should be $\Omega\left( \frac{L^{13}}{\varepsilon^8} \log^2 \frac{L}{\delta} +  \frac{L^{6}}{\varepsilon^4} \cdot\log \frac{L}{\delta}\cdot d \right)$ which is larger than ours by a factor of $ \frac{L^7}{\varepsilon^4} \log \frac{L}{\delta}$.
In \cref{sec:experiments}, we also empirically show that \cref{alg:ntk_random_features} requires far fewer features than random gradients.

\begin{algorithm}[t]
\caption{Random Features for ReLU NTK via \PolyS} \label{alg:ntk_random_features}
\begin{algorithmic}[1]
	\STATE {\bf input}: vector $x \in \RR^{d}$, network depth $L$, feature dimensions $m_0$, $m_1$, and $m_s$
	\STATE ${\psi}_{\tt rf}^{(0)}(x) \leftarrow {x}/{\|x\|_2}, \phi_{\tt rf}^{(0)}(x) \leftarrow {x}/{\|x\|_2}$
	\FOR{ $\ell=1$ to $L$}
		\STATE 
		$\dot{\phi}_{\tt rf}^{(\ell)}(x) \gets \Phi_0\left( \phi_{\tt rf}^{(\ell-1)}(x)\right)$, where $\Phi_0$ is defines as per \cref{eq:arccosine_random_features} with $m_0$ features \label{alg_phidot1}
		\STATE 
		$\phi_{\tt rf}^{(\ell)}(x) \gets \Phi_1 \left(\phi_{\tt rf}^{(\ell-1)}(x) \right)$, where $\Phi_1$ is defines as per \cref{eq:arccosine_random_features} with $m_1$ features \label{alg_phi1} 
        \STATE Draw a degree-$2$ \PolyS $\Q^2$ that maps to $\RR^{m_{s}}$ and compute: 
        $${\psi}_{\tt rf}^{(\ell)}(x) \gets  \phi_{\tt rf}^{(\ell)}(x) \oplus \Q^2 \cdot \left( \dot{\phi}_{\tt rf}^{(\ell)}(x) \otimes {\psi}_{\tt rf}^{(\ell-1)}(x) \right)$$
	\ENDFOR
	\vspace{-7pt}
	\STATE {\bf return} ${\Psi}_{\tt rf}^{(L)}(x) \gets \|x\|_2 \cdot {\psi}_{\tt rf}^{(L)}(x)$
\end{algorithmic}
\end{algorithm}

\subsection{Spectral Approximation for NTK via Leverage Scores Sampling} \label{sec:ntk_spectral}

Although the above NTK approximations can estimate the kernel function itself, it is still questionable how
it affects the performance of downstream tasks. Several works on kernel approximation adopt spectral approximation bound with regularization $\lambda > 0$ and approximation factor $\varepsilon >0$, that is, 
\begin{align}\label{eq:ntk-spectral-approximation}
(1-\varepsilon) ( \K^{(L)}_{\tt ntk} + \lambda \I)
\preceq
( \BPsi^{(L)} )^\top \BPsi^{(L)}  + \lambda \I
\preceq
(1+\varepsilon) ( \K^{(L)}_{\tt ntk} + \lambda \I),
\end{align}
where $\BPsi^{(L)} := \left[ \Psi^{(L)}(x_1), \dots, \Psi^{(L)}(x_n) \right]$ and $[ \K^{(L)}_{\tt ntk} ]_{i,j} = \Theta_{\tt ntk}^{(L)}(x_i,x_j)$. 
The spectral bound can provide rigorous guarantees for downstream applications including kernel ridge regression~\cite{avron2017random}, clustering and PCA~\cite{musco2017recursive}.
We first provide spectral bounds for arc-cosine kernels, 
then we present our spectral approximation bound for two-layer ReLU networks, which is the first in the literature.
To guarantee that the arc-cosine random features in \cref{eq:arccosine_random_features} provide spectral approximation, we will use the leverage score sampling framework of \cite{avron2017random,lee2020generalized}.  We reduce the variance of random features by performing importance sampling. The challenge is to find a proper modified distribution that certainly reduces the variance. It turns out that the original $0^{th}$ order arc-cosine random features has a small enough variance. More precisely, let $\K_0$ be the $0^{th}$ order arc-cosine kernel matrix, i.e., $[\K_0]_{i,j} = \kappa_0\left( \frac{\langle x_i, x_j \rangle}{\|x_i\|_2 \|x_j\|_2} \right)$, and $\BPhi_0 := \left[ \Phi_0(x_1), \ldots, \Phi_0(x_n) \right]$, where $\Phi_0(x)$ is defined in \cref{eq:arccosine_random_features}. If the number of features $m_0 \ge \frac{8}{3}\frac{n}{\lambda \varepsilon^{2} } \log\left( \frac{16 s_{\lambda}}{\delta} \right)$, then
\begin{align}
    \Pr\left[ (1 - \varepsilon) (\K_0 + \lambda \I) 
    \preceq
    \BPhi_0^\top \BPhi_0 + \lambda \I 
    \preceq 
    (1 + \varepsilon) (\K_0 + \lambda \I) \right] \ge 1 - \delta.
\end{align}

Next, we consider spectral approximation of the $1^{st}$ order arc-cosine kernel. Unlike the previous case, modifications of the sampling distribution are required. 
Specifically, for any $x \in \RR^d$, let
\begin{equation} \label{eq:arccosine_random_features_spectral}
\widetilde{\Phi}_1(x) = \sqrt{\frac{2d}{m_1}} \mathrm{ReLU} \left( \left[ \frac{w_1}{\|w_1\|_2}, \ldots, \frac{w_{m_1}}{\|w_{m_1}\|_2} \right]^\top x\right),
\end{equation}
where $w_1, \dots, w_{m_1} \in \R^d$ are i.i.d. samples from
$p(w) := \frac{1}{(2\pi)^{d/2} d} \norm{w}_2^2 \exp\left(-\frac{1}{2}\norm{w}_2^2\right)$.
For this modified features, let $\X\in \RR^{d \times n}$ be the dataset, $\K_1$ be the $1^{st}$ order arc-cosine kernel matrix, i.e., $[\K_1]_{i,j} = \|x_i\|_2 \|x_j\|_2 \cdot \kappa_1\left( \frac{\langle x_i, x_j \rangle}{\|x_i\|_2 \|x_j\|_2} \right)$, and $\BPhi_1 := \left[ \widetilde{\Phi}_1(x_1), \ldots, \widetilde{\Phi}_1(x_n) \right]$. 
If the number of features $m_1 \ge \frac{8}{3} \frac{d}{\varepsilon^{2} } \cdot \min\left\{ \rank(\X)^2, \frac{\|\X\|_2^2}{\lambda} \right\} \log\left( \frac{16 s_{\lambda}}{\delta} \right)$, then 
\begin{align}
    \Pr\left[(1 - \varepsilon) (\K_1 + \lambda \I) \preceq \BPhi_1^\top \BPhi_1 + \lambda \I \preceq (1 + \varepsilon) (\K_1 + \lambda \I)\right] \geq 1- \delta.
\end{align}

The details are provided in \cref{sec:proof-a0-spectral} and~ \cref{sec:proof-a1-spectral}. We are now ready to state our spectral approximation bound for our modified random features.
\begin{restatable}{theorem}{ntkspectral} \label{thm:ntk_spectral}
Given a dataset $\X \in \R^{d \times n}$ with $\|\X_{(:,i)}\|_2 \le 1$ for every $i \in [n]$, let $\K_{\tt ntk}, \K_0, \K_1$ be kernel matrices for two-layer ReLU NTK and arc-cosine kernels of $0^{th}$ and $1^{st}$ order, respectively. For any $\lambda > 0$, suppose $s_{\lambda}$ is the statistical dimension of $\K_{\tt ntk}$. 
Modify \cref{alg:ntk_random_features} by replacing $\Phi_1(\cdot)$ in line~\ref{alg_phi1} with $\widetilde{\Phi}_1(\cdot)$ defined in \cref{eq:arccosine_random_features_spectral}.
For any $\varepsilon ,\delta >0$, 
let $\BPsi_{\tt rf}^{(L)} \in \R^{(m_1 + m_{s}) \times n}$ be the output matrix of this algorithm with $L=1$.
There exist $m_0 = \bigo\left( \frac{n}{\varepsilon^{2} \lambda} \log \frac{ s_{\lambda}}{\delta} \right), 
m_1 = \bigo\left( \frac{d }{\varepsilon^{2}} \cdot \min \left\{ \rank(\X)^2 , \frac{\norm{\X}_2^2}{\lambda} \right\} \log \frac{ s_{\lambda}}{\delta} \right),
m_{s} = \bigo\left( \frac{1}{ \varepsilon^2}  \cdot \frac{n}{1+\lambda}  \log^3 \frac{n}{\varepsilon\delta} \right)$ such that,
\begin{align} \label{eq:spectral_approximation_ntk_features}
\Pr \left[(1-\varepsilon) \left( \K_{\tt ntk} + \lambda \I\right) \preceq
\left( \BPsi_{\tt rf}^{(L)} \right)^\top \BPsi_{\tt rf}^{(L)} + \lambda \I \preceq (1+\varepsilon) \left( \K_{\tt ntk} + \lambda \I\right) \right] \ge 1 - \delta.
\end{align}
\end{restatable}

For a proof see \cref{sec:proof_ntk_spectral}.
To generalize the current proof technique to deeper networks, one needs a monotone property of arc-cosine kernels, i.e., $\kappa_1(\X) \preceq \kappa_1(\Y)$ for $\X \preceq \Y$. However, this property does not hold in general and we leave the extension to deeper networks to future work.


\section{Sketching Convolutional Neural Tangent Kernel}

In this section, we design and analyze an efficient sketching method for the Convolutional Neural Tangent Kernel (CNTK).
We focus mainly on CNTK with Global Average Pooling (GAP), which exhibits superior empirical performance compared to vanilla CNTK with no pooling~\cite{arora2019exact}, however, our techniques can be applied to the vanilla version, as well.
Using the DP of \citet{arora2019exact}, the number of operations needed for exact computation of the depth-$L$ CNTK value $\Theta_{\tt cntk}^{(L)}(y,z)$ for images $y,z \in \RR^{d \times d}$ is $\Omega\left( d^4 \cdot L \right)$, which is extremely slow particularly due to its quadratic dependence on the number of pixels of input images $d^2$.
Fortunately, we are able to show that the CNTK for the important case of ReLU activation is a highly structured object that can be fully characterized in terms of tensoring and composition of arc-cosine kernels, and exploiting this special structure is key to designing efficient sketching methods for the CNTK.
Unlike the fully-connected NTK, CNTK is not a simple dot-product kernel function like~\cref{K_relu_NTK_correspondence}. The key reason being that CNTK works by partitioning its input images into patches and locally transforming the patches at each layer, as opposed to the NTK which operates on the entire input vectors.
We present our derivation of the ReLU CNTK function and its main properties in \cref{app-cntk-expression}.

Similar to \NTKS our method relies on approximating the arc-cosine kernels with low-degree polynomials via Taylor expansion, and then applying \PolyS to the resulting polynomial kernels. Our sketch computes the features for each pixel of the input image, by tensor product of arc-cosine sketches at consecutive layers, which in turn can be sketched efficiently using \PolyS. Additionally, the features of pixels that lie in the same patch get \emph{locally combined} at each layer via direct sum operation. This precisely corresponds to the convolution operation in neural networks. We present our \CNTKS algorithm in \cref{app-cntk-sketch} and give its performance guarantee in the following theorem.

\begin{restatable}{theorem}{maintheoremcntk}\label{maintheorem-cntk}
	For every positive integers $d_1,d_2,c$ and $L \ge 2$, and every $\varepsilon, \delta>0$, if we let $\Theta_{\tt cntk}^{(L)}:\RR^{d_1\times d_2 \times c} \times \RR^{d_1\times d_2\times c} \to \RR$ be the $L$-layer CNTK with ReLU activation and GAP given in \cite{arora2019exact},
	then there exist a randomized map $\Psi_{\tt cntk}^{(L)}: \RR^{d_1 \times d_2 \times c} \to \RR^{s^*}$ for some $s^* = \bigo\left( \frac{1}{\varepsilon^2} \log \frac{1}{\delta} \right)$ such that:
	\begin{enumerate}[wide, labelwidth=!, labelindent=0pt]
		\item For any images $y,z \in \RR^{d_1 \times d_2 \times c}$: 
		$$\Pr \left[ \left| \left< \Psi_{\tt cntk}^{(L)}(y) , \Psi_{\tt cntk}^{(L)}(z) \right> - \Theta_{\tt cntk}^{(L)}(y,z) \right| \leq \varepsilon \cdot \Theta_{\tt cntk}^{(L)}(y,z)\right] \geq 1 - \delta.$$
		\item For every image $x \in \RR^{d_1 \times d_2 \times c}$, time to compute $\Psi_{\tt cntk}^{(L)}(x)$ is $\bigo\left( \frac{L^{11}}{\varepsilon^{6.7}} \cdot (d_1d_2) \cdot \log^3 \frac{d_1d_2L}{\varepsilon\delta} \right)$.
	\end{enumerate}
\end{restatable}
The proof is in \cref{app-cntk-sketch}. Runtime of our \CNTKS is only linear in the number of image pixels $d_1d_2$, which is in stark contrast to quadratic scaling of the exact CNTK computation \cite{arora2019exact}.

\vspace{-4pt}
\section{Experiments}\label{sec:experiments}
\vspace{-4pt}

\begin{figure}[t]
	\centering
	\begin{subfigure}{0.48\textwidth}
		\includegraphics[width=0.495\textwidth]{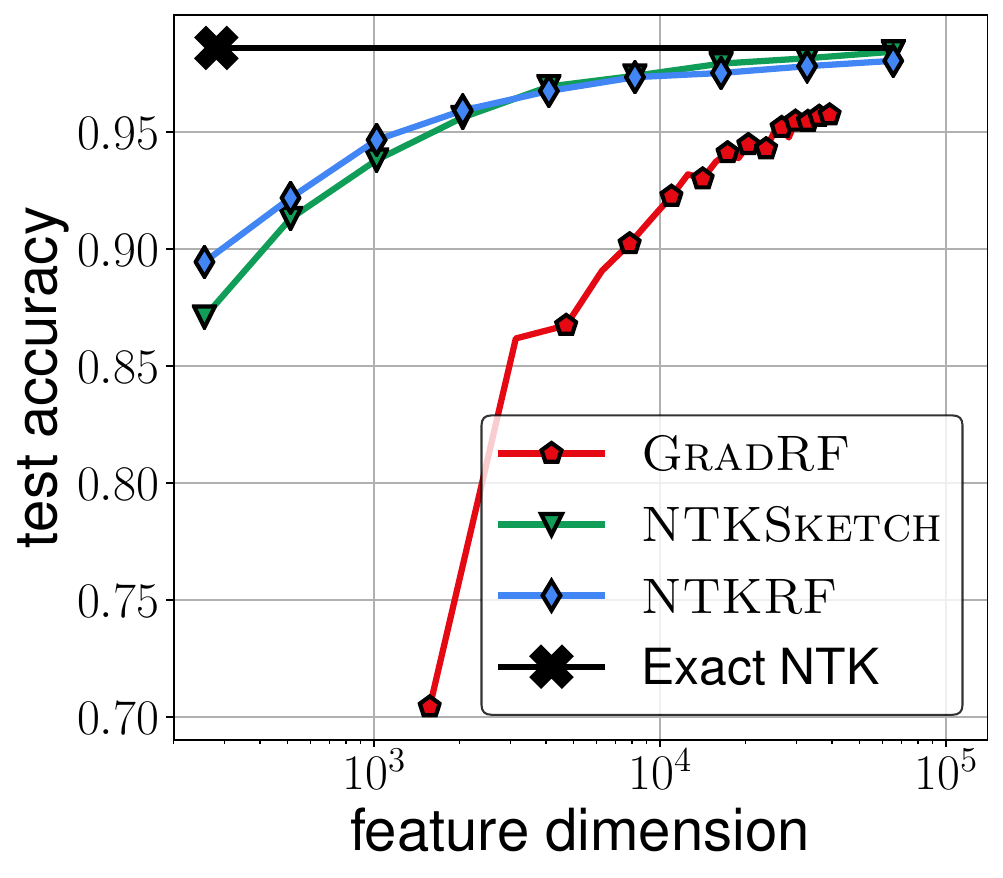}
		\includegraphics[width=0.495\textwidth]{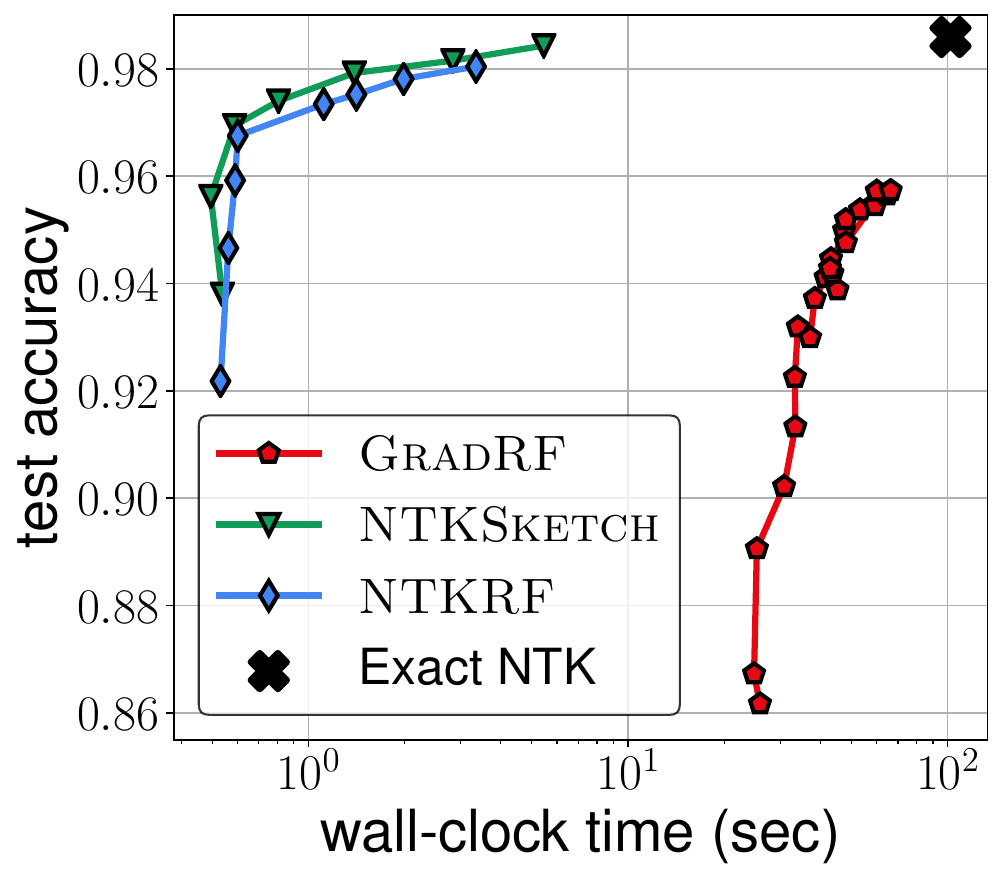}
    	\vskip -0.01in
		\caption{MNIST with NTK}\label{fig:MNIST-NTK}
	\end{subfigure}
	\begin{subfigure}{0.48\textwidth}
		\includegraphics[width=0.495\textwidth]{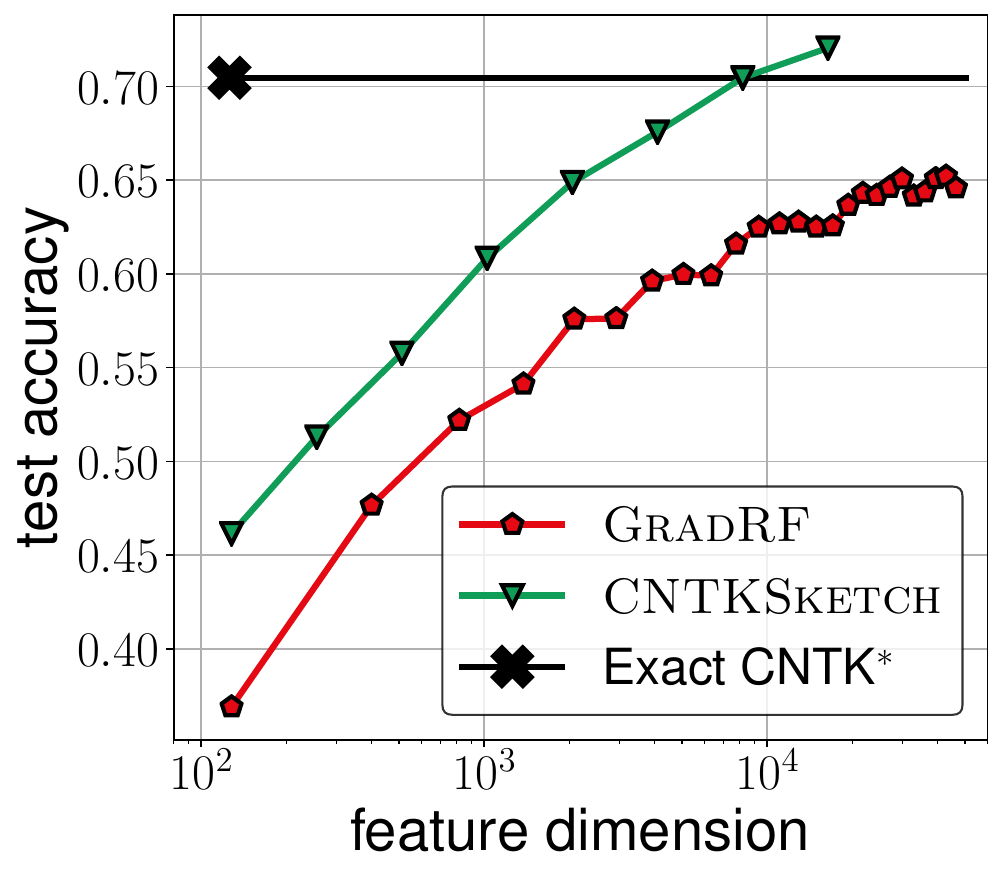}
		\includegraphics[width=0.495\textwidth]{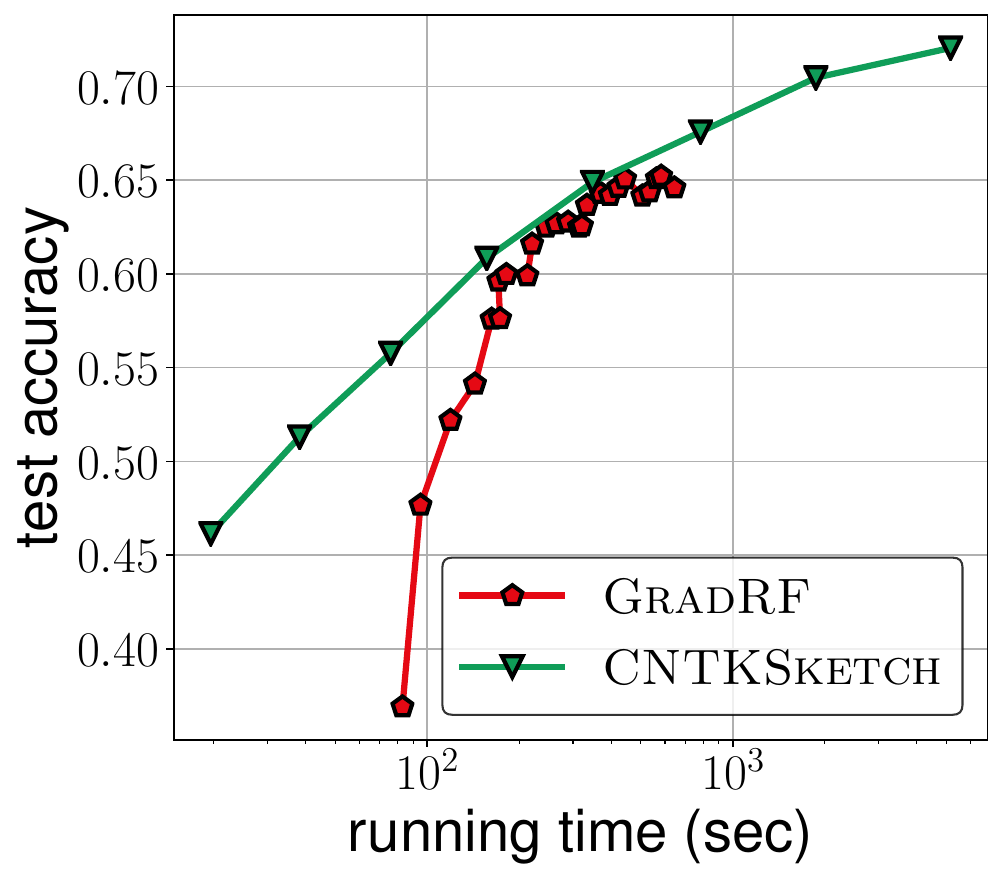}
    	\vskip -0.01in
		\caption{CIFAR-10 with CNTK}\label{fig:cifar-cntk}
	\end{subfigure}
	\vspace{-0.069in}
	\caption{Test accuracy of: (a) approximate NTK methods ({\sc GradRF}, \NTKS and {\sc NTKRF}) on MNIST and (b) approximate CNTK methods (\GradRF and {\sc CNTKSketch}) on CIFAR-10.} \label{fig:mse_vs_featdim}
	\vspace{-0.15in}
\end{figure}

In this section, we empirically show that running least squares regression on the features generated by our methods is extremely fast and effective for learning with NTK and CNTK kernel machines.
We run experiments on a system with an Intel E5-2630 CPU with 256 GB RAM and a single GeForce RTX 2080 GPUs with 12 GB RAM.
Codes are available at \url{https://github.com/insuhan/ntk-sketch-rf}.

\subsection{NTK Classification on MNIST}
We first benchmark our proposed NTK approximation algorithms on MNIST~\cite{lecun2010mnist} dataset and compare against gradient-based NTK random features~\cite{arora2019exact} ({\sc GradRF}) as a baseline method. To apply our methods and {\sc GradRF} into classification task, we encode class labels into one-hot vectors with zero-mean and solve the ridge regression problem. We search the ridge parameter with a random subset of training set and choose the one that achieves the best validation accuracy.
We use the ReLU network with depth $L = 1$. 
In \cref{fig:MNIST-NTK}, we observe that our random features (\NTKRF) achieves the best test accuracy. The \NTKS narrowly follows the performance of \NTKRF and the Grad-RF is the worst method which confirms the observations of \citet{arora2019exact}, i.e., gradient of a finite width network degrades practical performances.

\begin{rmk}[Optimizing \NTKS for Deeper Nets]
As shown in \cref{K_relu_NTK_correspondence}, the NTK is a normalized dot-product kernel characterized by the function $K^{(L)}_{\tt relu} (\alpha)$. This function can be easily computed using $\bigo(L)$ operations at any desired $\alpha \in [-1,1]$, therefore, we can efficiently fit a polynomial to this function using numerical methods (for instance, it is shown in~\cref{fig:relu-ntk} that a degree-$8$ polynomial can tightly approximate the depth-$3$ ReLU-NTK function $K_{\tt relu}^{(3)}$). Then, we can efficiently sketch the resulting polynomial kernel using \PolyS, as was previously done for Gaussian and general dot-product kernels \citep{ahle2020oblivious,woodruff2020near}. 
Therefore, we can accelerate our \NTKS for deeper networks ($L>2$), using this heuristic.
\end{rmk}

\subsection{CNTK Classification on CIFAR-10} 

Next we test our \CNTKS on CIFAR-10 dataset~\cite{Krizhevsky09learningmultiple}. We choose a convolutional network of depth $L=3$ and compare \CNTKS and \GradRF for various feature dimensions. We borrow results of both CNTK and CNN from \citet{arora2019exact}. The results are provided in \cref{fig:cifar-cntk} and \cref{tab:cifar-cntk}. Somewhat surprisingly, \CNTKS even performs better than the exact CNTK regression by achieving $72.06\%$ when feature dimension is set to $16{,}384$. The likely explanation is that \CNTKS takes advantages of implicit regularization effects of approximate feature map and powerful expressiveness of the CNTK. 
Moreover, computing the CNTK matrix takes over 250 hours (12 days) under our setting which is at least $150 \times$ slower than our {\sc CNTKSketch}. 

\begin{table}[t]
\addtolength{\tabcolsep}{1.3pt}
\vspace{-8pt}
\caption{Test accuracy and runtime to solve CNTK regression and its approximations on CIFAR-10. (*) means that the result is copied from \citet{arora2019exact}.} \label{tab:cifar-cntk}
\begin{center}
\scalebox{0.85}{
\begin{tabular}{@{}lcccccccc@{}}
\toprule
 & \multicolumn{3}{c}{\CNTKS} (ours) &  \multicolumn{3}{c}{\sc GradRF} & Exact CNTK & CNN\\ \cmidrule(lr){2-4}\cmidrule(lr){5-7}
Feature dimension & $4{,}096$ & $8{,}192$ & $16{,}384$ & $9{,}328$ & $17{,}040$ & $42{,}816$ &  \\
\midrule
Test accuracy (\%) & 67.58& 70.46 & {72.06} & 62.49 & 62.57 & 65.21 & 70.47$^*$ & 63.81$^*$ \\
Time (s)  & 780 & 1{,}870 & 5{,}160 & 300& 360& 580 & > 1,000,000 &  \\
\bottomrule
\end{tabular}
}
\end{center}
\vspace{-8pt}
\end{table}
\addtolength{\tabcolsep}{-1.3pt}

\addtolength{\tabcolsep}{1.2pt}
\begin{table*}[t]
	\caption{MSE and runtime on large-scale UCI datasets. We measure the entire time to solve kernel ridge regression. $(-)$ means Out-of-Memory error.} \label{tab:large-scale-uci}
	\centering
	\scalebox{0.83}{
		\begin{tabular}{@{}lcccccccc@{}}
			\toprule
			& \multicolumn{2}{c}{\MSD} & \multicolumn{2}{c}{\WorkLoads} & \multicolumn{2}{c}{\CT} & \multicolumn{2}{c}{\Protein} \\ \midrule
			\# of data points ($n$) & \multicolumn{2}{c}{467{,}315} & \multicolumn{2}{c}{179{,}585}  &  \multicolumn{2}{c}{53{,}500} & \multicolumn{2}{c}{39{,}617} \\\midrule
			& MSE   & Time (s) & MSE & Time (s) & MSE & Time (s) & MSE & Time (s) \\ \midrule
			RBF Kernel  & $-$ & $-$ & $-$ & $-$ & 35.37 & 59.23 & {18.96} & 46.45 \\
			RFF         & 109.50 & 231 & $4.034\times 10^{4}$ & 53.0 & 48.20 & 15.2 & 19.72 & 12.1 \\ \midrule
			NTK         & $-$ & $-$ & $-$ & $-$ & 30.52 & 72.10 & 20.24 & 76.93 \\
			\NTKRF (ours) & 94.27 & 95 & $3.554 \times 10^{4}$ & 35.7 & 46.91 & 2.12 & 20.51 & 4.3 \\
			\NTKS  (ours) & {92.83} & 36 & $3.538\times 10^{4}$ & 27.5 & 46.52 & 18.8 & 21.19 & 14.91 \\
			\bottomrule
		\end{tabular}
	}
\end{table*}
\addtolength{\tabcolsep}{-1.2pt}

\vspace{-4pt}
\subsection{Regression on Large-scale UCI Datasets} 
\vspace{-4pt}
We also demonstrate the computational efficiency of our \NTKS and \NTKRF using $4$ large-scale UCI regression datasets~\cite{dua2019} by comparing against exact NTK, RBF as well as Random Fourier Features (RFF). 
For our methods and RFF, we fix the output dimension to $m=8{,}192$ for all datasets.
In \cref{tab:large-scale-uci}, we report the runtime to compute feature map or kernel matrix and evaluate the averaged mean squared errors (MSE) on the test set via $4$-fold cross validation. The exact kernel methods face Out-of-Memory error on larger datasets. The proposed NTK approximations are significantly faster than the exact NTK, e.g., \NTKRF shows up to 30$\times$ speedup under \CT{} dataset.
We also verify that, except for \Protein{} dataset, our methods outperform RFF.

\vspace{-4pt}
\section{Discussion and Conclusion}
\vspace{-4pt}
In this work, we propose efficient low-rank feature maps for the NTK and CNTK kernel matrices based on both sketching and random features. Computing NTK have been raised severe computational problems when they apply to practical applications. Our methods runs remarkably faster than the NTK with performance improvement. 

{\bf Potential negative societal impact.} 
This is a technical work proposing provable algorithms 
which stand alone independently of data, e.g., do not learn any private information of input data. We think there is no particular potential negative
societal impact due to our work.


{\bf Limitations.} 
This paper only considers
fully-connected and convolutional neural networks,
and our ideas are not directly applicable to scale up NTK of other deep networks, 
e.g., transformers~\cite{hron2020infinite}.

\section*{Acknowledgments and Disclosure of Funding}
Amir Zandieh was partially supported by the Swiss NSF grant No. P2ELP2\textunderscore 195140. Haim Avron and Neta Shoham were partially supported by BSF grant 2017698 and ISF grant 1272/17. Jinwoo Shin was partially supported by the Engineering Research Center Program through the National Research Foundation of Korea (NRF) funded by the Korean Government MSIT (NRF\textunderscore2018R1A5A1059921) and Institute of Information \& communications Technology Planning \& Evaluation (IITP) grant funded by the Korea government (MSIT) (No.2019\textunderscore0\textunderscore00075, Artificial Intelligence Graduate School Program (KAIST)).

\bibliographystyle{plainnat}
\setcitestyle{numbers}
\bibliography{references}
\newpage
\appendix
\newcommand{\KNTK}{\boldsymbol \Theta}
\newcommand{\KNNGP}{\boldsymbol \Sigma}

\section{ReLU-NTK Expression}\label{appendix-relu-ntk}
\citet{arora2019exact} showed how to exactly compute the $L$-layer NTK with activation $\sigma:\RR \to \RR$ using the following dynamic program (DP):
\begin{enumerate}[wide, labelwidth=!, labelindent=0pt]
	\item For every $y,z\in \RR^d$, let $\Sigma^{(0)}(y,z) := \langle y , z\rangle$ and for every layer $h = 1,2, \ldots L $, recursively define the covariance $\Sigma^{(h)}: \RR^d \times \RR^d \to \RR$ as:
	\begin{equation}\label{eq:dp-covar}
		\begin{split}
			&\Lambda^{(h)}(y,z) := \begin{pmatrix}
				\Sigma^{(h-1)}(y,y) & \Sigma^{(h-1)}(y,z)\\
				\Sigma^{(h-1)}(z,y) & \Sigma^{(h-1)}(z,z)
			\end{pmatrix},\\
			&\Sigma^{(h)}(y,z) := \frac{\EE_{(u,v) \sim \mathcal{N}\left( 0, \Lambda^{(h)}(y,z) \right)} \left[ \sigma(u) \cdot \sigma(v) \right]}{\EE_{x\sim \mathcal{N}(0,1)} \left[ |\sigma(x)|^2 \right]}.
		\end{split}
	\end{equation}
	\item For $h = 1,2, \ldots L$, define the derivative covariance as,
	\begin{equation}\label{eq:dp-derivative-covar}
		\dot{\Sigma}^{(h)}(y,z) := \frac{\EE_{(u,v) \sim \mathcal{N}\left( 0, \Lambda^{(h)}(y,z) \right)} \left[ \dot{\sigma}(u) \cdot \dot{\sigma}(v) \right]}{\EE_{x\sim \mathcal{N}(0,1)} \left[ |\sigma(x)|^2 \right]}.
	\end{equation}
	\normalsize
	\item Let $\Theta_{\tt ntk}^{(0)}(y,z) := \Sigma^{(0)}(y,z)$ and for every integer $L \ge 1$, the depth-$L$ NTK expression is defined recursively as:
	\begin{equation}\label{eq:dp-ntk}
		\Theta_{\tt ntk}^{(L)}(y,z) := \Theta_{\tt ntk}^{(L-1)}(y,z) \cdot \dot{\Sigma}^{(L)}(y,z) + \Sigma^{(L)}(y,z).
	\end{equation}
\end{enumerate}

While using this DP, one can compute the kernel value $\Theta_{\tt ntk}^{(L)}(y,z)$ for any pair of vectors $y,z \in \RR^{d}$ using $\bigo(d + L)$ operations, it is hard to gain insight into the structure of this kernel using the expression above. 
In particular, the NTK expression involves recursive computation of nontrivial expectations which makes it is unclear whether there exist efficient sketching solutions for this kernel in its current form.
However, we show that for the important case of ReLU activation, this kernel takes an extremely nice and highly structured form. In fact, the NTK can be fully characterized by a \emph{univariate function} $K_{\tt relu}^{(L)}:[-1,1]\to \RR$, and exploiting this special structure is the key to designing efficient sketching methods for this kernel.

Now we proceed to prove \cref{K_relu_NTK_correspondence}, that is, 
\begin{align*}
    \Theta_{\tt ntk}^{(L)}(y,z) \equiv \|y\|_2 \|z\|_2 \cdot K_{\tt relu}^{(L)}\left( \frac{\langle y, z \rangle}{\|y\|_2 \|z\|_2} \right), \text{~~~for any } y,z \in \RR^d.\tag{5}
\end{align*}

First note that the main component of the DP given in \cref{eq:dp-covar}, \cref{eq:dp-derivative-covar}, and \cref{eq:dp-ntk} is the \emph{Activation Covariances}:
	\[\EE_{w \sim \mathcal{N}(0, \I_d)} \left[ \sigma(w^\top y) \cdot \sigma(w^\top z) \right],~ \text{ and } \EE_{w \sim \mathcal{N}(0, \I_d)} \left[ \dot{\sigma}(w^\top y) \cdot \dot{\sigma}(w^\top z) \right] \text{~~ for every }y,z \in \RR^d. \]
	
It is worth noting that the above activation covariance functions are positive definite and hence they both define valid kernel functions in $\RR^d \times \RR^d$.
The connection between the ReLU activation covariance functions and arc-cosine kernel functions defined in \cref{relu-activ-cov} of \cref{def:relu-ntk} is proved in \citet{cho2009kernel}. We restate this result as follows,
\begin{equation}\label{relu-covariance}
    \begin{split}
        \EE_{w \sim \mathcal{N}(0, \I_d)} \left[ \mathrm{ReLU}(w^\top y) \cdot \mathrm{ReLU}(w^\top z) \right] &= \frac{\|y\|_2\|z\|_2}{2} \cdot \kappa_1 \left(\frac{\langle y, z \rangle}{\|y\|_2\|z\|_2}\right)\\
        \EE_{w \sim \mathcal{N}(0, \I_d)} \left[ \mathrm{Step}(w^\top y) \cdot \mathrm{Step}(w^\top z) \right] &= \frac{1}{2} \cdot \kappa_0\left(\frac{\langle y, z \rangle}{\|y\|_2\|z\|_2}\right).
    \end{split}
\end{equation}

{\it Proof of \cref{K_relu_NTK_correspondence}:}
	Consider the NTK expression given in \cref{eq:dp-covar}, \cref{eq:dp-derivative-covar}, and \cref{eq:dp-ntk}.
	We first prove by induction on $h =0,1,2, \ldots L$ that the covariance function $\Sigma^{(h)}(y,z)$ defined in \cref{eq:dp-covar} satisfies: 
	\[\Sigma^{(h)}(y,z) = \|y\|_2 \|z\|_2 \cdot \Sigma_{\tt relu}^{(h)} \left( \frac{\langle y,z \rangle}{\|y\|_2 \|z\|_2} \right).\]
	The {\bf base of induction} is trivial for $h=0$ due to $\Sigma^{(0)}(y,z) = \langle y,z \rangle = \|y\|_2 \|z\|_2 \cdot \Sigma_{\tt relu}^{(0)} \left( \frac{\langle y,z \rangle}{\|y\|_2 \|z\|_2} \right)$.
	
	To prove the {\bf inductive step}, suppose that the inductive hypothesis holds for $h-1$, i.e.,
	\[\Sigma^{(h-1)}(y,z) = \|y\|_2 \|z\|_2 \cdot \Sigma_{\tt relu}^{(h-1)} \left( \frac{\langle y,z \rangle}{\|y\|_2 \|z\|_2} \right)\]
	The $2 \times 2$ covariance matrix $\Lambda^{(h)}(y,z)$, defined in \cref{eq:dp-covar}, can be decomposed as $\Lambda^{(h)}(y,z) = \begin{pmatrix} f^\top \\ g^\top \end{pmatrix} \cdot \begin{pmatrix}
		f & g \end{pmatrix}$, where $f,g \in \RR^2$. Now note that $\|f\|_2^2 = \Sigma^{(h-1)}(y,y)$, hence, by inductive hypothesis, we have,
	$\|f\|_2^2 = \|y\|_2^2 \cdot \Sigma_{\tt relu}^{(h-1)} \left( \frac{\langle y,y \rangle}{\|y\|_2^2} \right) = \|y\|_2^2 \cdot \Sigma_{\tt relu}^{(h-1)}(1) = \|y\|_2^2.$
	
	Using a similar argument we have $\|g\|_2^2 = \|z\|_2^2$. Therefore, by \cref{relu-covariance}, we can write
	\begin{align*}
		\Sigma^{(h)}(y,z) &= \frac{1}{\EE_{x\sim \mathcal{N}(0,1)} \left[ |\sigma(x)|^2 \right]} \cdot \EE_{w \sim \mathcal{N}\left( 0, I_2 \right)} \left[ \sigma(w^\top f) \cdot \sigma(w^\top g) \right]\\
		&= \frac{2}{\kappa_1(1)} \cdot \EE_{w \sim \mathcal{N}\left( 0, I_2 \right)} \left[ \sigma(w^\top f) \cdot \sigma(w^\top g) \right]\\
		&= \|f\|_2 \|g\|_2 \cdot \kappa_1\left( \frac{\langle f, g \rangle}{\|f\|_2 \|g\|_2} \right) = \|y\|_2 \|z\|_2 \cdot \kappa_1\left( \frac{\langle f, g \rangle}{\|y\|_2 \|z\|_2} \right).
	\end{align*}
	Since we assumed that $\Lambda^{(h)}(y,z) = \begin{pmatrix} f^\top \\ g^\top \end{pmatrix} \cdot \begin{pmatrix}
		f & g \end{pmatrix}$, we have $\langle f, g \rangle = \Sigma^{(h-1)}(y,z)$. By inductive hypothesis along with \cref{eq:dp-covar-relu}, we find that
	\[ \Sigma^{(h)}(y,z) = \|y\|_2 \|z\|_2 \cdot \kappa_1\left( \Sigma_{\tt relu}^{(h-1)} \left( \frac{\langle y,z \rangle}{\|y\|_2 \|z\|_2} \right) \right) = \|y\|_2 \|z\|_2 \cdot \Sigma_{\tt relu}^{(h)}\left( \frac{\langle y, z\rangle}{\|y\|_2 \|z\|_2}  \right), \]
	which completes the induction and proves that for every $h =0,1,\dots,L$,
	\begin{equation}\label{eq:covariance-dotproduct}
		\Sigma^{(h)}(y,z) = \|y\|_2 \|z\|_2 \cdot \Sigma_{\tt relu}^{(h)} \left( \frac{\langle y,z \rangle}{\|y\|_2 \|z\|_2} \right).
	\end{equation}
	
	For obtaining the final NTK expression we also need to simplify the derivative covariance function defined in \cref{eq:dp-derivative-covar}.
	Recall that we proved before, the covariance matrix $\Lambda^{(h)}(y,z)$ can be decomposed as $\Lambda^{(h)}(y,z) = \begin{pmatrix} f^\top \\ g^\top \end{pmatrix} \cdot \begin{pmatrix}
		f & g \end{pmatrix}$, where $f,g \in \RR^2$ with $\|f\|_2 = \|y\|_2$ and $\|g\|_2=  \|z\|_2$. Therefore, by \cref{relu-covariance}, we can write
	\begin{align*}
		\dot{\Sigma}^{(h)}( y, z ) &= \frac{1}{\EE_{x\sim \mathcal{N}(0,1)} \left[ |\sigma(x)|^2 \right]} \cdot \EE_{w \sim \mathcal{N}\left( 0, I_2 \right)} \left[ \dot{\sigma}(w^\top f) \cdot \dot{\sigma}(w^\top g) \right] \\
		&=\frac{2}{\kappa_1(1)} \cdot \EE_{w \sim \mathcal{N}\left( 0, I_2 \right)} \left[ \dot{\sigma}(w^\top f) \cdot \dot{\sigma}(w^\top g) \right]\\
		&= \kappa_0\left( \frac{\langle f, g \rangle}{\|y\|_2 \|z\|_2} \right).
	\end{align*}
	Since we assumed that $\Lambda^{(h)}(y,z) = \begin{pmatrix} f^\top \\ g^\top \end{pmatrix} \cdot \begin{pmatrix}
		f & g \end{pmatrix}$, $\langle f, g \rangle = \Sigma^{(h-1)}(y,z) = \|y\|_2 \|z\|_2 \cdot \Sigma_{\tt relu}^{(h-1)} \left( \frac{\langle y,z \rangle}{\|y\|_2 \|z\|_2} \right)$. Therefore, by \cref{eq:dp-covar-relu}, for every $h \in \{1,2, \dots L\}$,
	\begin{equation}\label{eq:der-cov-dotproduct}
		\dot{\Sigma}^{(h)}(y,z) = \kappa_0\left( \Sigma_{\tt relu}^{(h-1)}\left( \frac{\langle y, z \rangle}{\|y\|_2 \|z\|_2} \right) \right) = \dot{\Sigma}_{\tt relu}^{(h)}\left( \frac{\langle y, z \rangle}{\|y\|_2 \|z\|_2} \right).
	\end{equation}
	
	Now we prove by induction on integer $L$ that the NTK with $L$ layers and ReLU activation given in \cref{eq:dp-ntk} satisfies 
	\[\Theta_{\tt ntk}^{(L)}(y,z) = \|y\|_2 \|z\|_2 \cdot K_{\tt relu}^{(L)}\left( \frac{\langle y,z\rangle}{\|y\|_2 \|z\|_2} \right).\]
	The {\bf base of induction}, trivially holds because, by \cref{eq:covariance-dotproduct}:
	\[\Theta_{\tt ntk}^{(0)}(y,z)=\Sigma^{(0)}(y,z)=\|y\|_2 \|z\|_2 \cdot K^{(0)}_{\tt relu}\left( \frac{\langle y,z\rangle}{\|y\|_2 \|z\|_2} \right).\]
	
	To prove the {\bf inductive step}, suppose that the inducive hypothesis holds for $L-1$, that is $\Theta_{\tt ntk}^{(L-1)}(y,z) = \|y\|_2 \|z\|_2 \cdot K_{\tt relu}^{(L-1)}\left( \frac{\langle y,z\rangle}{\|y\|_2 \|z\|_2} \right)$. Now using the recursive definition of $\Theta_{\tt ntk}^{(L)}(y,z)$ given in \cref{eq:dp-ntk} along with \cref{eq:covariance-dotproduct} and \cref{eq:der-cov-dotproduct} we can write,
	\begin{align*}
		\Theta_{\tt ntk}^{(L)}(y,z)
		&= \Theta_{\tt ntk}^{(L-1)}(y,z) \cdot \dot{\Sigma}^{(L)}(y,z) + \Sigma^{(L)}(y,z)\\
		&= \|y\|_2 \|z\|_2 \cdot K_{\tt relu}^{(L-1)}\left( \frac{\langle y,z\rangle}{\|y\|_2 \|z\|_2} \right) \cdot \dot{\Sigma}_{\tt relu}^{(h)}\left( \frac{\langle y, z \rangle}{\|y\|_2 \|z\|_2} \right) +  \|y\|_2 \|z\|_2 \cdot \Sigma_{\tt relu}^{(h)} \left( \frac{\langle y,z \rangle}{\|y\|_2 \|z\|_2} \right)\\
		&\equiv \|y\|_2 \|z\|_2 \cdot K_{\tt relu}^{(L)}\left( \frac{\langle y,z\rangle}{\|y\|_2 \|z\|_2} \right).
	\end{align*}
	This completes the proof of \cref{K_relu_NTK_correspondence}.
\qed

\section{Sketching Preliminaries: \PolyS and \SRHT}\label{appendix-sketch-prelims}

Our sketching algorithms use the Subsampled Randomized Hadamard Transform (SRHT)~\cite{ailon2009fast} to reduce the dimensionality of the intermediate vectors that arise in our computations. Next lemma gives the performance of SRHT sketches which is proved, for instance, in Theorem 9 of \cite{cohen2016optimal},
\begin{lemma}[SRHT Sketch]\label{lem:srht}
	For every positive integer $d$ and every $\varepsilon, \delta>0$, there exists a distribution on random matrices $\S \in \RR^{m \times d}$ with $m = \bigo\left(\frac{1}{\varepsilon^2} \cdot \log^2 \frac{1}{\varepsilon\delta} \right)$, called {\bf SRHT sketch}, such that for any vector $x \in \RR^{d}$,
	$\Pr\left[ \|\S x\|_2^2 \in (1\pm\varepsilon)\|x\|_2^2 \right] \ge 1 - \delta$.
	Moreover, $\S x$ can be computed in time $\bigo\left(\frac{1}{\varepsilon^2} \cdot \log^2 \frac{1}{\varepsilon\delta} + d\log d \right)$.
\end{lemma}

Now we restate the \cref{soda-result} and present the proof,

\sodaresults*

{\it Proof of \cref{soda-result}:}
    The fourth statement of the lemma immediately follows from Theorem 1.3 of \citet{ahle2020oblivious}.
	Moreover, by invoking Theorem 1.2 of \citet{ahle2020oblivious}, we find that there exists a random sketch $\Q^p \in\RR^{m \times d^p}$ such that $m = C \cdot \frac{p}{\varepsilon^2} \log^3 \frac{1}{\varepsilon\delta}$, for some absolute constant $C$, and for any $y\in \RR^{d^p}$, 
	\begin{align*}
	    \Pr\left[ \|\Q^p y \|_2^2 \in (1\pm\varepsilon) \|y\|_2^2\right] \ge 1- \delta.
	\end{align*}
	
	This immediately proves the first statement of the lemma.
	
	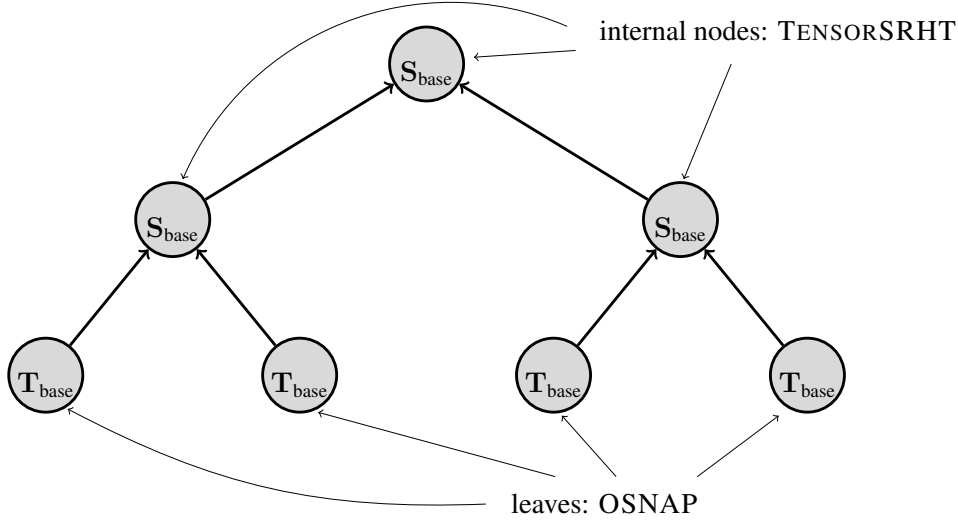
\begin{figure*}
		\centering
		
		\scalebox{0.9}{
			\begin{tikzpicture}[<-, level/.style={sibling distance=75mm/#1,level distance = 2.3cm}]
				\node [arn_t] (z){}
				child {node [arn_t] (a){}edge from parent [electron]
					child {node [arn_t] (b){}edge from parent [electron]
					}
					child {node [arn_t] (e){}edge from parent [electron]
					}
				}
				child { node [arn_t] (h){}edge from parent [electron]
					child {node [arn_t] (i){}edge from parent [electron]
					}
					child {node [arn_t] (l){}edge from parent [electron]
					}
				};
				
				\node []	at (z.south)	[label=\large{${\bf S_{\text{base}}}$}]	{};

				\node []	at (a.south)	[label=\large{${\bf S_{\text{base}}}$}]	{};

				\node []	at (b.south)	[label=\large${\bf T_{\text{base}}}$]	{};
				\node []	at (e.south)	[label=\large${\bf T_{\text{base}}}$] {};
				\node []	at (h.south)	[label=\large{${\bf S_{\text{base}}}$}] {};

				\node []	at (i.south)	[label=\large${\bf T_{\text{base}}}$] {};
				\node []	at (l.south)	[label=\large${\bf T_{\text{base}}}$] {};

				\draw[draw=black, ->] (2.2,0.2) -- (0.7,0.1);
				\draw[draw=black, ->] (4.5,0) -- (3.8,-1.7);
				
				\node [] at (2.2,0.5) [label=right:\large{ internal nodes: {\sc TensorSRHT}}]	{}
				edge[->, bend right=45] (-3.6,-1.7);
				
				\draw[draw=black, ->] (4.0,-6.1) -- (5.2,-5.2);
				\draw[draw=black, ->] (2.8,-6.1) -- (2,-5.2);
				\draw[draw=black, ->] (1.9,-6.1) -- (-1.6,-5.15);
				
				\node [] at (1,-6.5) [label=right:\large{leaves: {\sc OSNAP}}]	{}
				edge[->, bend left=15] (-5.3,-5.1);
				
			\end{tikzpicture}
		}
		\par

		\caption{The structure of sketch $\Q^p$ proposed in Theorem 1.2 of \cite{ahle2020oblivious}: the sketch matrices in nodes of the tree labeled with $\S_{\text{base}}$ and $\T_{\text{base}}$ are independent instances of {\sc TensorSRHT} and {\sc OSNAP}, respectively.} \label{sketchingtree}
	\end{figure*}
	
	As shown in \cite{ahle2020oblivious}, the sketch $\Q^p$ can be applied to tensor product vectors of the form $v_1 \otimes v_2 \otimes \ldots v_p$ by recursive application of $\bigo(p)$ independent instances of OSNAP transform~\cite{nelson2013osnap} and a novel variant of the \SRHT, proposed in \cite{ahle2020oblivious} called {\sc TensorSRHT}, on vectors $v_i$ and their sketched versions. The sketch $\Q^p$, as shown in \cref{sketchingtree}, can be represented by a binary tree with $p$ leaves where the leaves are {\sc OSNAP} sketches and the internal nodes are the {\sc TensorSRHT}. The use of {\sc OSNAP} in the leaves of this sketch structure ensures excellent runtime for sketching sparse input vectors. However, note that if the input vectors are not sparse, i.e., ${\rm nnz}(v_i) = \widetilde{\Omega}(d)$ for input vectors $v_i$, then we can simply remove the {\sc OSNAP} transforms from the leaves of this structure and achieve improved runtime, without hurting the approximation guarantee. Therefore, the sketch $\Q^p$ that satisfies the statement of the lemma is exactly the one introduced in \cite{ahle2020oblivious} for sparse input vectors and for non-sparse inputs is obtained by removing the {\sc OSNAP} transforms from the leaves of the sketch structure given in \cref{sketchingtree}.
	
	{\bf Runtime analysis:} By Theorem 1.2 of \cite{ahle2020oblivious}, for any vector $x \in \RR^d$, $\Q^p x^{\otimes p}$ can be computed in time $\bigo\left( p m \log m + \frac{p^{3/2}}{\varepsilon} \log \frac{1}{\delta} \cdot {\rm nnz}(x) \right)$. 
	From the binary tree structure of the sketch, shown in \cref{sketchingtree}, it follows that once we compute $\Q^p x^{\otimes p}$, then $\Q^p \left(x^{\otimes p-1} \otimes {e}_1\right)$ can be computed by updating the path from one of the leaves to the root of the binary tree which amounts to applying an instance of {\sc OSNAP} transform on the ${e}_1$ vector and then applying $\bigo(\log p)$ instances of \TSRHT on the intermediate nodes of the tree. This can be computed in a total additional runtime of $\bigo( m\log m \log p )$. By this argument, it follows that $\Q^p \left( x^{\otimes p-j} \otimes {e}_1^{j} \right)$ can be computed sequentially for all $j=0,1,2, \cdots p$ in total time $\bigo\left( p m \log p \log m + \frac{p^{3/2}}{\varepsilon} \log\frac{1}{\delta} \cdot {\rm nnz}(x) \right)$. By plugging in the value $m = \bigo\left( \frac{p}{\varepsilon^2} \log^3 \frac{1}{\varepsilon\delta} \right)$, this runtime will be $\bigo\left( \frac{p^2 \log^2 \frac{p}{\varepsilon}}{\varepsilon^2} \log^3 \frac{1}{\varepsilon\delta} + \frac{p^{3/2}}{\varepsilon} \log\frac{1}{\delta} \cdot {\rm nnz}(x) \right)$, which gives the second statement of the lemma for sparse input vectors $x$. If $x$ is non-sparse, as we discussed in the above paragraph, we just need to omit the {\sc OSNAP} transforms from our sketch construction which translates into a runtime of $\bigo\left( \frac{p^2 \log^2 \frac{p}{\varepsilon}}{\varepsilon^2} \log^3 \frac{1}{\varepsilon\delta} + pd \log d \right)$. Therefore, the final runtime bound is $\bigo\left( \frac{p^2 \log^2 \frac{p}{\varepsilon}}{\varepsilon^2} \log^3 \frac{1}{\varepsilon\delta} + \min \left\{\frac{p^{3/2}}{\varepsilon} \log\frac{1}{\delta} \cdot {\rm nnz}(x), pd\log d \right\} \right)$, which proves the second statement of the lemma.
	
	Furthermore, the sketch $\Q^p$ can be applied to tensor product of any collection of $p$ vectors. The time to apply $\Q^p$ to the tensor product $v_1\otimes v_2\otimes \ldots  v_p$ consists of time of applying OSNAP to each of the vectors $v_1, v_2 \ldots v_p$ and time of applying $\bigo(p)$ instances of \TSRHT to intermediate vectors which are of size $m$. This runtime can be upper bounded by $\bigo\left( \frac{p^2 \log \frac{p}{\varepsilon}}{\varepsilon^2} \log^3 \frac{1}{\varepsilon\delta} + \frac{p^{3/2}}{\varepsilon} d \cdot \log\frac{1}{\delta} \right)$, which proves the third statement of the \cref{soda-result}.
\qed

\section{NTK Sketch: Claims and Invariants}\label{appendix-ntk-sketch}
We start by proving that the polynomials $P_{\tt relu}^{(p)}(\cdot)$ and $\dot{P}_{\tt relu}^{(p')}(\cdot)$ defined in \cref{eq:poly-approx-krelu} of \cref{alg-def-ntk-sketch} closely approximate the arc-cosine functions $\kappa_1(\cdot)$ and $\kappa_0(\cdot)$ on the interval $[-1,1]$.

{\bf Remark.} Observe that $\kappa_0(\alpha) = \frac{d}{d\alpha}\left( \kappa_1(\alpha) \right)$.

\begin{lemma}[Polynomial Approximations to $\kappa_1$ and $\kappa_0$]\label{lem:polynomi-approx-krelu}
	If we let $\kappa_1(\cdot)$ and $\kappa_0(\cdot)$ be defined as in \cref{relu-activ-cov} of \cref{def:relu-ntk}, then for any integer $p \ge \frac{1}{9 \varepsilon^{2/3}}$, the polynomial $P_{\tt relu}^{(p)}(\cdot)$ defined in \cref{eq:poly-approx-krelu} of \cref{alg-def-ntk-sketch} satisfies,
	\[ \max_{\alpha \in [-1,1]} \left| P_{\tt relu}^{(p)}(\alpha) - \kappa_1(\alpha) \right| \le \varepsilon.\]
	Moreover, for any integer $p' \ge \frac{1}{26 \varepsilon^2}$, the polynomial $\dot{P}_{\tt relu}^{(p')}(\cdot)$ defined as in \cref{eq:poly-approx-krelu} of \cref{alg-def-ntk-sketch}, satisfies,
	\[ \max_{\alpha \in [-1,1]} \left| \dot{P}_{\tt relu}^{(p')}(\alpha) - \kappa_0(\alpha) \right| \le \varepsilon.\]
\end{lemma}
{\it Proof of \cref{lem:polynomi-approx-krelu}:}
	We start by Taylor series expansion of $\kappa_0(\cdot)$ around $\alpha=0$, $\kappa_0(\alpha) = \frac{1}{2} + \frac{1}{\pi} \cdot \sum_{i=0}^\infty \frac{(2i)!}{2^{2i} \cdot (i!)^2 \cdot (2i+1)} \cdot \alpha^{2i+1}$.
	Therefore, we have
	\begin{align*}
		\max_{\alpha \in [-1,1]} \left| \dot{P}_{\tt relu}^{(p')}(\alpha) - \kappa_0(\alpha) \right| &= \frac{1}{\pi} \cdot \sum_{i=p'+1}^\infty \frac{(2i)!}{2^{2i} \cdot (i!)^2 \cdot (2i+1)}\\
		&\le \frac{1}{\pi} \cdot \sum_{i = p'+1}^\infty \frac{e \cdot e^{-2i} \cdot (2i)^{2i + 1/2}}{ 2\pi \cdot 2^{2i} \cdot e^{-2i} \cdot n^{2i + 1} \cdot (2i+1)}\\
		&= \frac{e}{\sqrt{2} \pi^2} \cdot \sum_{i = p'+1}^\infty \frac{1}{\sqrt{i} \cdot (2i+1)}\\
		&\le \frac{e}{\sqrt{2} \pi^2} \cdot \int_{p'}^\infty \frac{1}{\sqrt{x} \cdot (2x+1)} dx\\
		&\le \frac{e}{\sqrt{2} \pi^2} \cdot \frac{1}{\sqrt{p'}} \le \varepsilon.
	\end{align*}
	
	To prove the second part of the lemma, we consider the Taylor expansion of $\kappa_1(\cdot)$ at $\alpha=0$. Since $\kappa_0(\alpha) = \frac{d}{d\alpha}\left(\kappa_1(\alpha)\right)$, the Taylor series of $\kappa_1(\alpha)$ can be obtained from the Taylor series of $\kappa_0(\alpha)$ as follows,
	\[ \kappa_1(\alpha) = \frac{1}{\pi} + \frac{\alpha}{2} + \frac{1}{\pi} \cdot \sum_{i=0}^\infty \frac{(2i)!}{2^{2i} \cdot (i!)^2 \cdot (2i+1)\cdot (2i+2)} \cdot \alpha^{2i+2}. \]
	Hence, we have
	\begin{align*}
		\max_{\alpha \in [-1,1]} \left| P_{\tt relu}^{(p)}(\alpha) - \kappa_1(\alpha) \right| &= \frac{1}{\pi} \cdot \sum_{i=p+1}^\infty \frac{(2i)!}{2^{2i} \cdot (i!)^2 \cdot (2i+1) \cdot (2i+2)}\\
		&\le \frac{1}{\pi} \cdot \sum_{i = p+1}^\infty \frac{e \cdot e^{-2i} \cdot (2i)^{2i + 1/2}}{ 2\pi \cdot 2^{2i} \cdot e^{-2i} \cdot n^{2i + 1} \cdot (2i+1) \cdot (2i+2)}\\
		&= \frac{e}{\sqrt{2} \pi^2} \cdot \sum_{i = p+1}^\infty \frac{1}{\sqrt{i} \cdot (2i+1) \cdot (2i+2) }\\
		&\le \frac{e}{\sqrt{2} \pi^2} \cdot \int_{p}^\infty \frac{1}{\sqrt{x} \cdot (2x+1) \cdot (2x+2)} dx\\
		&\le \frac{e}{\sqrt{2} \pi^2} \cdot \frac{1}{6 \cdot p^{3/2}} \le \varepsilon.
	\end{align*}
This completes the proof of \cref{lem:polynomi-approx-krelu}.
\qed.

Therefore, it is possible to approximate the function $\kappa_0(\cdot)$ up to error $\varepsilon$ using a polynomial of degree $\bigo\left(\frac{1}{ \varepsilon^2}\right)$. Also if we want to approximate $\kappa_1(\cdot)$ using a polynomial up to error $\varepsilon$ on the interval $[-1,1]$, it suffices to use a polynomial of degree $\bigo\left(\frac{1}{\varepsilon^{2/3} }\right)$.
One can see that since the Taylor expansions of $\kappa_1$ and $\kappa_0$ contain non-negative coefficients only, both of these functions are positive definite. Additionally, the polynomial approximations $P_{\tt relu}^{(p)}$ and $\dot{P}_{\tt relu}^{(p')}$ given in \cref{eq:poly-approx-krelu} of \cref{alg-def-ntk-sketch} are positive definite functions.

In order to prove \cref{mainthm-ntk}, we also need the following lemma on the error sensitivity of polynomials $P_{\tt relu}^{(p)}$ and $\dot{P}_{\tt relu}^{(p')}$,
\begin{lemma}[Sensitivity of $P_{\tt relu}^{(p)}$ and $\dot{P}_{\tt relu}^{(p)}$]\label{lema:sensitivity-polynomial}
	For any integer $p\ge 3$, any $\alpha \in [-1,1]$, and any $\alpha'$ such that $|\alpha - \alpha'| \le \frac{1}{6p}$, if we let the polynomials $P_{\tt relu}^{(p)}(\alpha)$ and $\dot{P}_{\tt relu}^{(p)}(\alpha)$ be defined as in \cref{eq:poly-approx-krelu} of \cref{alg-def-ntk-sketch}, then 
	\[ \left| P_{\tt relu}^{(p)}(\alpha) - P_{\tt relu}^{(p)}(\alpha') \right| \le |\alpha - \alpha'|, \]
	and
	\[ \left| \dot{P}_{\tt relu}^{(p)}(\alpha) - \dot{P}_{\tt relu}^{(p)}(\alpha') \right| \le \sqrt{p} \cdot |\alpha - \alpha'|. \]
\end{lemma}
{\it Proof of \cref{lema:sensitivity-polynomial}:}
	Note that an $\alpha' $ that satisfies the preconditions of the lemm, is in the range $\left[-1 - \frac{1}{6p}, 1 + \frac{1}{6p} \right]$. Now we bound the derivative of the polynomial $\dot{P}_{\tt relu}^{(p)}$ on the interval $\left[-1 - \frac{1}{6p}, 1 + \frac{1}{6p} \right]$,
	\begin{align*}
		\max_{\alpha \in \left[-1 - \frac{1}{6p}, 1 + \frac{1}{6p} \right]} \left| \frac{d}{d\alpha} \left(\dot{P}_{\tt relu}^{(p)}(\alpha)\right) \right| &= \frac{1}{\pi} \cdot \sum_{i=0}^p \frac{(2i)!}{2^{2i} \cdot (i!)^2} \cdot \left( 1 + \frac{1}{6p} \right)^{2i}\\
		&\le \frac{1}{\pi} + \frac{e^{4/3}}{\sqrt{2} \pi^2} \cdot \sum_{i=1}^p \frac{1}{\sqrt{i}}\\
		&\le \frac{1}{\pi} + \frac{e^{4/3}}{\sqrt{2} \pi^2} \cdot \int_{0}^p \frac{1}{\sqrt{x}} dx\\
		&\le \sqrt{p},
	\end{align*}
	therefore, the second statement of lemma holds.
	
	To prove the first statement of lemma, we bound the derivative of the polynomial $P_{\tt relu}^{(p)}$ on the interval $\left[-1 - \frac{1}{6p}, 1 + \frac{1}{6p} \right]$ as follows,
	\begin{align*}
		\max_{\alpha \in \left[-1 - \frac{1}{6p}, 1 + \frac{1}{6p} \right]} \left| \frac{d}{d\alpha} \left(P_{\tt relu}^{(p)}(\alpha)\right) \right| &= \frac{1}{\pi} \cdot \sum_{i=0}^p \frac{(2i)!}{2^{2i} \cdot (i!)^2 \cdot (2i+1)} \cdot \left( 1 + \frac{1}{6p} \right)^{2i+1}\\
		&\le \frac{19}{18\pi} + \frac{e^{25/18}}{\sqrt{2} \pi^2} \cdot \sum_{i=1}^p \frac{1}{\sqrt{i} \cdot (2i+1)}\\
		&\le \frac{19}{18\pi} + \frac{e^{25/18}}{\sqrt{2} \pi^2} \cdot \int_{0}^p \frac{1}{\sqrt{x} \cdot (2x+1)} dx\\
		&\le 1,
	\end{align*}
	therefore, the second statement of the lemma follows. This completes the proof of \cref{lema:sensitivity-polynomial}.
\qed

For the rest of this section, we need two basic properties of tensor products and direct sums:
\begin{align}
    \langle x \otimes y , z \otimes w\rangle = \langle x  , z\rangle \cdot \langle y  , w \rangle,\quad
    \langle x \oplus y , z \oplus w\rangle = \langle x  , z\rangle + \langle y  , w \rangle
\end{align}
for vectors $x,y,z,w$ with conforming sizes.


Now we are in a position to analyze the invariants that are maintained throughout the execution of \NTKS (\cref{alg-def-ntk-sketch}):
\begin{lemma}[Invariants of the \NTKS algorithm]
	\label{thm:ntk-sketch-corr}
	For every positive integers $d$ and $L$, every $\varepsilon, \delta>0$, every vectors $y,z \in \RR^d$, if we let $\Sigma_{\tt relu}^{(h)}:[-1,1] \to \RR$ and $K_{\tt relu}^{(h)}:[-1,1] \to \RR$ be the functions defined in \cref{eq:dp-covar-relu} and \cref{eq:dp-ntk-relu} of \cref{alg-def-ntk-sketch}, then with probability at least $1-\delta$ the following invariants hold for every $h =0, 1, 2, \ldots L$:
	\begin{enumerate}
		\item The mapping $\phi^{(h)}(\cdot)$ computed by \NTKS in line~\ref{eq:map-covar-zero} and \cref{eq:map-covar} of \cref{alg-def-ntk-sketch} satisfy
		\[ \left| \left< \phi^{(h)}(y), \phi^{(h)}(z) \right> - \Sigma_{\tt relu}^{(h)}\left( \frac{\langle y , z \rangle}{\|y\|_2 \|z\|_2} \right) \right| \le ({h+1}) \cdot \frac{\varepsilon^2}{60L^3}. \]
		\item The mapping $\psi^{(h)}(\cdot)$ computed by \NTKS in line~\ref{eq:map-relu-zero} and \cref{eq:map-relu} of \cref{alg-def-ntk-sketch} satisfy
		\[ \left| \left< \psi^{(h)}(y), \psi^{(h)}(z) \right> - K_{\tt relu}^{(h)}\left( \frac{\langle y , z \rangle}{\|y\|_2 \|z\|_2} \right) \right| \le \varepsilon \cdot \frac{h^2+1}{10L}. \]
	\end{enumerate}
\end{lemma}

{\it Proof of \cref{thm:ntk-sketch-corr}:}
	The proof is by induction on the value of $h=0,1,2, \ldots L$. 
	More formally, consider the following statements for every $h=0,1,2, \ldots L$:
	\begin{enumerate}[leftmargin=1.5cm]
		\item[${\bf P_1(h):}$]
		\[ \begin{split}
			&\left| \left< \phi^{(h)}(y), \phi^{(h)}(z) \right> - \Sigma_{\tt relu}^{(h)}\left( \frac{\langle y , z \rangle}{\|y\|_2 \|z\|_2} \right) \right| \le  ({h+1}) \cdot \frac{\varepsilon^2}{60L^3},\\
			&\left| \left\| \phi^{(h)}(y) \right\|_2^2 - 1 \right| \le ({h+1}) \cdot \frac{\varepsilon^2}{60L^3}, \text{ and } \left| \left\| \phi^{(h)}(z) \right\|_2^2 - 1 \right| \le ({h+1}) \cdot \frac{\varepsilon^2}{60L^3}.
		\end{split} \]
		\item[${\bf P_2(h):}$]
		\[ \begin{split}
			&\left| \left< \psi^{(h)}(y), \psi^{(h)}(z) \right> - K_{\tt relu}^{(h)}\left( \frac{\langle y , z \rangle}{\|y\|_2 \|z\|_2} \right) \right| \le \varepsilon \cdot \frac{h^2+1}{10L},\\
			&\left| \left\| \psi^{(h)}(y) \right\|_2^2 - K_{\tt relu}^{(h)}(1) \right| \le \varepsilon \cdot \frac{h^2+1}{10L}, \text{ and } \left| \left\| \psi^{(h)}(z) \right\|_2^2 - K_{\tt relu}^{(h)}(1) \right| \le \varepsilon \cdot \frac{h^2+1}{10L}.
		\end{split} \]
	\end{enumerate}
	We prove that the following holds,
	\begin{align}
	\Pr[ P_1(0)] \ge 1 - \bigo\left(\frac{\delta}{L}\right),~~\text{ and }~~\Pr[ P_2(0)|P_1(0)] \ge 1 - \bigo\left(\frac{\delta}{L}\right).	    
	\end{align}
	Additionally, by induction, we prove that for every $h = 1,2, \ldots L$,
	\begin{align}
	   &\Pr \left[ P_1(h) | P_1(h-1) \right] \ge 1 - \bigo\left(\frac{\delta}{L}\right),~~\text{ and } \nonumber \\
	   &\Pr\left[ P_2(h) | P_2(h-1), P_1(h), P_1(h-1) \right] \ge 1 - \bigo\left(\frac{\delta}{L}\right).
	\end{align}
	
	{\bf (1) Base of induction ($h=0$):}
	By line~\ref{eq:map-covar-zero} of \cref{alg-def-ntk-sketch}, $\phi^{(0)}(y) = \frac{1}{\|y\|_2} \cdot \S \cdot \Q^{1} \cdot y$ and $\phi^{(0)}(z) = \frac{1}{\|z\|_2} \cdot \S \cdot \Q^{1} \cdot z$, thus, \cref{lem:srht} implies the following
	\begin{align}
	    \Pr\left[ \left| \left< \phi^{(0)}(y), \phi^{(0)}(z)\right> - \frac{\left< \Q^{1} y, \Q^{1} z \right>}{\|y\|_2\|z\|_2}\right| \le \bigo\left(\frac{\varepsilon^2}{L^3}\right)\cdot \frac{\left\| \Q^{1} y \right\|_2 \| \Q^{1} z \|_2}{\|y\|_2 \|z\|_2} \right] \ge 1 - \bigo\left(\frac{\delta}{L}\right). 
	\end{align} 
	By using the above together with \cref{soda-result} and union bound as well as triangle inequality, we have
	\begin{align}
	    \Pr\left[ \left| \left< \phi^{(0)}(y), \phi^{(0)}(z)\right> - \frac{\left<  y,  z \right>}{\|y\|_2 \|z\|_2} \right| \le \bigo\left(\frac{\varepsilon^2}{L^3}\right) \right] \ge 1 - \bigo\left(\frac{\delta}{L}\right).
	\end{align}
	Similarly, we can prove that
	\begin{align*}
	    &\Pr\left[ \left| \left\| \phi^{(0)}(y)\right\|_2^2 - 1 \right| \le \bigo\left(\frac{\varepsilon^2}{L^3}\right) \right] \ge 1 - \bigo\left(\frac{\delta}{L}\right), \text{ and }\\
	    &\Pr\left[ \left| \left\| \phi^{(0)}(z)\right\|_2^2 - 1 \right| \le \bigo\left(\frac{\varepsilon^2}{L^3}\right) \right] \ge 1 - \bigo\left(\frac{\delta}{L}\right).
	\end{align*}
	Using union bound, this proves the base of induction for statement $P_1(0)$, i.e., 
	\begin{align}
	    \Pr[ P_1(0) ] \ge 1 - \bigo\left(\frac{\delta}{L}\right).
	\end{align}
	
	Moreover, by line~\ref{eq:map-relu-zero} of \cref{alg-def-ntk-sketch}, $\psi^{(0)}(y) = \V \cdot \phi^{(0)}(y)$ and $\psi^{(0)}(z) = \V \cdot \phi^{(0)}(z)$, thus, \cref{lem:srht} implies that,
	\small
	\begin{align}
	    \Pr\left[ \left| \left< \psi^{(0)}(y), \psi^{(0)}(z)\right> - \left< \phi^{(0)}(y), \phi^{(0)}(z) \right> \right| \le \bigo\left(\frac{\varepsilon}{L}\right)\cdot \left\| \phi^{(0)}(y) \right\|_2 \left\| \phi^{(0)}(z)\right\|_2 \right] \ge 1 - \bigo\left(\frac{\delta}{L}\right).
	\end{align}
	\normalsize
	By conditioning on $P_1(0)$ and using the above together with triangle inequality it follows that,
	\begin{align}
	    \Pr\left[ \left| \left< \psi^{(0)}(y), \psi^{(0)}(z)\right> - K_{\tt relu}^{(0)}\left( \frac{\langle y , z \rangle}{\|y\|_2 \|z\|_2} \right) \right| \le \frac{\varepsilon}{10L} \right] \ge 1 - \bigo\left(\frac{\delta}{L}\right).
	\end{align}

	Similarly we can prove that with probability $1 - \bigo\left(\frac{\delta}{L}\right)$ we have $\left| \left\| \psi^{(0)}(y)\right\|_2^2 - K_{\tt relu}^{(0)}( 1 ) \right| \le \frac{\varepsilon}{10L}$ and $\left| \left\| \psi^{(0)}(z)\right\|_2^2 - K_{\tt relu}^{(0)}( 1 ) \right| \le \frac{\varepsilon}{10L}$,	
	which proves the base of induction for the second statement, i.e., $\Pr[P_2(0)|P_1(0)] \ge 1 - \bigo\left({\delta}/{L}\right)$. This completes the base of induction.
	
    {\bf (2) Inductive step:}
	Assume that the inductive hypothesis holds for $h-1$.
	First, note that by \cref{lem:srht} and using \cref{eq:map-covar} we have the following,
	\begin{equation}\label{eq:phi-h}
		\Pr\left[ \left| \left< \phi^{(h)}(y), \phi^{(h)}(z)\right> - \sum_{j=0}^{2p+2} c_j \left<Z^{(h)}_{j}(y), Z^{(h)}_{j}(z)\right> \right| \le \bigo\left(\frac{\varepsilon^2}{L^3}\right) \cdot A \right] \ge 1 - \bigo\left(\frac{\delta}{L}\right),
	\end{equation}
	where $A := \sqrt{\sum_{j=0}^{2p+2} c_j \|Z^{(h)}_{j}(y)\|_2^2} \cdot \sqrt{\sum_{j=0}^{2p+2} c_j \|Z^{(h)}_{j}(z)\|_2^2}$ and the collection of vectors $\left\{Z^{(h)}_{j}(y)\right\}_{j=0}^{2p+2}$ and $\left\{Z^{(h)}_{j}(z)\right\}_{j=0}^{2p+2}$ and coefficients $\{c_j\}_{j=0}^{2p+2}$ are defined as per \cref{eq:map-covar} and \cref{eq:poly-approx-krelu}, respectively. 
	
	By \cref{soda-result} together with union bound, the following inequalities simultaneously hold for all $j =0, \dots, 2p+2$, with probability at least $1 - \bigo\left( \delta / L\right)$:
	\begin{align}
		&\left|\left<Z^{(h)}_{j}(y), Z^{(h)}_{j}(z)\right> - \left<\phi^{(h-1)}(y), \phi^{(h-1)}(z)\right>^j \right| \le \bigo\left( \frac{\varepsilon^2}{L^3} \right) \cdot \left\| \phi^{(h-1)}(y) \right\|_2^j \left\| \phi^{(h-1)}(z)\right\|_2^j \nonumber\\
		&\left\| Z^{(h)}_{j}(y)\right\|_2^2 \le \frac{11}{10} \cdot \left\| \phi^{(h-1)}(y) \right\|_2^{2j} \label{eq:Zj-sketch}\\
		&\left\| Z^{(h)}_{j}(z)\right\|_2^2 \le \frac{11}{10} \cdot \left\| \phi^{(h-1)}(z) \right\|_2^{2j} \nonumber
	\end{align}
	Therefore, by plugging \cref{eq:Zj-sketch} to \cref{eq:phi-h} and using union bound, triangle inequality and Cauchy–Schwarz inequality we find that,
	\small
	\begin{equation} \label{eq:phi-inner-prod-bound}
		\Pr\left[ \left| \left< \phi^{(h)}(y), \phi^{(h)}(z)\right> - P^{(p)}_{\tt relu}\left( \left<\phi^{(h-1)}(y), \phi^{(h-1)}(z)\right> \right) \right| \le \bigo\left(\frac{\varepsilon^2}{L^3}\right) \cdot B \right] \ge 1 - \bigo\left(\frac{\delta}{L}\right),
	\end{equation}
	\normalsize
	where $B:= \sqrt{P^{(p)}_{\tt relu}\left(\|\phi^{(h-1)}(y)\|_2^2\right) \cdot P^{(p)}_{\tt relu}\left(\|\phi^{(h-1)}(z)\|_2^2\right)}$ and $P^{(p)}_{\tt relu}(\alpha) = \sum_{j=0}^{2p+2} c_j \cdot \alpha^j$ is the polynomial defined in \cref{eq:poly-approx-krelu}. Using the inductive hypothesis $P_1(h-1)$, we have that
	\begin{align}
	    \left| \left\| \phi^{(h-1)}(y) \right\|_2^2 - 1 \right| \le h \cdot \frac{\varepsilon^2}{60L^3},~~\text{ and }~~\left| \left\| \phi^{(h-1)}(z) \right\|_2^2 - 1 \right| \le h \cdot \frac{\varepsilon^2}{60L^3}.
	\end{align}
	Therefore, by \cref{lema:sensitivity-polynomial} we have $\left| P_{\tt relu}^{(p)}\left(\|\phi^{(h-1)}(y)\|_2^2\right) - P_{\tt relu}^{(p)}(1) \right| \le h \cdot \frac{\varepsilon^2}{60L^3}$ and $\left| P_{\tt relu}^{(p)}\left(\|\phi^{(h-1)}(z)\|_2^2\right) - P_{\tt relu}^{(p)}(1) \right| \le h \cdot \frac{\varepsilon^2}{60L^3}$. Consequently, since $P_{\tt relu}^{(p)}(1) \le P_{\tt relu}^{(+\infty)}(1) = 1$, we obtain that $B \le \frac{11}{10}.$ By plugging this into \cref{eq:phi-inner-prod-bound} we have,
	\begin{equation} \label{eq:phi-inner-prod-final}
		\Pr\left[ \left| \left< \phi^{(h)}(y), \phi^{(h)}(z)\right> - P^{(p)}_{\tt relu}\left( \left<\phi^{(h-1)}(y), \phi^{(h-1)}(z)\right> \right) \right| \le \bigo\left(\frac{\varepsilon^2}{L^3}\right) \right] \ge 1 - \bigo\left(\frac{\delta}{L}\right).
	\end{equation}
	Furthermore, the inductive hypothesis $P_1(h-1)$ implies that 
	\begin{align}
	    \left| \left< \phi^{(h-1)}(y), \phi^{(h-1)}(z) \right> - \Sigma_{\tt relu}^{(h-1)}\left( \frac{\langle y , z \rangle}{\|y\|_2 \|z\|_2} \right) \right| \le h \cdot \frac{\varepsilon^2}{60L^3}.
	\end{align}
	Hence, using \cref{lema:sensitivity-polynomial} we find that,
	\begin{align}
	    \left|P^{(p)}_{\tt relu}\left( \left<\phi^{(h-1)}(y), \phi^{(h-1)}(z)\right> \right) - P^{(p)}_{\tt relu}\left( \Sigma_{\tt relu}^{(h-1)}\left( \frac{\langle y , z \rangle}{\|y\|_2 \|z\|_2} \right) \right) \right| \le h \cdot \frac{\varepsilon^2}{60L^3}.
    \end{align}
	By incorporating the above inequality into \cref{eq:phi-inner-prod-final} using triangle inequality we find that,
	\small
	\begin{equation} \label{eq:phi-inner-prod-}
		\Pr\left[ \left| \left< \phi^{(h)}(y), \phi^{(h)}(z)\right> - P^{(p)}_{\tt relu}\left(\Sigma_{\tt relu}^{(h-1)}\left( \frac{\langle y , z \rangle}{\|y\|_2 \|z\|_2} \right)\right) \right| \le  \frac{h \cdot\varepsilon^2}{60L^3} +  \bigo\left(\frac{\varepsilon^2}{L^3}\right) \right] \ge 1 - \bigo\left(\frac{\delta}{L}\right).
	\end{equation}
	\normalsize
	Now, by invoking \cref{lem:polynomi-approx-krelu} and using the fact that $p = \left\lceil 2 L^2 /{\varepsilon}^{4/3} \right\rceil$ we have,
	\begin{align}
	    \left| P_{\tt relu}^{(p)}\left(\Sigma_{\tt relu}^{(h-1)}\left( \frac{\langle y , z \rangle}{\|y\|_2 \|z\|_2} \right)\right) - \kappa_1\left(\Sigma_{\tt relu}^{(h-1)}\left( \frac{\langle y , z \rangle}{\|y\|_2 \|z\|_2} \right)\right) \right| \le \frac{\varepsilon^2}{76 L^3}.
	\end{align}
	By combining the above inequality with \cref{eq:phi-inner-prod-} using triangle inequality and using the fact that $\kappa_1\left(\Sigma_{\tt relu}^{(h-1)}\left( \frac{\langle y , z \rangle}{\|y\|_2 \|z\|_2} \right)\right) = \Sigma_{\tt relu}^{(h)}\left( \frac{\langle y , z \rangle}{\|y\|_2 \|z\|_2} \right)$ (by \cref{eq:dp-covar-relu}), we get the following bound,
	\begin{align}
	    \Pr\left[ \left| \left< \phi^{(h)}(y), \phi^{(h)}(z)\right> - \Sigma_{\tt relu}^{(h)}\left( \frac{\langle y , z \rangle}{\|y\|_2 \|z\|_2} \right) \right| \le (h+1) \cdot \frac{\varepsilon^2}{60L^3} \right] \ge 1 - \bigo\left(\frac{\delta}{L}\right).
	\end{align}
	Similarly, we can prove that
	\begin{align*}
	    &\Pr\left[ \left| \left\| \phi^{(h)}(y) \right\|_2^2 - 1 \right| >  \frac{(h+1) \cdot\varepsilon^2}{60L^3} \right] \le \bigo\left( \frac{\delta}{L} \right), \text{ and } \\
	    &\Pr\left[ \left| \left\| \phi^{(h)}(z) \right\|_2^2 - 1 \right| > \frac{(h+1) \cdot\varepsilon^2}{60L^3} \right] \le \bigo\left( \frac{\delta}{L} \right).
	\end{align*}
	This is sufficient to prove the inductive step by union bound, i.e., $\Pr[P_1(h)|P_1(h-1)] \ge 1 - \bigo\left(\frac{\delta}{L}\right)$.
	
	Now we prove the inductive step for statement $P_2(h)$, that is, we prove that conditioned on $P_2(h-1), P_1(h), P_1(h-1)$, statement $P_2(h)$ holds with probability at least $1-\bigo\left(\frac{\delta}{L}\right)$.
	First, note that by \cref{lem:srht} and using \cref{eq:map-derivative-covar} we have,
	\begin{equation}\label{eq:phi-dot-h}
		\Pr\left[ \left| \left< \dot{\phi}^{(h)}(y), \dot{\phi}^{(h)}(z)\right> - \sum_{j=0}^{2p'+1} b_j \left<Y^{(h)}_{j}(y), Y^{(h)}_{j}(z)\right> \right| \le O\left(\frac{\varepsilon}{L}\right) \cdot \widehat{A} \right] \ge 1 - \bigo\left(\frac{\delta}{L}\right),
	\end{equation}
	where $\widehat{A} := \sqrt{\sum_{j=0}^{2p'+1} b_j \|Y^{(h)}_{j}(y)\|_2^2} \cdot \sqrt{\sum_{j=0}^{2p'+1} b_j \|Y^{(h)}_{j}(z)\|_2^2}$ and the collection of vectors $\left\{Y^{(h)}_{j}(y)\right\}_{j=0}^{2p'+1}$ and $\left\{Y^{(h)}_{j}(z)\right\}_{j=0}^{2p'+1}$ and coefficients $\{b_j\}_{j=0}^{2p'+1}$ are defined as per \cref{eq:map-derivative-covar} and \cref{eq:poly-approx-krelu}, respectively. By invoking \cref{soda-result} along with union bound, with probability at least $1 - \bigo\left( \frac{\delta}{L} \right)$, the following inequalities hold true simultaneously for all $j =0,1,\ldots 2p'+1$
	\begin{align}
		& \left|\left<Y^{(h)}_{j}(y), Y^{(h)}_{j}(z)\right> - \left<\phi^{(h-1)}(y), \phi^{(h-1)}(z)\right>^j \right| \le \bigo\left( \frac{\varepsilon}{L} \right) \cdot \left\| \phi^{(h-1)}(y) \right\|_2^j \left\| \phi^{(h-1)}(z)\right\|_2^j \nonumber\\
		& \left\| Y^{(h)}_{j}(y)\right\|_2^2 \le \frac{11}{10} \cdot \left\| \phi^{(h-1)}(y) \right\|_2^{2j} \label{eq:Yj-sketch}\\
		& \left\| Y^{(h)}_{j}(z)\right\|_2^2 \le \frac{11}{10} \cdot \left\| \phi^{(h-1)}(z) \right\|_2^{2j} \nonumber
	\end{align}
	Therefore, by plugging \cref{eq:Yj-sketch} into \cref{eq:phi-dot-h} and using union bound, triangle inequality and Cauchy–Schwarz inequality we find that,
	\small
	\begin{equation} \label{eq:phi-dot-inner-prod-bound}
		\Pr\left[ \left| \left< \dot{\phi}^{(h)}(y), \dot{\phi}^{(h)}(z)\right> - \dot{P}^{(p')}_{\tt relu}\left( \left<\phi^{(h-1)}(y), \phi^{(h-1)}(z)\right> \right) \right| \le \bigo\left(\frac{\varepsilon}{L}\right) \cdot \widehat{B} \right] \ge 1 - \bigo\left(\frac{\delta}{L}\right),
	\end{equation}
	\normalsize
	where $\widehat{B}:= \sqrt{\dot{P}^{(p')}_{\tt relu}\left(\|\phi^{(h-1)}(y)\|_2^2\right) \cdot \dot{P}^{(p')}_{\tt relu}\left(\|\phi^{(h-1)}(z)\|_2^2\right)}$ and $\dot{P}^{(p')}_{\tt relu}(\alpha) = \sum_{j=0}^{2p'+1} b_j \cdot \alpha^j$ is the polynomial defined in \cref{eq:poly-approx-krelu}.
	By inductive hypothesis $P_1(h-1)$ we have $\left| \left\| \phi^{(h-1)}(y) \right\|_2^2 - 1 \right| \le h \cdot \frac{\varepsilon^2}{60L^3}$ and $\left| \left\| \phi^{(h-1)}(z) \right\|_2^2 - 1 \right| \le h \cdot \frac{\varepsilon^2}{60L^3}$. 
	Therefore, using the fact that $p' = \left\lceil 9L^2 /\varepsilon^{2} \right\rceil$ and \cref{lema:sensitivity-polynomial}, $\left| \dot{P}_{\tt relu}^{(p')}\left(\|\phi^{(h-1)}(y)\|_2^2\right) - \dot{P}_{\tt relu}^{(p')}(1) \right| \le  \frac{h \cdot\varepsilon}{20L^2}$ and $\left| \dot{P}_{\tt relu}^{(p')}\left(\|\phi^{(h-1)}(z)\|_2^2\right) - \dot{P}_{\tt relu}^{(p')}(1) \right| \le  \frac{h \cdot\varepsilon}{20L^2}$. Consequently, since $\dot{P}_{\tt relu}^{(p')}(1) \le \dot{P}_{\tt relu}^{(+\infty)}(1) = 1$, we find that $\widehat{B} \le \frac{11}{10}.$
	By plugging this into \cref{eq:phi-dot-inner-prod-bound} we have,
	\begin{equation} \label{eq:phi-dot-inner-prod-final}
		\Pr\left[ \left| \left< \dot{\phi}^{(h)}(y), \dot{\phi}^{(h)}(z)\right> - \dot{P}^{(p')}_{\tt relu}\left( \left<\phi^{(h-1)}(y), \phi^{(h-1)}(z)\right> \right) \right| \le \bigo\left(\frac{\varepsilon}{L}\right) \right] \ge 1 - \bigo\left(\frac{\delta}{L}\right).
	\end{equation}
	
	Furthermore, inductive hypothesis $P_1(h-1)$ implies $\left| \left< \phi^{(h-1)}(y), \phi^{(h-1)}(z) \right> - \Sigma_{\tt relu}^{(h-1)}\left( \frac{\langle y , z \rangle}{\|y\|_2 \|z\|_2} \right) \right| \le h \cdot \frac{\varepsilon^2}{60L^3}$,
	hence, by invoking \cref{lema:sensitivity-polynomial} we find that,
	\begin{align*}
	    \left|\dot{P}^{(p')}_{\tt relu}\left( \left<\phi^{(h-1)}(y), \phi^{(h-1)}(z)\right> \right) - \dot{P}^{(p')}_{\tt relu}\left( \Sigma_{\tt relu}^{(h-1)}\left( \frac{\langle y , z \rangle}{\|y\|_2 \|z\|_2} \right) \right) \right| \le  \frac{h \cdot \varepsilon}{20L^2}.
	\end{align*}
	By plugging the above inequality into \cref{eq:phi-dot-inner-prod-final} using triangle inequality, we find that,
	\small
	\begin{equation} \label{eq:phi-dot-inner-prod-}
		\Pr\left[ \left| \left< \dot{\phi}^{(h)}(y), \dot{\phi}^{(h)}(z)\right> - \dot{P}^{(p')}_{\tt relu}\left(\Sigma_{\tt relu}^{(h-1)}\left( \frac{\langle y , z \rangle}{\|y\|_2 \|z\|_2} \right)\right) \right| \le \frac{h \cdot \varepsilon}{20L^2} +  \bigo\left(\frac{\varepsilon}{L}\right) \right] \ge 1 - \bigo\left(\frac{\delta}{L}\right).
	\end{equation}
	\normalsize
	Now, by invoking \cref{lem:polynomi-approx-krelu} and using the fact that $p' = \left\lceil 9L^2 / {\varepsilon}^2 \right\rceil$ we have,
	\begin{align}
	    \left| \dot{P}_{\tt relu}^{(p')}\left(\Sigma_{\tt relu}^{(h-1)}\left( \frac{\langle y , z \rangle}{\|y\|_2 \|z\|_2} \right)\right) - \kappa_0 \left(\Sigma_{\tt relu}^{(h-1)}\left( \frac{\langle y , z \rangle}{\|y\|_2 \|z\|_2} \right)\right) \right| \le \frac{\varepsilon}{15 L}.
	\end{align}
	By combining the above inequality with \cref{eq:phi-dot-inner-prod-} using triangle inequality and using the fact that $\kappa_0\left(\Sigma_{\tt relu}^{(h-1)}\left( \frac{\langle y , z \rangle}{\|y\|_2 \|z\|_2} \right)\right) = \dot{\Sigma}_{\tt relu}^{(h)}\left( \frac{\langle y , z \rangle}{\|y\|_2 \|z\|_2} \right)$ (by \cref{eq:dp-covar-relu}), we get the following bound,
	\begin{equation}\label{eq:phi-dot-error-bound}
		\Pr\left[ \left| \left< \dot{\phi}^{(h)}(y), \dot{\phi}^{(h)}(z)\right> - \dot{\Sigma}_{\tt relu}^{(h)}\left( \frac{\langle y , z \rangle}{\|y\|_2 \|z\|_2} \right) \right| \le  \frac{\varepsilon}{8L} \right] \ge 1 - \bigo\left(\frac{\delta}{L}\right).
	\end{equation}
	Similarly we can show that,
	\begin{equation}\label{eq:phi-dot-norm-bound}
		\Pr\left[ \left| \left\| \dot{\phi}^{(h)}(y)\right\| - 1 \right| > \frac{\varepsilon}{8L} \right] \le \bigo\left(\frac{\delta}{L}\right), ~\text{ and}~ \Pr\left[ \left| \left\| \dot{\phi}^{(h)}(z)\right\| - 1 \right| >  \frac{\varepsilon}{8L} \right] \le \bigo\left(\frac{\delta}{L}\right).
	\end{equation}

	Now let $f := \psi^{(h-1)}(y) \otimes \dot{\phi}^{(h)}(y)$ and $g := \psi^{(h-1)}(z) \otimes \dot{\phi}^{(h)}(z)$. Then by \cref{lem:srht} and using \cref{eq:map-relu} we have the following,
	\small
	\begin{equation}\label{eq:psi-h-inner-prod}
		\Pr\left[ \left| \left< \psi^{(h)}(y) , \psi^{(h)}(z) \right> - \left< \Q^2  f \oplus \phi^{(h)}(y), \Q^2  g \oplus \phi^{(h)}(z) \right> \right| \le \bigo\left( \frac{\varepsilon}{L} \right) \cdot D \right] \ge 1 - \bigo\left(\frac{\delta}{L}\right), 
	\end{equation}
	\normalsize
	where $D := \left\| \Q^2  f \oplus \phi^{(h)}(y) \right\|_2 \left\| \Q^2  g \oplus \phi^{(h)}(z) \right\|_2$. By the fact that we conditioned on $P_1(h)$, 
	\[ D \le  \sqrt{\left\| \Q^2  f \right\|_2^2 + \frac{11}{10}} \cdot \sqrt{\left\| \Q^2  g \right\|_2^2 + \frac{11}{10}}. \]
	By \cref{soda-result}, we can further obtain an upper bound:
	\[ D \le \frac{11}{10} \cdot \sqrt{\left\| f \right\|_2^2 + 1} \cdot \sqrt{\left\| g \right\|_2^2 + 1}. \]
	Now note that because we conditioned on $P_2(h-1)$ and using \cref{eq:phi-dot-norm-bound}, with probability at least $1 - \bigo\left(\frac{\delta}{L}\right)$ the following holds:
	\[ \left\| f \right\|_2^2 = \left\| \psi^{(h-1)}(y) \right\|_2^2 \left\| \dot{\phi}^{(h)}(y) \right\|_2^2 \le \frac{11}{10} \cdot K^{(h-1)}_{\tt relu}(1) = \frac{11}{10}h. \]
	Similarly, $\left\| g \right\|_2^2 \le \frac{11}{10}h$ with probability at least $1-\bigo\left(\frac{\delta}{L}\right)$, thus, by union bound:
	\[ \Pr[D \le 2 (h+1) | P_2(h-1), P_1(h), P_1(h-1)] \ge 1 - \bigo\left(\frac{\delta}{L}\right). \]
	Therefore, by combining the above with \cref{eq:psi-h-inner-prod} via union bound we find that,
	\small
	\begin{equation}\label{eq:psi-h-inner-prod-bound}
		\Pr\left[ \left| \left< \psi^{(h)}(y) , \psi^{(h)}(z) \right> - \left< \Q^2  f \oplus \phi^{(h)}(y), \Q^2  g \oplus \phi^{(h)}(z) \right> \right| \le \bigo\left(\frac{\varepsilon h}{L}\right) \right] \ge 1 - \bigo\left(\frac{\delta}{L}\right),
	\end{equation}
	\normalsize
	Now note that $\left< \Q^2  f \oplus \phi^{(h)}(y), \Q^2  g \oplus \phi^{(h)}(z) \right> = \left< \Q^2  f, \Q^2  g \right> + \left< \phi^{(h)}(y), \phi^{(h)}(z) \right>$. We proceed by bounding the term $\left| \left< \Q^2  f, \Q^2  g \right> - \left< f, g \right> \right|$ using \cref{soda-result}, as follows,
	\begin{align}
	    \Pr\left[ \left| \left< \Q^2  f, \Q^2  g \right> - \left< f, g \right> \right| \le \bigo\left( \frac{\varepsilon}{L} \right) \cdot \|f\|_2 \|g\|_2 \right] \ge 1 - \bigo\left(\frac{\delta}{L}\right).
	\end{align}
	We proved that conditioned on $P_2(h-1)$ and $P_1(h-1)$, $\left\| f \right\|_2^2 \le 11h/10$ and $\left\| g \right\|_2^2 \le 11h/10$ with probability at least $1-\bigo\left(\frac{\delta}{L}\right)$. Hence, by union bound we find that,
	\begin{equation}\label{eq:first-term-kernel-bound}
		\Pr\left[ \left.\left| \left< \Q^2  f, \Q^2  g \right> - \left< f, g \right> \right| \le \bigo\left( \frac{\varepsilon h}{L} \right) \right| P_2(h-1), P_1(h-1) \right] \ge 1 - \bigo\left(\frac{\delta}{L}\right).
	\end{equation}
	Note that $\left< f, g \right> = \left< \psi^{(h-1)}(y), \psi^{(h-1)}(z) \right> \cdot \left< \dot{\phi}^{(h)}(y), \dot{\phi}^{(h)}(z) \right>$, thus by conditioning on inductive hypothesis $P_2(h-1)$ and \cref{eq:phi-dot-error-bound} we have,
	\[ \begin{split}
		\left| \left< f, g \right> - K_{\tt relu}^{(h-1)}\left( \frac{\langle y , z \rangle}{\|y\|_2 \|z\|_2} \right) \cdot \dot{\Sigma}_{\tt relu}^{(h)}\left( \frac{\langle y , z \rangle}{\|y\|_2 \|z\|_2} \right) \right| &\le \frac{\varepsilon}{8L}  \left(h+\varepsilon \cdot \frac{(h-1)^2+1}{10L} \right) + \varepsilon \cdot \frac{(h-1)^2+1}{10L}
	\end{split}
	\]
	By combining the above inequality with \cref{eq:first-term-kernel-bound}, $P_1(h)$, and \cref{eq:psi-h-inner-prod-bound} using triangle inequality and union bound we get the following inequality,
	\small
	\[ 	\Pr\left[ \left| \left< \psi^{(h)}(y) , \psi^{(h)}(z) \right> - K_{\tt relu}^{(h-1)}\left( \frac{\langle y , z \rangle}{\|y\| \|z\|} \right) \cdot \dot{\Sigma}_{\tt relu}^{(h)}\left( \frac{\langle y , z \rangle}{\|y\| \|z\|} \right) - \Sigma_{\tt relu}^{(h)}\left( \frac{\langle y , z \rangle}{\|y\| \|z\|} \right) \right| > \varepsilon \cdot \frac{h^2+1}{10L} \right] \le O\left(\frac{\delta}{L}\right). \]
	\normalsize
	By noting that $K_{\tt relu}^{(h-1)}\left( \alpha \right) \cdot \dot{\Sigma}_{\tt relu}^{(h)}\left( \alpha \right) + \Sigma_{\tt relu}^{(h)}\left( \alpha \right) = K_{\tt relu}^{(h)}\left( \alpha \right)$ (see \cref{eq:dp-ntk-relu}) we have proved that
	\begin{align}
	    \Pr\left[ \left| \left< \psi^{(h)}(y) , \psi^{(h)}(z) \right> - K_{\tt relu}^{(h)}\left( \frac{\langle y , z \rangle}{\|y\|_2 \|z\|_2} \right) \right| \le \varepsilon \cdot \frac{h^2+1}{10L} \right] \ge 1 - \bigo\left(\frac{\delta}{L}\right).
	\end{align}
	Similarly we can prove the following inequalities hold with probability at least $1 - \bigo\left(\frac{\delta}{L}\right)$,
	\[ \left| \left\| \psi^{(h)}(y) \right\|_2^2 - K_{\tt relu}^{(h)}( 1) \right| \le \varepsilon \cdot \frac{h^2+1}{10L}, \text{ and } \left| \left\| \psi^{(h)}(z) \right\|_2^2 - K_{\tt relu}^{(h)}( 1) \right| \le \varepsilon \cdot \frac{h^2+1}{10L}. \]
	This proves the inductive step for the statement $P_2(h)$ follows, i.e.,
	\[\Pr[ P_2(h) | P_2(h-1), P_1(h), P_1(h-1) ] \ge 1 - \bigo\left(\frac{\delta}{L}\right).\]
	Therefore, by union bounding over all $h=0,1,2, \ldots L$, it follows that the statements of the lemma hold simultaneously for all $h$ with probability at least $1 - \delta$. This completes the proof of \cref{thm:ntk-sketch-corr}.
\qed

We now analyze the runtime of the \NTKS algorithm:
\begin{lemma}[Runtime of {\sc NTKSketch}]
	\label{thm:ntk-sketch-runtime}
	For every positive integers $d$ and $L$, every $\varepsilon, \delta>0$, every vector $x \in \RR^d$, the time to compute \NTKS $\Psi_{\tt ntk}^{(L)}(x) \in \RR^{s^*}$, for $s^*=\bigo\left( \frac{1}{\varepsilon^2} \cdot \log \frac{1}{\delta} \right)$, using the procedure given in \cref{alg-def-ntk-sketch} is bounded by,
	\[ \bigo\left( \frac{L^{11}}{\varepsilon^{6.7}} \cdot \log^3 \frac{L}{\varepsilon\delta} + \frac{L^3}{\varepsilon^2} \cdot \log \frac{L}{\varepsilon\delta} \cdot {\rm nnz}(x) \right). \]
\end{lemma}
{\it Proof of \cref{thm:ntk-sketch-runtime}:}
	There are three main components to the runtime of this procedure that we have to account for. The first is the time to apply the sketch $\Q^1$ to $x$ in line~\ref{eq:map-covar-zero} of \cref{alg-def-ntk-sketch}. By \cref{soda-result}, the runtime of computing $\Q^1\cdot x$ is $\bigo\left( \frac{L^6}{\varepsilon^4} \cdot \log^3 \frac{L}{\varepsilon\delta} + \frac{L^3}{\varepsilon^2} \cdot \log \frac{L}{\varepsilon\delta} \cdot {\rm nnz}(x) \right)$. The second heavy operation corresponds to computing vectors $Z_j^{(h)}(x) = \Q^{2p+2} \cdot \left(\left[ \phi^{(h-1)}(x) \right]^{\otimes j} \otimes {e}_1^{\otimes 2p+2-j}\right)$ for $j=0,1,2, \ldots 2p+2$ and $h=1,2, \ldots L$ in \cref{eq:map-covar}. By \cref{soda-result}, the time to compute $Z_j^{(h)}(x)$ for a fixed $h$ and all $j=0,1,2, \ldots 2p+2$ is bounded by,
	\[ \bigo\left( \frac{L^{10}}{\varepsilon^{20/3}} \cdot \log^2 \frac{L}{\varepsilon} \cdot \log^3 \frac{L}{\varepsilon\delta} + \frac{L^{8}}{\varepsilon^{16/3}} \cdot \log^3 \frac{L}{\varepsilon\delta} \right) = \bigo\left( \frac{L^{10}}{\varepsilon^{6.7}} \cdot \log^3 \frac{L}{\varepsilon\delta} \right). \]
	The total time to compute vectors $Z_j^{(h)}(x)$ for all $h=1,2, \ldots L$ and all $j=0,1,2, \ldots 2p+2$ is thus $\bigo\left( \frac{L^{11}}{\varepsilon^{6.7}} \cdot \log^3 \frac{L}{\varepsilon\delta} \right)$. 
	Finally, the last computationally expensive operation is computing vectors $Y_j^{(h)}(x) = \Q^{2p'+1} \cdot \left(\left[ \phi^{(h-1)}(x) \right]^{\otimes j} \otimes {e}_1^{\otimes 2p'+1-j}\right)$ for $j=0,1,2, \ldots 2p'+1$ and $h=1,2, \ldots L$ in \cref{eq:map-derivative-covar}.
	By \cref{soda-result}, the runtime of computing $Y_j^{(h)}(x)$ for a fixed $h$ and all $j=0,1,2, \ldots 2p'+1$ is bounded by,
	\[ \bigo\left( \frac{L^{6}}{\varepsilon^6} \cdot \log^2 \frac{L}{\varepsilon} \cdot \log^3 \frac{L}{\varepsilon\delta} + \frac{L^{8}}{\varepsilon^6} \cdot \log^3 \frac{L}{\varepsilon\delta} \right) = \bigo\left( \frac{L^{8}}{\varepsilon^6}  \cdot \log^3 \frac{L}{\varepsilon\delta} \right). \]
	Hence, the total time to compute vectors $Y_j^{(h)}(x)$ for all $h=1,2, \ldots L$ and all $j=0,1,2, \ldots 2p'+1$ is $\bigo\left( \frac{L^{9}}{\varepsilon^6} \cdot \log^3 \frac{L}{\varepsilon\delta} \right)$. The total runtime of the NTK Sketch is obtained by summing up these three contributions. This completes the proof of \cref{thm:ntk-sketch-runtime}.
\qed

Now we are ready to prove the main theorem on {\sc NTKSketch}. \mainthmntk*

{\it Proof of \cref{mainthm-ntk}:}
	Let $\psi^{(L)}:\RR^d \to \RR^s$ for $s=\bigo\left( \frac{L^2}{\varepsilon^2} \cdot \log^2 \frac{L}{\varepsilon\delta} \right)$ be the mapping defined in \cref{eq:map-relu} of \cref{alg-def-ntk-sketch}. By \cref{Psi-ntk-def}, the NTK Sketch $\Psi_{\tt ntk}^{(L)}(x)$ is defined as 
	\[\Psi_{\tt ntk}^{(L)}(x):= {\|x\|_2} \cdot \G \cdot \psi^{(L)}( x).\]
	Because $\G$ is a matrix of i.i.d normal entries with $s^{*} = C \cdot \frac{1}{\varepsilon^2} \cdot \log\frac{1}{\delta}$ rows, for a large enough constant $C$, $\G$ is a JL transform~\cite{dasgupta2003elementary} and hence $\Psi_{\tt ntk}^{(L)}$ satisfies the following,
	\[ \Pr \left[ \left| \left< \Psi_{\tt ntk}^{(L)}(y) , \Psi_{\tt ntk}^{(L)}(z) \right> - {\|y\|_2}{\|z\|_2}  \cdot \left< \psi^{(L)}( y),  \psi^{(L)}(z) \right> \right| \le \bigo(\varepsilon) \cdot A \right] \ge 1 - \bigo(\delta), \]
	where $A := {\|y\|_2}{\|z\|_2} \left\| \psi^{(L)}({y}) \right\|_2  \left\| \psi^{(L)}({z}) \right\|_2$. By \cref{thm:ntk-sketch-corr} and using the fact that $K_{\tt relu}^{(L)}(1) = L+1$, the following bounds hold with probability at least $1 - \bigo(\delta)$:
	\[ \left\| \psi^{(L)}( {y}) \right\|_2^2 \le \frac{11}{10} \cdot (L+1), \text{ and } \left\| \psi^{(L)}( {z}) \right\|_2^2 \le \frac{11}{10} \cdot (L+1). \]
	Therefore, by union bound we find that,
	\[ \Pr \left[ \left| \left< \Psi_{\tt ntk}^{(L)}(y) , \Psi_{\tt ntk}^{(L)}(z) \right> - {\|y\|_2}{\|z\|_2}  \cdot \left< \psi^{(L)}({y}),  \psi^{(L)}( {z}) \right> \right| \le \bigo(\varepsilon L)\cdot {\|y\|_2}{\|z\|_2} \right] \ge 1 - \bigo(\delta). \]
	Additionally, by \cref{thm:ntk-sketch-corr}, the following holds with probability at least $1 - \bigo(\delta)$:
	\[ \left| \left< \psi^{(L)}( {y}),  \psi^{(L)}( {z}) \right> - K_{\tt relu}^{(L)}\left( \frac{\langle y, z \rangle }{\|y\|_2\|z\|_2} \right) \right| \le \frac{\varepsilon (L+1)}{10}. \]
	Hence by union bound and triangle inequality we have,
	\[ \Pr \left[ \left| \left< \Psi_{\tt ntk}^{(L)}(y) , \Psi_{\tt ntk}^{(L)}(z) \right> - {\|y\|_2}{\|z\|_2}  \cdot K_{\tt relu}^{(L)}\left( \frac{\langle y, z \rangle }{\|y\|_2\|z\|_2} \right) \right| \le \frac{\varepsilon (L+1)}{9} \cdot {\|y\|_2}{\|z\|_2} \right] \ge 1 - \bigo(\delta). \]
	Now note that by \cref{K_relu_NTK_correspondence}, ${\|y\|_2}{\|z\|_2}  \cdot K_{\tt relu}^{(L)}\left( \frac{\langle y, z \rangle }{\|y\|_2\|z\|_2} \right) = \Theta_{\tt ntk}^{(L)}(y,z)$, and also note that for every $L\ge 2$ and any $\alpha \in [-1,1]$, $K_{\tt relu}^{(L)}\left( \alpha \right) \ge (L+1)/9$, therefore,
	\[ \Pr \left[ \left| \left< \Psi_{\tt ntk}^{(L)}(y) , \Psi_{\tt ntk}^{(L)}(z) \right> - \Theta_{\tt ntk}^{(L)}(y,z) \right| \le \varepsilon \cdot \Theta_{\tt ntk}^{(L)}(y,z) \right] \ge 1 - \delta. \] 
	
	{\bf Remark on the fact that $K_{\tt relu}^{(L)}\left( \alpha \right) \ge (L+1)/9$ for every $L\ge 2$ and any $\alpha \in [-1,1]$.}  Note that from the definition of $\Sigma_{\tt relu}^{(h)}$ in \cref{eq:dp-covar-relu}, we have that for any $\alpha \in [-1,1]$: $\Sigma_{\tt relu}^{(0)}(\alpha) \ge -1$, $\Sigma_{\tt relu}^{(1)}(\alpha) \ge 0$, $\Sigma_{\tt relu}^{(2)}(\alpha) \ge \frac{1}{\pi}$, and $\Sigma_{\tt relu}^{(h)}(\alpha) \ge \frac{1}{2}$ for every $h \ge 3$ because $\kappa_1(\cdot)$ is a monotonically increasing function on the interval $[-1,1]$. Moreover, using the definition of $\dot{\Sigma}_{\tt relu}^{(h)}$ in \cref{eq:dp-covar-relu}, we have that for any $\alpha \in [-1,1]$: $\dot{\Sigma}_{\tt relu}^{(1)}(\alpha) \ge 0$, $\dot{\Sigma}_{\tt relu}^{(2)}(\alpha) \ge \frac{1}{2}$, and $\dot{\Sigma}_{\tt relu}^{(h)}(\alpha) \ge \frac{3}{5}$ for every $h \ge 3$ because $\kappa_0(\cdot)$ is a monotonically increasing function on the interval $[-1,1]$. By an inducive proof and using the definition of $K_{\tt relu}^{(L)}$ in \cref{eq:dp-ntk-relu}, we can show that $K_{\tt relu}^{(L)}\left( \alpha \right) \ge (L+1)/9$ for every $L\ge 2$ and any $\alpha \in [-1,1]$
	
	{\bf Runtime analysis:} By \cref{thm:ntk-sketch-runtime}, runtime to compute the \NTKS is 
	\begin{align}
	    \bigo\left( \frac{L^{11}}{\varepsilon^{6.7}} \log^3 \frac{L}{\varepsilon\delta} + \frac{L^3}{\varepsilon^2} \log \frac{L}{\varepsilon\delta} \cdot {\rm nnz}(x) \right).
	\end{align}
This completes the proof of \cref{mainthm-ntk}.\qed

\section{NTK Random Features: Claims and Proofs}\label{sec:proof-ntk-random-features-error}

\subsection{Proof of \cref{thm:ntk-random-features-error}}
In this section we prove \cref{thm:ntk-random-features-error}. We first restate the theorem:
\randomfeatureserror*

In the proof of this theorem we use the following results from the literature,

\begin{lemma}[Corollary 16 in \citep{daniely2016toward}] \label{lmm:a1-entry-diff}
	Given integer $\ell>0$ and $x, x' \in \RR^{d}$ such that $\norm{y}_2=\norm{z}_2=1$, let $\phi_{\tt rf}^{(\ell)}(y), \phi_{\tt rf}^{(\ell)}(z)$ be defined as per line~\ref{alg_phi1} of \cref{alg:ntk_random_features}. For $\delta_1, \varepsilon_1 \in (0,1)$, there exists a constant $C_1 > 0$ such that for any $m_1 \geq C_1 \frac{L^2}{\varepsilon_1^2} \log\left( \frac{ L}{\delta_1}\right)$ the following holds:
	\begin{align}
		\Pr\left[ \abs{
		  \inner{\phi_{\tt rf}^{(\ell)}(y), \phi_{\tt rf}^{(\ell)}(z)} - \Sigma_{\tt relu}^{(\ell)}(\langle y, z \rangle)
		} \leq \varepsilon_1 \right] \ge 1 - \delta_1.
	\end{align}
\end{lemma}
We also need the following lemma,
\begin{lemma}[Lemma E.5 in \cite{arora2019exact}] \label{lmm:a0-entry-diff}
	Given $x, x' \in \R^d$ with $\norm{x}_2 = \norm{x'}_2 = 1$, integer $\ell >0$, let $\phi_{\tt rf} ^{(\ell)}(x), \phi_{\tt rf} ^{(\ell)}(x')$ be defined as per line~\ref{alg_phi1} of \cref{alg:ntk_random_features}. For any $\varepsilon_2 \in (0,1)$ assume that
	\begin{align}
		\abs{ \inner{ \phi_{\tt rf}^{(\ell)}(x), \phi_{\tt rf}^{(\ell)}(x')} - \Sigma_{\tt relu}^{(\ell)}(\langle x, x' \rangle)} \leq \frac{\varepsilon_2^2}{2}.
	\end{align}
	Then, for $\dot{\phi}_{\tt rf} ^{(\ell)}(x), \dot{\phi}_{\tt rf} ^{(\ell)}(x')$ defined as in line~\ref{alg_phidot1} of \cref{alg:ntk_random_features}, and any $\delta_2>0$ the following holds:
	\begin{align}
		\Pr \left[ \abs{ \inner{\dot{\phi}_{\tt rf}^{(\ell)}(x), \dot{\phi}_{\tt rf}^{(\ell)}(x')} - \dot{\Sigma}_{\tt relu}^{(\ell)}(\langle x, x' \rangle) } \leq \varepsilon_2 + \sqrt{\frac{2}{m_0} \log\left( \frac{6}{\delta_2}\right)} \right] \ge 1-\delta_2.
	\end{align}
\end{lemma}
{\it Proof of \cref{thm:ntk-random-features-error}:}
For fixed $y, z \in \RR^d$ and $\ell = 0, \dots, L$, we denote the estimation error as
\begin{align*}
\Delta_{\ell} := 
\max_{(x, x') \in \{ (y,z), (y,y), (z, z)\}}
\abs{\inner{\psi_{\tt rf}^{(\ell)}(x), \psi_{\tt rf}^{(\ell)}(x')} - K_{\tt relu}^{(\ell)}\left( \frac{\langle x, x' \rangle}{\|x\|_2 \|x'\|_2} \right)}
\end{align*}
and note that $\Delta_0 = 0$. Recall that, by \cref{def:relu-ntk} and \cref{K_relu_NTK_correspondence}:
\begin{align}
    \Theta_{\tt ntk}^{(\ell)}(x,x') = \|x\|_2 \|x'\|_2 \cdot K_{\tt relu}^{(\ell)}\left( \frac{\langle x, x' \rangle}{\|x\|_2 \|x'\|_2} \right),
\end{align}
where for every $\alpha \in [-1,1]$ and $\ell=1,\dots, L$:
\begin{align*}
K_{\tt relu}^{(\ell)}(\alpha) &:= K_{\tt relu}^{(\ell-1)}(\alpha)\cdot \dot{\Sigma}_{\tt relu}^{(\ell)}(\alpha) + \Sigma_{\tt relu}^{(\ell)}(\alpha),\\
\dot{\Sigma}_{\tt relu}^{(\ell)}(\alpha) &:= \kappa_0 \left(\Sigma_{\tt relu}^{(\ell-1)}(\alpha) \right), \\
\Sigma_{\tt relu}^{(\ell)}(\alpha) &:= \kappa_1\left(\Sigma_{\tt relu}^{(\ell-1)}(\alpha) \right).
\end{align*}

We use the recursive relation to approximate:
\begin{align*}
{\inner{ \psi_{\tt rf}^{(\ell)}(x), \psi_{\tt rf}^{(\ell)}(x')}} &= \inner{ \phi_{\tt rf}^{(\ell)}(x), \phi_{\tt rf}^{(\ell)}(x')}\\ &\qquad + \inner{ \Q^2 \cdot \left( \dot{\phi}_{\tt rf}^{(\ell)}(x) \otimes {\psi}_{\tt rf}^{(\ell-1)}(x) \right), \Q^2 \cdot \left( \dot{\phi}_{\tt rf}^{(\ell)}(x') \otimes {\psi}_{\tt rf}^{(\ell-1)}(x') \right)}\\
&\approx \inner{ \phi_{\tt rf}^{(\ell)}(x), \phi_{\tt rf}^{(\ell)}(x')} + 
\inner{ \dot{\phi}_{\tt rf}^{(\ell)}(x) \otimes {\psi}_{\tt rf}^{(\ell-1)}(x), 
 \dot{\phi}_{\tt rf}^{(\ell)}(x') \otimes {\psi}_{\tt rf}^{(\ell-1)}(x')
} \\
&\approx \left( \Sigma_{\tt relu}^{(\ell)} +  \dot{\Sigma}_{\tt relu}^{(\ell)} \cdot  K_{\tt relu}^{(\ell-1)} \right)\left( \frac{\langle x, x' \rangle}{\|x\|_2 \|x'\|_2} \right) = K_{\tt relu}^{(\ell)}\left( \frac{\langle x, x' \rangle}{\|x\|_2 \|x'\|_2} \right).
\end{align*}

For notational simplicity, we define the following events:
\small
\begin{align}
\mathcal{E}_{\phi}^{(\ell)} (\varepsilon) &:= 
	\left\{
		\abs{ \inner{ \phi_{\tt rf}^{(\ell)}(x), \phi_{\tt rf}^{(\ell)}(x')} - \Sigma_{\tt relu}^{(\ell)}\left( \frac{\langle x, x' \rangle}{\|x\|_2 \|x'\|_2} \right) } \leq \varepsilon	: \forall (x, x') \in \{ (y,z), (y,y), (z, z)\}
	\right\}, \\
\dot{\mathcal{E}}_{\phi}^{(\ell)} (\varepsilon) &:= 
    \left\{
        \abs{ \inner{ \dot{\phi}_{\tt rf}^{(\ell)}(x), \dot{\phi}_{\tt rf}^{(\ell)}(x')} - \dot{\Sigma}_{\tt relu}^{(\ell)}\left( \frac{\langle x, x' \rangle}{\|x\|_2 \|x'\|_2} \right) } \leq \varepsilon	: \forall (x, x') \in \{ (y,z), (y,y), (z, z)\}
    \right\}.
\end{align}
\normalsize
Our proof is based on the following claims:

First we claim that, there exists a constant $C_1 > 0$ such that for any $m_1 \geq C_1 \frac{L^6}{\varepsilon^4} \log\left( \frac{ L}{\delta}\right)$:
\begin{equation}\label{claim1}
	\Pr\left[
	\mathcal{E}_{\phi}^{(\ell)} \left( \frac{\varepsilon^2}{100L^2} \right)
	\right] \geq 1 - \frac{\delta}{3L}.
\end{equation}
which directly follows by invoking \cref{lmm:a1-entry-diff} with $(x, x') \in \left\{ \left(\frac{y}{\|y\|_2},\frac{z}{\|z\|_2}\right), \left(\frac{y}{\|y\|_2},\frac{y}{\|y\|_2}\right), \left(\frac{z}{\|z\|_2}, \frac{z}{\|z\|_2}\right) \right\}$ and setting $\varepsilon_1 = \frac{\varepsilon^2}{100L^2}$ and $\delta_1 = \frac{\delta}{9L}$ and applying union bound over choices of $(x, x')$. 

Our second claim is that there exists a constant $C_0 > 0$ such that if $m_0 \geq C_0 \frac{L^2}{\varepsilon^2}  \log\left( \frac{L}{\delta} \right)$ then
\begin{equation}\label{claim2}
	\Pr \left[ \left.
	\dot{\mathcal{E}}_{\phi}^{(\ell)} \left( \frac{\varepsilon}{8L} \right)
	\right|
	\mathcal{E}_{\phi}^{(\ell)} \left( \frac{\varepsilon^2}{100L^2} \right)
	\right] \geq 1 - \frac{\delta}{3L}.
\end{equation}
The above statement is a direct consequence of invoking \cref{lmm:a0-entry-diff} with $(x, x') \in \left\{ \left(\frac{y}{\|y\|_2},\frac{z}{\|z\|_2}\right), \left(\frac{y}{\|y\|_2},\frac{y}{\|y\|_2}\right), \left(\frac{z}{\|z\|_2}, \frac{z}{\|z\|_2}\right) \right\}$ and union bounding over choices of $(x, x')$.
In \cref{lmm:a0-entry-diff}, we choose $\varepsilon_2 = \frac{\varepsilon}{10L}, \delta_2 = \frac{\delta}{9L}$ and $m_0 \geq \frac{3200L^2}{\varepsilon^2} \log \left(\frac{54L}{\delta} \right)$ for $\varepsilon, \delta \in(0,1)$ to obtain \cref{claim2} by union bound.

Our next claim is the following,
\begin{restatable}{claim}{claimthree} \label{claim3}
Let $f(x):= \dot{\phi}_{\tt rf}^{(\ell)}(x) \otimes {\psi}_{\tt rf}^{(\ell-1)}(x)$  for every $x \in \RR^d$. There exists a constant $C_2 > 0$ such that if $m_{s}\geq C_2 \frac{L^2}{\varepsilon^2} \log\frac{L}{\varepsilon \delta}$ then conditioned on the events $\dot{\mathcal{E}}_{\phi}^{(\ell)} \left( \frac{\varepsilon}{8L} \right)$ and $	\mathcal{E}_{\phi}^{(\ell)} \left( \frac{\varepsilon^2}{100L^2} \right)$ the following holds with probability at least $1 - \frac{\delta}{3L}$, for all $(x, x') \in \left\{ \left(\frac{y}{\|y\|_2},\frac{z}{\|z\|_2}\right), \left(\frac{y}{\|y\|_2},\frac{y}{\|y\|_2}\right), \left(\frac{z}{\|z\|_2}, \frac{z}{\|z\|_2}\right) \right\}$,
\begin{align*}
\left| \inner{ \Q^2 \cdot f(x) , \Q^2 \cdot f(x')} - \inner{f(x), f(x')} \right| \le \frac{\varepsilon}{100L}(\ell + \Delta_{\ell-1}).
\end{align*}
\end{restatable}
The proof of the above claim is provided in \cref{sec:proof-claim1}.
Now, by combining \cref{claim1}, \cref{claim2} and \cref{claim3}, we can prove the following claim which provides a recursive relation for bounding $\Delta_\ell$.
\begin{restatable}{claim}{claimfour} \label{claim4}
If the events in \cref{claim1}, \cref{claim2} and \cref{claim3} hold, then 
\begin{align} \label{eq:delta-recur}
	\Delta_\ell \leq \left(1 + \frac{\varepsilon}{7L}\right) \Delta_{\ell-1} + \frac{\varepsilon}{7L} \ell.
\end{align}
\end{restatable}
The proof of \cref{claim4} is provided in \cref{sec:proof-claim1}.

Note that by union bound, with probability $1 - \frac{\delta}{L}$, all preconditions of \cref{claim4} hold and thus \cref{eq:delta-recur} holds with probability $1 - \frac{\delta}{L}$.
Applying union bound on \cref{eq:delta-recur} for all $\ell \in [L]$ and solving the recurrence, we obtain that with probability at least $1 - \delta$, the following bound holds
\begin{align}
\Delta_\ell \leq \frac{\varepsilon}{9L} \cdot \ell^2.
\end{align}

When $L\ge 2$, we showed in the proof of \cref{mainthm-ntk} that $K_{\tt relu}^{(L)}(\cdot) \ge \frac{L+1}{9}$, therefore,
\begin{align}
\abs{\inner{\psi_{\tt rf}^{(L)}(x), \psi_{\tt rf}^{(L)}(x')} - K_{\tt relu}^{(L)}\left( \frac{\langle x, x' \rangle}{\|x\|_2 \|x'\|_2} \right)} \leq \varepsilon \cdot K_{\tt relu}^{(L)}\left( \frac{\langle x, x' \rangle}{\|x\|_2 \|x'\|_2} \right).
\end{align}

Since ${\Psi}_{\tt rf}^{(L)}(x) = \|x\|_2 \cdot {\psi}^{(L)}(x)$ and ${\Psi}_{\tt rf}^{(L)}(x') = \|x'\|_2 \cdot {\psi}^{(L)}(x')$, this implies that,
\[ \Pr\left[ \abs{\inner{\Psi_{\tt rf}^{(L)}(x), \Psi_{\tt rf}^{(L)}(x')} - \Theta_{\tt ntk}^{(L)}(  x, x' )} \leq \varepsilon \cdot \Theta_{\tt ntk}^{(L)}(  x, x' ) \right] \ge 1 - \delta.\]
This completes the proof of \cref{thm:ntk-random-features-error}.
\qed

\subsection{Proof of Auxiliary Claims} \label{sec:proof-claim1}

\claimthree*
{\it Proof of \cref{claim3}:}
The proof is based on \cref{soda-result} that provides an upper bound on variance of the {\sc PolySketch}.
By using the definition of $f(x), f(x')$ and \cref{soda-result}, with probability at least $1- \frac{\delta}{9L}$, we have
\begin{align}
	&\abs{\inner{ \Q^2 \cdot f(x) , \Q^2 \cdot f(x')} - \inner{f(x), f(x')}} \nonumber\\
	&\qquad\leq \frac{\varepsilon}{200L} \norm{\dot{\phi}_{\tt rf}^{(\ell)}(x)}_2 \norm{\dot{\phi}_{\tt rf}^{(\ell)}(x')}_2 \norm{\psi_{\tt rf}^{(\ell-1)}(x)}_2 \norm{\psi_{\tt rf}^{(\ell-1)}(x')}_2 \nonumber\\
	&\qquad\leq \frac{\varepsilon}{200L} \left( 1 + \frac{\varepsilon}{8L} \right) \norm{\psi_{\tt rf}^{(\ell-1)}(x)}_2 \norm{\psi_{\tt rf}^{(\ell-1)}(x')}_2 \nonumber\\
	&\qquad \leq \frac{\varepsilon}{100L} \left( \ell + \Delta_{\ell-1} \right) \nonumber
\end{align}
where the second inequality follows from the assumption that $\dot{\mathcal{E}}_{\phi}^{(\ell)} \left( \frac{\varepsilon}{8L} \right)$ holds and the third one follows from the fact that 
$K_{\tt relu}^{(\ell-1)}(\alpha) \leq \ell$ for any $\alpha \in [-1,1]$ and
\begin{align} \label{eq:norm-psi}
	\norm{\psi_{\tt rf}^{(\ell-1)}(x)}_2^2 \leq K_{\tt relu}^{(\ell-1)}(x,x') + \Delta_{\ell-1} \leq \ell + \Delta_{\ell-1}.
\end{align}
Union bounding over the choices of $(x,x')$ completes the proof of \cref{claim3}.
\qed

\claimfour*
{\it Proof of \cref{claim4}:}
Recall that 
\begin{align*}
    \Delta_{\ell} := 
\max_{(x, x') \in \{ (y,z), (y,y), (z, z)\}}
\abs{\inner{\psi_{\tt rf}^{(\ell)}(x), \psi_{\tt rf}^{(\ell)}(x')} - K_{\tt relu}^{(\ell)}\left( \frac{\langle x, x' \rangle}{\|x\|_2 \|x'\|_2} \right)}.
\end{align*}
Observe that the estimation error $\Delta_{\ell}$ can be decomposed into three parts:
\small
\begin{align} \label{eq:three-part}
&\abs{\inner{\psi_{\tt rf}^{(\ell)}(x), \psi_{\tt rf}^{(\ell)}(x')} - K_{\tt relu}^{(\ell)}\left( \frac{\langle x, x' \rangle}{\|x\|_2 \|x'\|_2} \right)}
 \leq \abs{\inner{\phi_{\tt rf}^{(\ell)}(x), \phi_{\tt rf}^{(\ell)}(x')} - \Sigma_{\tt relu}^{(\ell)}\left( \frac{\langle x, x' \rangle}{\|x\|_2 \|x'\|_2} \right)} \nonumber \\
&\qquad \qquad + \abs{\inner{ \Q^2 \cdot f(x) , \Q^2 \cdot f(x')} - \inner{f(x), f(x')}} \\
&\qquad \qquad + \abs{\inner{\dot{\phi}_{\tt rf}^{(\ell)}(x) \otimes {\psi}_{\tt rf}^{(\ell-1)}(x), \dot{\phi}_{\tt rf}^{(\ell)}(x') \otimes {\psi}_{\tt rf}^{(\ell-1)}(x')} - \dot{\Sigma}_{\tt relu}^{(\ell)}\left( \frac{\langle x, x' \rangle}{\|x\|_2 \|x'\|_2} \right) K_{\tt relu}^{(\ell-1)}\left( \frac{\langle x, x' \rangle}{\|x\|_2 \|x'\|_2} \right)}. \nonumber
\end{align} 
\normalsize
for $(x, x') \in \{ (y,z), (y,y), (z, z)\}$. By the assumption that \cref{claim1} holds, the definition of $\mathcal{E}_{\phi}^{(\ell)} \left( \frac{\varepsilon^2}{100L^2} \right)$ implies that,
\begin{align} \label{eq:part1}
	\abs{ \inner{\phi_{\tt rf}^{(\ell)}(x),\phi_{\tt rf}^{(\ell)}(x')} - \Sigma_{\tt relu}^{(\ell)}\left( \frac{\langle x, x' \rangle}{\|x\|_2 \|x'\|_2} \right) }  \leq  \frac{\varepsilon^2}{100L^2}
\end{align}

Furthermore, the assumption that \cref{claim3} holds, implies that,
\begin{align} \label{eq:part2}
\left| \inner{ \Q^2 \cdot f(x) , \Q^2 \cdot f(x')} - \inner{f(x), f(x')} \right| \le \frac{\varepsilon}{100L}(\ell + \Delta_{\ell-1}).
\end{align}

For the third part in \cref{eq:three-part}, we observe that 
\begin{align}
	&\abs{\inner{\dot{\phi}_{\tt rf}^{(\ell)}(x) \otimes {\psi}_{\tt rf}^{(\ell-1)}(x), \dot{\phi}_{\tt rf}^{(\ell)}(x') \otimes {\psi}_{\tt rf}^{(\ell-1)}(x')} - \dot{\Sigma}_{\tt relu}^{(\ell)}\left( \frac{\langle x, x' \rangle}{\|x\|_2 \|x'\|_2} \right) K_{\tt relu}^{(\ell-1)}\left( \frac{\langle x, x' \rangle}{\|x\|_2 \|x'\|_2} \right)} \nonumber\\
	&\leq 
	\abs{\inner{\dot{\phi}_{\tt rf}^{(\ell)}(x), \dot{\phi}_{\tt rf}^{(\ell)}(x')}} \cdot
	\abs{\inner{{\psi}_{\tt rf}^{(\ell-1)}(x), {\psi}_{\tt rf}^{(\ell-1)}(x')} - K_{\tt relu}^{(\ell-1)}\left( \frac{\langle x, x' \rangle}{\|x\|_2 \|x'\|_2} \right)} \nonumber \\
	&\quad \qquad + K_{\tt relu}^{(\ell-1)}\left( \frac{\langle x, x' \rangle}{\|x\|_2 \|x'\|_2} \right) \cdot \abs{\inner{\dot{\phi}_{\tt rf}^{(\ell)}(x), \dot{\phi}_{\tt rf}^{(\ell)}(x')} - \dot{\Sigma}_{\tt relu}^{(\ell)}\left( \frac{\langle x, x' \rangle}{\|x\|_2 \|x'\|_2} \right)} \nonumber \\
	&\leq \left( 1+ \frac{\varepsilon}{8L} \right)\cdot \Delta_{\ell-1} + \ell \cdot \frac{\varepsilon}{8L}, \label{eq:part3}
\end{align}
where the second inequality comes from that the assumption that \cref{claim2} holds along with $\abs{\dot{\Sigma}_{\tt relu}^{(\ell)}(\cdot)} \leq 1$ and $\abs{K_{\tt relu}^{(\ell-1)}(\cdot)} \leq \ell$. 
Putting \cref{eq:part1}, \cref{eq:part2} and \cref{eq:part3} into \cref{eq:three-part}, we have 
\begin{align}
	\Delta_{\ell} \leq \frac{\varepsilon^2}{100L^2} + \frac{\varepsilon}{100L} \left( \ell + \Delta_{\ell-1} \right) + \left( 1+ \frac{\varepsilon}{8L} \right) \Delta_{\ell-1} + \frac{\ell \cdot \varepsilon}{8L}
	= 
	\left(1 + \frac{\varepsilon}{7L}\right) \Delta_{\ell-1} + \frac{\varepsilon}{7L} \ell.
\end{align}
This completes the proof of \cref{claim4}.
\qed


\section{Spectral Approximation via Leverage Scores Sampling}
\subsection{Zeroth Order Arc-Cosine Kernels} \label{sec:proof-a0-spectral}

The proofs here rely on Theorem 3.3 in \cite{lee2020generalized} which states spectral approximation bounds of random features for general kernels equipped with the leverage score sampling. This result is a generalization of~\cite{avron2017random} on the Random Fourier Features.

\begin{theorem}[Theorem 3.3 in \citep{lee2020generalized}] \label{thm:generalized_leverage_score}
Suppose $\K\in \R^{n \times n}$ is a kernel matrix with statistical dimension $s_{\lambda}$ for some $\lambda\in (0, \norm{\K}_2)$. Let $\BPhi(\w) \in \R^n$ be a feature map with a random vector $\w \sim p(\w)$ satisfying that
$\K = \E_{\w}\left[ \BPhi(\w) \BPhi(\w)^\top \right]$.
Define $\tau_{\lambda}(\w) := p(\w) \cdot \BPhi(\w)^\top (\K + \lambda \I)^{-1} \BPhi(\w)$. Let $\widetilde{\tau}(\w)$ be any measurable function such that $\widetilde{\tau}(\w) \geq \tau_{\lambda}(\w)$ for all $\w$. Assume that $s_{\widetilde{\tau}} := \int \widetilde{\tau}(\w) d \w$ is finite.
Consider random vectors $\w_1, \dots, \w_m$ sampled from $q(\w):= \widetilde{\tau}(\w) / s_{\widetilde{\tau}}$ and define that
\begin{align}
\overline{\boldsymbol \Phi} := \frac1{\sqrt{m}} \left[ \sqrt{\frac{p(\w_1)}{q(\w_1)}} \BPhi(\w_1), \ \dots \ , \sqrt{\frac{p(\w_m)}{q(\w_m)}} \BPhi(\w_m) \right]^\top.
\end{align}
If $m \geq \frac{8}{3} \varepsilon^{-2} s_{\widetilde{\tau}} \log \left( 16 s_{\lambda} / \delta\right)$ then  
\begin{align}
\left(1 - \varepsilon \right) \left( \K + \lambda \I\right)
\preceq
\overline{\boldsymbol \Phi}^\top \overline{\boldsymbol \Phi} + \lambda \I
\preceq
\left(1 + \varepsilon \right) \left( \K + \lambda \I\right)
\end{align}
holds with probability at least $1-\delta$.
\end{theorem}

We now ready to provide spectral approximation guarantee for arc-cosine kernels of order zero.

\begin{theorem} \label{thm:a0_spectral}
Given dataset $\X \in \R^{d \times n}$, let $\K_0 \in \R^{n \times n}$ be the arc-cosine kernel matrix of $0$-th order with $\X$ and denote $\BPhi_0 := \sqrt{\frac{2}{m}} \mathrm{Step}(\W \X) \in \R^{m \times n}$ where each entry in $\W\in \R^{m \times d}$ is an i.i.d. sample from $\mathcal{N}(0,1)$. 
For $\lambda \in (0, \norm{\K_0}_2)$, let $s_{\lambda}$ be the statistical dimension of $\K_0$.  Given $\varepsilon \in (0, 1/2)$ and $\delta \in (0,1)$, if $m\geq \frac{8}{3} \frac{n}{\lambda \varepsilon^{2} } \log\left( \frac{16 s_{\lambda}}{\delta} \right)$, then
it holds that
\begin{align*}
(1 - \varepsilon) (\K_0 + \lambda \I) 
\preceq
\BPhi_0^\top \BPhi_0 + \lambda \I 
\preceq 
(1 + \varepsilon) (\K_0 + \lambda \I)
\end{align*}
with probability at least $1-\delta$.
\end{theorem}

{\it Proof of \cref{thm:a0_spectral}:}
Let $\BPhi_0(w) := \sqrt{2} \ \mathrm{Step}(\X^\top w) \in \R^n$ for $w \in \R^d$ and $p(w)$ be the 
probability density function of the standard normal distribution. As studied in~\cite{cho2009kernel}, $\BPhi_0(w)$ is a random feature of $\K_0$ such that
\begin{align}
\K_0 = \E_{w \sim p(w)} \left[ \BPhi_0(w) \BPhi_0(w)^\top\right].
\end{align}
In order to utilize \cref{thm:generalized_leverage_score}, we need an upper bound of $\tau_{\lambda}(w)$ as below:
\begin{align}
\tau_{\lambda}(w) &:= p(w) \cdot \BPhi_0(w)^\top \left( \K_0 + \lambda \I \right)^{-1} \BPhi_0(w) \\
&\leq p(w) \norm{(\K_0 + \lambda \I)^{-1}}_{2} \norm{\BPhi_0(w)}_2^2 \\
&\leq p(w) \frac{\norm{\BPhi_0(w)}_2^2 }{\lambda} \\
&\leq p(w) \frac{2n}{\lambda}
\end{align}
where the inequality in second line holds from the definition of matrix operator norm and the inequality in third line follows from the fact that smallest eigenvalue of $\K_0 + \lambda \I$ is equal to or greater than $\lambda$. The last inequality is from that $\norm{\mathrm{Step}(x)}_2^2 \leq n$ for any $x \in \R^n$. Note that $\int_{\R^d} p(w) \frac{2n}{\lambda} dw= \frac{2n}{\lambda}$ and since it is constant the modified random features correspond to the original ones. Putting all together into \cref{thm:generalized_leverage_score}, we can obtain the result.
This completes the proof of \cref{thm:a0_spectral}. \qed

\subsection{First Order Arc-Cosine Kernels} \label{sec:proof-a1-spectral}
\begin{theorem} \label{thm:a1_spectral}
Given dataset $\X \in \R^{d \times n}$, let $\K_1 \in \R^{n \times n}$ be the arc-cosine kernel matrix of $1$-th order with $\X$ and $v_1, \dots, v_m \in \R^d$ be i.i.d. random vectors from probability distribution 
$q(v) = \frac{1}{(2\pi)^{d/2} d} \norm{v}_2^2 \exp\left(-\frac{1}{2}\norm{v}_2^2\right).$
Denote
$
\BPhi_1 := \sqrt{\frac{2d}{m}} \left[
\frac{\mathrm{ReLU}( \X^\top v_1)}{\norm{v_1}_2} , \ \dots \ , \frac{\mathrm{ReLU}( \X^\top v_m )}{\norm{v_m}_2} \right]^\top
$
and for $\lambda \in (0, \norm{\K_1}_2)$, let $s_{\lambda}$ be the statistical dimension of $\K_1$. Given $\varepsilon \in (0, 1/2)$ and $\delta \in (0,1)$, if $m\geq \frac{8}{3} \frac{d}{\varepsilon^2} \min\left\{\rank(\X)^2, \frac{ \norm{\X}_2^2}{\lambda} \right\}\log\left( \frac{16 s_{\lambda}}{\delta} \right)$, then it holds that
\begin{align*}
(1 - \varepsilon) (\K_1 + \lambda I) 
\preceq
\BPhi_1^\top \BPhi_1 + \lambda \I 
\preceq 
(1 + \varepsilon) (\K_1 + \lambda I)
\end{align*}
with probability at least $1-\delta$.
\end{theorem}

{\it Proof of \cref{thm:a1_spectral}:}
Let $\BPhi_1(w) := \sqrt{2}~\mathrm{ReLU}( \X^\top w) \in \R^n$ for $w \in \R^d$ and $p(w)$ be the probability density function of standard normal distribution.
\citet{cho2009kernel} also showed that $\BPhi_1(w)$ is a random feature of $\K_1$ such that
\begin{align}
\K_1 = \E_{w \sim p(w)} \left[ \BPhi_1(w) \BPhi_1(w)^\top\right].
\end{align}
Then, 
\begin{align*}
\tau_{\lambda}(w) &:= p(w) \cdot \BPhi_1(w)^\top (\K_1 + \lambda \I)^{-1} \BPhi_1(w) \\
&\leq p(w) \norm{(\K_1 + \lambda \I)^{-1}}_2 \norm{\BPhi_1(w)}_2^2 \\
&= 2~p(w) \frac{\norm{\mathrm{ReLU}(\X^\top w)}_2^2}{\lambda} \\
&\leq 2~p(w) \frac{\norm{\X^\top w}_2^2}{\lambda} \\
&\leq 2~p(w) \norm{w}_2^2 \frac{\norm{\X}_2^2}{\lambda}
\end{align*}
where the inequality in fourth line holds from the fact that $\norm{\mathrm{ReLU}(x)}_2^2 \leq \norm{x}_2^2$ for any vector $x$. 

On the other hand, if we write the ReLU function in terms of a matrix form, i.e., $\BPhi_1(w) = \sqrt{2}~\mathrm{ReLU}(\X^\top w) = \sqrt{2}~\S \X^\top w$ where $\S$ is a diagonal matrix such that $\S_{ii}=1$ if $[\X^\top w]_i > 0$ else $\S_{ii}=0$ for $i \in [n]$, then we can obtain that 
\begin{align}
\tau_{\lambda}(w) &:= p(w) \cdot \BPhi_1(w)^\top (\K_1 + \lambda \I)^{-1} \BPhi_1(w) \nonumber\\
&= 2~p(w)\cdot w^\top \X \S (\K_1 + \lambda \I)^{-1} \S \X^\top w. \label{ls-def-k1}
\end{align}
Now note that by definition of $\K_1$, we have $[\K_1]_{i,j} = \|x_i\|_2 \|x_j\|_2 \cdot \kappa_1\left( \frac{\langle x_i, x_j \rangle}{\|x_i\|_2 \|x_j\|_2} \right)$. Therefore, using the Taylor expansion of the function $\kappa_1(\alpha) = \frac{1}{\pi} + \frac{\alpha}{2} + \frac{1}{\pi} \cdot \sum_{i=0}^\infty \frac{(2i)!}{2^{2i} \cdot (i!)^2 \cdot (2i+1)\cdot (2i+2)} \cdot \alpha^{2i+2}$ and the fact that its Taylor coefficients are all non-negative, we have,
\[ \frac{1}{2} \X^\top \X \preceq \K_1. \]
Using the above inequality along with \cref{ls-def-k1}, we can write,
\begin{align}
\tau_{\lambda}(w)
&\leq 2~p(w)\cdot w^\top \X \S \left(\frac12 \X^\top \X + \lambda \I\right)^{-1} \S \X^\top w \nonumber\\
&\leq 2~p(w)\cdot \norm{w}_2^2 \cdot \norm{\X \S \left(\frac12 \X^\top \X + \lambda \I\right)^{-1} \S \X^\top }_2 \label{eq:spectral-norm-XSX}
\end{align}
To obtain an upper bound of the third term in \cref{eq:spectral-norm-XSX}, we consider the singular value decomposition of $\X = \V \BSigma \U^\top$. And we have
\begin{align}
\norm{\X \S \left(\frac12 \X^\top \X + \lambda \I\right)^{-1} \S \X^\top }_2 
&= \norm{ \V \BSigma \U^\top \S \left(\frac12 \U\BSigma^2\U^\top + \lambda \I\right)^{-1} \S \U \BSigma \V^\top}_2 \nonumber\\
&= \norm{ \BSigma \U^\top \S \left(\frac12 \U\BSigma^2\U^\top + \lambda \I\right)^{-1} \S \U \BSigma}_2 \nonumber\\
&= \norm{ \BSigma \U^\top \S \U \left(\frac12 \BSigma^2 + \lambda \I\right)^{-1} \U^\top \S \U \BSigma}_2\label{k1-norm-ls-bound}
\end{align}
Now we observe that
\begin{align}
    [\U^\top \S \U]_{ij} \leq \norm{u_i}_2 \norm{u_j}_2
\end{align}
which leads us to $\norm{\U^\top \S \U}_F \leq \rank(\U)=\rank(\X)$. Since $\U^\top \S \U$ is a positive semi-definite matrix we have,
\[ 0 \preceq \U^\top \S \U \preceq \rank(\X) \cdot \I. \]
Therefore, plugging this into \cref{k1-norm-ls-bound} and \cref{eq:spectral-norm-XSX} gives,
\[ \tau_{\lambda}(w) \le 2~p(w)\cdot \norm{w}_2^2 \cdot \rank(\X)^2. \]

Denote $\widetilde{\tau}(w) := 2~p(w) \norm{w}_2^2 \min\left\{ \rank(\X)^2,  \frac{\norm{\X}_2^2}{\lambda} \right\}$ and it holds that
\begin{align}
\int_{\R^d} \widetilde{\tau}(w) dw 
= 2 d \min\left\{ \rank(\X)^2,  \frac{\norm{\X}_2^2}{\lambda} \right\}
\end{align}
since $\int_{\R^d}  p(w) \norm{w}_2^2 = \mathtt{tr}(\I_d) = d$ for $w \sim \mathcal{N}(\boldsymbol{0}, \I_d)$.
We define the modified distribution as
\begin{align} \label{eq:pdf_weighted_normal}
q(w):= \frac{\widetilde{\tau}(w)}{\int_{\R^d} \widetilde{\tau}(w) dw} = p(w) \frac{\norm{w}_2^2}{d} = 
\frac{1}{(2\pi)^{d/2} d} \norm{w}_2^2 \exp\left(-\frac{1}{2}\norm{w}_2^2\right)
\end{align}
and recall that the modified random features are defined as
\begin{align}
\BPhi_1 &= \frac1{\sqrt{m}}
\left[
\sqrt{\frac{p(w_1)}{q(w_1)}} \ \BPhi_1(w_1), \ \dots \ , \sqrt{\frac{p(w_m)}{q(w_m)}} \ \BPhi_1(w_m)
\right]^\top \\
&= \sqrt{\frac{2d}{m}}
\left[
\frac{\mathrm{ReLU}(\X^\top w_1)}{\norm{w_1}_2} , \ \dots \ ,  \frac{\mathrm{ReLU}(\X^\top w_m)}{\norm{w_m}_2}
\right]^\top.
\end{align}
Putting all together into \cref{thm:generalized_leverage_score}, we derive the result. 
This completes the proof of \cref{thm:a1_spectral}. \qed

{\bf Approximate sampling.}
It is not trivial to sample a vector $w \in \R^d$ from the distribution $q(\cdot)$ defined in \cref{eq:pdf_weighted_normal}. Thus, we suggest to perform an approximate sampling via Gibbs sampling. The algorithm starts with a random initialized vector $w$ and then iteratively replaces $[w]_i$ with a sample 
\begin{align}
    q([w]_i | [w]_{\setminus \{i\}}) \propto \frac{\norm{w}_2^2}{1 + \norm{w}_2^2 - [w]_i^2} \exp\left( -\frac{[w]_i^2}{2}\right)
\end{align}
for $i\in [d]$ and repeats this process for $T$ iterations. Sampling from $q([w]_i | [w]_{\setminus \{i\}})$ can be done via the inverse transformation method.\footnote{It requires the CDF of $q([w]_i | [w]_{\setminus \{i\}})$ which is equivalent to $\frac{\mathrm{erf}\left( {[w]_i }/{\sqrt{2}}\right)+1}{2} - \frac{ [w]_i  \exp\left( -[w]_i^2/2\right)}{\sqrt{2 \pi}(1 + \norm{w}_2^2 - [w]_i^2)}$.} We empirically verify that $T=1$ is enough for promising performances. The running time of Gibbs sampling becomes $\bigo(m_1 d  T)$ where $m_1$ corresponds to the number of independent samples from $q(v)$. This is negligible compared to the feature map construction with \PolyS for $T=\bigo(1)$. The pseudo-code for the modified random features of $A_1$ using Gibbs sampling is outlined in \cref{alg:gibbs}.

\begin{algorithm}[t]
\caption{
Gibbs Sampling for Approximating \cref{eq:pdf_weighted_normal} via Inverse Transformation Method
} \label{alg:gibbs}
\begin{algorithmic}[1]
\STATE {\bf Input}: $\X \in \R^{d \times n}$, feature dimension $m$, Gibbs iterations $T$
\STATE Draw i.i.d. $w_i \sim \mathcal{N}({\bf 0}, \I_d)$ for $i \in [m]$
\FOR{ $i = 1$ to $m$}
    \STATE $q(x, z) \leftarrow $ inverse of $\frac{\mathrm{erf}\left( {x}/{\sqrt{2}}\right)+1}{2} - \frac{ x \exp\left( -x^2/2\right)}{\sqrt{2 \pi}(z+1)}$ 
    \hfill(i.e., CDF of $\Pr([w_i]_j | [w_i]_{\setminus \{j\}})$)
    \FOR{ $t = 1$ to $T$}
        \FOR{ $j = 1$ to $d$}
            \STATE $u \leftarrow$ sample from $[0,1]$ at uniformly random
            \STATE $[w_i]_j \leftarrow q\left(u, \sum_{k \in [d]\setminus\{j\}} [w_i]_k^2\right)$
        \ENDFOR
    \ENDFOR
\ENDFOR
\STATE {\bf return } $\sqrt{\frac{2d}{m}} \left[
\frac{\mathrm{ReLU}(\X^\top w_1)}{\norm{w_1}_2} , \ \dots \ , \frac{\mathrm{ReLU}(\X^\top w_m)}{\norm{w_m}_2} \right]$
\end{algorithmic}
\end{algorithm}

\subsection{Proof of \cref{thm:ntk_spectral}} \label{sec:proof_ntk_spectral}

Our proof relies on spectral approximation bounds of \PolyS given in the fourth part of \cref{soda-result}.

\ntkspectral*

{\it Proof of \cref{thm:ntk_spectral}:}
Note that the NTK of two-layer ReLU network can be formulated as
\begin{align}
\K_{\tt ntk} = \K_1 + \K_0 \odot(\X^\top \X)
\end{align}
where $\K_0$ and $\K_1$ are the arc-cosine kernel matrices of order $0$ and $1$ with dataset $\X$, respectively. 

Let $\BPhi_0$ and $\BPhi_1$ be the random features of $\K_0$ and $\K_1$, defined as per \cref{thm:a0_spectral} and \cref{thm:a1_spectral}, respectively. Also let $\BPsi_{\tt rf}$ be the feature matrix that \cref{alg:ntk_random_features} outputs, that is each column of this matrix is obtained by applying this algorithm on the dataset $\X$. By basic properties of tensor products we have,
\begin{align}
\BPsi_{\tt rf} := \BPhi_1 \oplus \Q^2 \cdot \left( \BPhi_0 \otimes \X \right).
\end{align}
Our proof is a combination of spectral analysis of $\BPhi_0^\top \BPhi_0, \BPhi_1^\top \BPhi_1$ and $\left(\Q^2\left( \BPhi_0 \otimes \X \right) \right)^\top \cdot \Q^2\left( \BPhi_0 \otimes \X \right)$ which are stated in \cref{thm:a0_spectral}, \cref{thm:a1_spectral} and  \cref{soda-result}, respectively.

From \cref{thm:a1_spectral}, if $m_1 \geq \frac{16}{3} \frac{d}{\varepsilon^2} \min\left\{\rank(\X)^2, \frac{ \norm{\X}_2^2}{\lambda} \right\} \log\left( \frac{48 s_{\lambda}}{\delta} \right)$ then with probability at least $1 - \frac{\delta}{3}$ the following holds,
\begin{align} \label{eq:a1_spectral}
(1 - \varepsilon) \left(\K_1 + \frac{\lambda}{2} \I\right) 
\preceq
\BPhi_1^\top \BPhi_1 + \frac{\lambda}{2} \I 
\preceq 
(1 + \varepsilon) \left( \K_1 + \frac{\lambda}{2} \I\right).
\end{align}

From \cref{thm:a0_spectral}, if $m_0 \geq 48\frac{n}{\lambda \varepsilon^{2} } \log\left( \frac{48 s_{\lambda}}{\delta} \right)$ then with probability at least $1 - \frac{\delta}{3}$ it holds that,
\begin{align} \label{eq:a0_spectral}
\left(1 - \frac{\varepsilon}{3}\right) \left(\K_0 + \frac{\lambda}{2} \I\right)
\preceq
\BPhi_0^\top  \BPhi_0 + \frac{\lambda}{2} \I 
\preceq 
\left(1 + \frac{\varepsilon}{3}\right) \left(\K_0 + \frac{\lambda}{2} \I\right)
\end{align}

Rearranging \cref{eq:a0_spectral}, we get 
\begin{align*} 
\BPhi_0^\top \BPhi_0 \preceq  \left(1 + \frac{\varepsilon}{3}\right) \K_0 + \frac{\varepsilon}{6} \lambda \I.
\end{align*}
Now we bound the trace of $\left(\BPhi_0 \otimes \X \right)^\top \cdot \left( \BPhi_0 \otimes \X \right) = \BPhi_0^\top \BPhi_0 \odot \X^\top \X$:
\begin{align*}
    \tr \left(\BPhi_0^\top \BPhi_0 \odot \X^\top \X\right) &= \sum_{j \in [n]} [\BPhi_0^\top \BPhi_0]_{j,j} \cdot [\X^\top X]_{j,j}\\
    &\le \sum_{j \in [n]} [\BPhi_0^\top \BPhi_0]_{j,j}\\
    &\le n.
\end{align*}
Now note that, we can write,
\begin{align}\label{stat-dim-bound}
s_{\lambda} \left( \left(\BPhi_0 \otimes \X \right)^\top  (\BPhi_0 \otimes \X ) \right) \le \frac{\tr \left(\BPhi_0^\top \BPhi_0 \odot \X^\top \X\right)}{\tr \left(\BPhi_0^\top \BPhi_0 \odot \X^\top \X\right)/n + \lambda} \le \frac{n}{1+\lambda}.
\end{align}

To guarantee spectral approximation of $\left(\Q^2\left( \BPhi_0 \otimes \X \right) \right)^\top \cdot \Q^2\left( \BPhi_0 \otimes \X \right)$, we will use the result of \cref{soda-result}.
Using \cref{stat-dim-bound} along with \cref{soda-result} and the fact that
$m_{s}\geq \frac{C}{ \varepsilon^2} \cdot \frac{n}{1 + \lambda} \log^3 \frac{n}{\varepsilon\delta}$ for some constant $C$, and union bound, with probability at least $1 - \frac{\delta}{2}$, we have
\begin{align} 
\left(\Q^2\left( \BPhi_0 \otimes \X \right) \right)^\top  \Q^2\left( \BPhi_0 \otimes \X \right) + \frac{\lambda}{2} \I
&\preceq
\left(1 + \frac{\varepsilon}{3} \right) \left( \BPhi_0^\top \BPhi_0 \odot \X^\top \X  + \frac{\lambda}{2} \I \right)\nonumber\\
&\preceq
\left(1 + \frac{\varepsilon}{3}\right) \left( \left[\left(1 + \frac{\varepsilon}{3}\right) \K_0 + \frac{\varepsilon}{6} \lambda \I\right] \odot \X^\top \X  + \frac{\lambda}{2} \I \right) \nonumber\\
&= \left(1 + \frac{\varepsilon}{3}\right)  \left(\left(1 + \frac{\varepsilon}{3}\right) \left(\K_0 \odot \X^\top \X\right) + \frac{\varepsilon}{6} \lambda (\I \odot \X^\top \X ) + \frac{\lambda}{2} \I \right) \nonumber\\
&\preceq  \left(1 + \frac{\varepsilon}{3}\right) \left(1 + \frac{\varepsilon}{3}\right) \left(\K_0 \odot \X^\top \X +  \frac{\lambda}{2} \I  \right) \nonumber\\
&\preceq  \left(1 + \varepsilon \right) \left(\K_0 \odot \X^\top \X +  \frac{\lambda}{2} \I  \right) \label{eq:cs_a0_spectral}
\end{align}
where the inequality in second line follows from \cref{lmm:sdim_upperbound}
and the fourth line follows from the assumption $\norm{\X_{(:,i)}}_2 \le 1$ for all $i \in [n]$ which leads that $\I \odot (\X^\top \X) \preceq \I$. The last inequality holds since $\varepsilon \in (0, 1/2)$.

Similarly, we can obtain the following lower bound:
\begin{align}
\left(\Q^2\left( \BPhi_0 \otimes \X \right) \right)^\top  \Q^2\left( \BPhi_0 \otimes \X \right) + \frac{\lambda}{2} \I
\succeq \left(1 - \varepsilon\right) \left(\K_0 \odot \X^\top \X + \frac{\lambda}{2} \I \right) \label{eq:cs_a0_spectral2}.
\end{align}

Combining \cref{eq:a1_spectral}, \cref{eq:cs_a0_spectral} and \cref{eq:cs_a0_spectral2} gives
\begin{align} \label{eq:ntk_spectral}
(1 - \varepsilon) \left( \K_{\tt ntk} + \lambda \I\right)
\preceq
\BPsi_{\tt rf}^\top \BPsi_{\tt rf} + \lambda \I 
\preceq
(1 + \varepsilon) \left( \K_{\tt ntk} + \lambda \I\right).
\end{align}

Furthermore, by taking a union bound over all events, 
\cref{eq:ntk_spectral} holds with probability at least $1- \delta$.
This completes the proof of \cref{thm:ntk_spectral}. \qed

\subsection{Auxiliary Lemmas}

\begin{lemma} \label{lmm:sdim_upperbound}
If $\A, \B, \C$ are positive semi-definite matrices of conforming sizes such that $\B \preceq \C$, then,
\[ \A \odot \B \preceq \A \odot \C. \] 
\end{lemma}

{\it Proof of \cref{lmm:sdim_upperbound}:}
We want to show that for any vector $v$, $v^\top \A \odot \B v \preceq v^\top \A \odot \C v$. Because the matrices $\A,\B,\C$ are PSD, there exist matrices $\X, \Y, \Z$ of appropriate sizes such that we can decompose these matrices as follows,
\[ \A = \X^\top \X, ~~~~ \B = \Y^\top \Y, ~~~~ \C = \Z^\top \Z. \]
Using this and basic properties of tensor products, we have the following for any vector $v$,
\begin{align*}
    v^\top \A \odot \B v &= \| \X \otimes \Y v \|_2^2\\
    &= \| \X \cdot \text{\tt diag}(v) \cdot \Y^\top \|_F^2\\
    &= 
    \sum_{i} (\X_{(i,:)} \odot v)^\top \cdot \B \cdot (\X_{(i,:)} \odot v)\\
    &\le \sum_{i} (\X_{(i,:)} \odot v)^\top \cdot \C \cdot (\X_{(i,:)} \odot v)\\
    &= v^\top \A \odot \C v.
\end{align*}
This completes the proof of \cref{lmm:sdim_upperbound}.
\qed

\section{ReLU-CNTK: Expression and Main Properties}\label{app-cntk-expression}
In this section we prove that the depth-$L$ CNTK corresponding to ReLU activation is highly structured and can be fully characterized in terms of tensoring and composition of arc-cosine kernel functions $\kappa_1(\cdot)$ and $\kappa_0(\cdot)$. We refer to this kernel function as {\bf ReLU-CNTK}. 
First we start by restating the DP proposed by \citet{arora2019exact} for computing the $L$-layer CNTK kernel corresponding to an arbitrary activation function $\sigma:\RR \to \RR$ and convolutional filters of size $q \times q$, with GAP:
\begin{enumerate}[wide, labelwidth=!, labelindent=0pt]
	\item Let $y,z \in \RR^{d_1\times d_2 \times c}$ be two input images, where $c$ is the number of channels ($c=3$ for RGB images). 
	Define $\Gamma^{(0)}: \RR^{d_1\times d_2 \times c} \times \RR^{d_1 \times d_2 \times c} \to \RR^{d_1 \times d_2 \times d_1 \times d_2}$ and $\Sigma^{(0)}: \RR^{d_1\times d_2 \times c} \times \RR^{d_1 \times d_2 \times c} \to \RR^{d_1 \times d_2 \times d_1 \times d_2}$ as follows for every $i,i' \in [d_1]$ and $j,j' \in [d_2]$:
	\begin{equation}\label{eq:dp-cntk-zero}
	\begin{split}
		&\Gamma^{(0)}(y,z)  := \sum_{l=1}^c y_{(:,:,l)} \otimes z_{(:,:,l)}, \\ &\Sigma^{(0)}_{i,j,i',j'}(y,z) := \sum_{a=-\frac{q-1}{2}}^{\frac{q-1}{2}} \sum_{b=-\frac{q-1}{2}}^{\frac{q-1}{2}}  \Gamma^{(0)}_{i+a,j+b,i'+a,j'+b}(y,z).
	\end{split}
	\end{equation}
	\item For every layer $h = 1,2, \ldots , L$ of the network and every $i,i' \in [d_1]$ and $j,j' \in [d_2]$, define $\Gamma^{(h)}: \RR^{d_1\times d_2 \times c} \times \RR^{d_1 \times d_2 \times c} \to \RR^{d_1 \times d_2 \times d_1 \times d_2}$ recursively as:
	\begin{equation}\label{eq:dp-cntk-covar}
		\begin{split}
			&\Lambda^{(h)}_{i,j,i',j'}(y,z) := \begin{pmatrix}
				\Sigma^{(h-1)}_{i,j,i,j}(y,y) & \Sigma^{(h-1)}_{i,j,i',j'}(y,z)\\
				&\\
				\Sigma^{(h-1)}_{i',j',i,j}(z,y) & \Sigma^{(h-1)}_{i',j',i',j'}(z,z)
			\end{pmatrix},\\
			&\Gamma^{(h)}_{i,j,i',j'}(y,z) := \frac{1}{q^2 \cdot \EE_{w\sim \mathcal{N}(0,1)} \left[ |\sigma(w)|^2 \right]} \cdot \EE_{(u,v) \sim \mathcal{N}\left( 0, \Lambda^{(h)}_{i,j,i',j'}(y,z) \right)} \left[ \sigma(u) \cdot \sigma(v) \right],\\
			&\Sigma^{(h)}_{i,j,i',j'}(y,z) := \sum_{a=-\frac{q-1}{2}}^{\frac{q-1}{2}} \sum_{b=-\frac{q-1}{2}}^{\frac{q-1}{2}}  \Gamma^{(h)}_{i+a,j+b,i'+a,j'+b}(y,z),
		\end{split}
	\end{equation}
	\item For every $h = 1,2, \ldots L $, every $i,i' \in [d_1]$ and $j,j' \in [d_2]$, define $\dot{\Gamma}^{(h)}(y,z) \in \RR^{d_1 \times d_2 \times d_1 \times d_2}$ as:
	\begin{equation}\label{eq:dp-cntk-derivative-covar}
		\dot{\Gamma}^{(h)}_{i,j,i',j'}(y,z) := \frac{1}{q^2 \cdot \EE_{w\sim \mathcal{N}(0,1)} \left[ |\sigma(w)|^2 \right]} \cdot \EE_{(u,v) \sim \mathcal{N}\left( 0, \Lambda^{(h)}_{i,j,i',j'}(y,z) \right)} \left[ \dot{\sigma}(u) \cdot \dot{\sigma}(v) \right].
	\end{equation}
	\item Let $\Pi^{(0)}(y,z) := 0$ and for every $h = 1,2, \ldots, L-1$, every $i,i' \in [d_1]$ and $j,j' \in [d_2]$, define $\Pi^{(h)}: \RR^{d_1\times d_2 \times c} \times \RR^{d_1\times d_2 \times c} \to \RR^{d_1\times d_2\times d_1 \times d_2}$ recursively as:
	\small
	\begin{equation}\label{eq:dp-cntk-pi}
		\begin{split}
			\Pi^{(h)}_{i,j,i',j'}(y,z) &:= \sum_{a=-\frac{q-1}{2}}^{\frac{q-1}{2}} \sum_{b=-\frac{q-1}{2}}^{\frac{q-1}{2}}  \left[\Pi^{(h-1)}(y,z) \odot \dot{\Gamma}^{(h)}(y,z) + \Gamma^{(h)}(y,z)\right]_{i+a,j+b,i'+a,j'+b},\\
			\Pi^{(L)}(y,z) &:= \Pi^{(L-1)}(y,z) \odot \dot{\Gamma}^{(L)}(y,z).
		\end{split}
	\end{equation}
	\normalsize
	\item The final CNTK expressions is defined as:
	\begin{equation}\label{eq:dp-cntk-gap-def}
		\Theta_{\tt cntk}^{(L)}(y,z) := \frac{1}{d_1^2d_2^2} \cdot \sum_{i , i' \in [d_1]} \sum_{j , j' \in [d_2]} \Pi_{i,j,i',j'}^{(L)}(y,z).
	\end{equation}
\end{enumerate}

Now we show how to recursively compute the ReLU-CNTK as follows,
\begin{defn}[ReLU-CNTK] \label{relu-cntk-def}
	For every positive integers $q,L$, the $L$-layer CNTK for ReLU activation function and convolutional filter size of $q \times q$ is defined as follows
	\begin{enumerate}[wide, labelwidth=!, labelindent=0pt]
		\item For $x\in \RR^{d_1\times d_2 \times c}$, every $i \in [d_1]$ and $j \in [d_2]$ let $N_{i,j}^{(0)} (x) :=q^2 \cdot \sum_{l=1}^c \left| x_{i+a,j+b,l} \right|^2$, and for every $h \ge 1$, recursively define,
		\begin{equation}\label{eq:dp-cntk-norm-simplified}
			N^{(h)}_{i,j}(x):=\frac{1}{q^2} \cdot \sum_{a=-\frac{q-1}{2}}^{\frac{q-1}{2}} \sum_{b=-\frac{q-1}{2}}^{\frac{q-1}{2}} N^{(h-1)}_{i+a,j+b}(x).
		\end{equation}
		\item Define $\Gamma^{(0)}(y,z)  := \sum_{l=1}^c y_{(:,:,l)} \otimes z_{(:,:,l)}$. Let $\kappa_1:[-1,1]\to \RR$ be the function defined in \cref{relu-activ-cov} of \cref{def:relu-ntk}. For every layer $h = 1,2, \ldots , L$, every $i,i' \in [d_1]$ and $j,j' \in [d_2]$, define $\Gamma^{(h)}: \RR^{d_1\times d_2 \times c} \times \RR^{d_1 \times d_2 \times c} \to \RR^{d_1 \times d_2 \times d_1 \times d_2}$ recursively as:
		\small
		\begin{equation}\label{eq:dp-cntk-covar-simplified}
			\Gamma^{(h)}_{i,j,i',j'}(y,z) := \frac{\sqrt{N^{(h)}_{i,j}(y) \cdot N^{(h)}_{i',j'}(z)}}{q^2} \cdot \kappa_1\left( \frac{\sum_{a=-\frac{q-1}{2}}^{\frac{q-1}{2}} \sum_{b=-\frac{q-1}{2}}^{\frac{q-1}{2}}  \Gamma^{(h-1)}_{i+a,j+b,i'+a,j'+b}(y,z)}{\sqrt{N^{(h)}_{i,j}(y) \cdot N^{(h)}_{i',j'}(z)}} \right).
		\end{equation}
		\normalsize
		\item Let $\kappa_0:[-1,1]\to \RR$ be the function defined in \cref{relu-activ-cov} of \cref{def:relu-ntk}. For every $h = 1,2, \ldots L$, every $i,i' \in [d_1]$ and $j,j' \in [d_2]$, define $\dot{\Gamma}^{(h)}(y,z) \in \RR^{d_1 \times d_2 \times d_1 \times d_2}$ as:
		\begin{equation}\label{eq:dp-cntk-derivative-covar-simplified}
			\dot{\Gamma}^{(h)}_{i,j,i',j'}(y,z) := \frac{1}{q^2} \cdot \kappa_0\left( \frac{\sum_{a=-\frac{q-1}{2}}^{\frac{q-1}{2}} \sum_{b=-\frac{q-1}{2}}^{\frac{q-1}{2}}  \Gamma^{(h-1)}_{i+a,j+b,i'+a,j'+b}(y,z)}{\sqrt{N^{(h)}_{i,j}(y) \cdot N^{(h)}_{i',j'}(z)}} \right).
		\end{equation}

		\item Let $\Pi^{(0)}(y,z) := 0$ and for every $h = 1,2, \ldots, L-1$, every $i,i' \in [d_1]$ and $j,j' \in [d_2]$, define $\Pi^{(h)}: \RR^{d_1\times d_2 \times c} \times \RR^{d_1\times d_2 \times c} \to \RR^{d_1\times d_2\times d_1 \times d_2}$ recursively as:
		\small
		\begin{equation}\label{eq:dp-cntk}
			\Pi^{(h)}_{i,j,i',j'}(y,z) := \sum_{a=-\frac{q-1}{2}}^{\frac{q-1}{2}} \sum_{b=-\frac{q-1}{2}}^{\frac{q-1}{2}}  \left[\Pi^{(h-1)}(y,z) \odot \dot{\Gamma}^{(h)}(y,z) + \Gamma^{(h)}(y,z) \right]_{i+a,j+b,i'+a,j'+b}.
		\end{equation}
		\normalsize
		Furthermore, for $h=L$ define: 
		\begin{equation}\label{eq:dp-cntk-last-layer}
			\Pi^{(L)}(y,z) := \Pi^{(L-1)}(y,z) \odot \dot{\Gamma}^{(L)}(y,z).
		\end{equation}
		\item The final CNTK expressions for ReLU activation is:
		\begin{equation}\label{eq:dp-cntk-finalkernel}
			\Theta_{\tt cntk}^{(L)}(y,z) := \frac{1}{d_1^2d_2^2} \cdot \sum_{i , i' \in [d_1]} \sum_{j , j' \in [d_2]} \Pi_{i,j,i',j'}^{(L)}(y,z).
		\end{equation}
	\end{enumerate}
\end{defn}

In what follows we prove that the procedure in \cref{relu-cntk-def} precisely computes the CNTK kernel function corresponding to ReLU activation and additionally, we present useful corollaries and consequences of this fact.

\begin{lemma}\label{thm:cntk-relu}
	For every positive integers $d_1,d_2, c$, odd integer $q$, and every integer $h \ge 0$, if the activation function is ReLU, then the tensor covariances $\Gamma^{(h)} , \dot{\Gamma}^{(h)}(y,z): \RR^{d_1\times d_2 \times c} \times \RR^{d_1 \times d_2 \times c} \to \RR^{d_1 \times d_2 \times d_1 \times d_2}$ defined in \cref{eq:dp-cntk-covar} and \cref{eq:dp-cntk-derivative-covar}, are precisely equal to the tensor covariances defined in \cref{eq:dp-cntk-covar-simplified} and \cref{eq:dp-cntk-derivative-covar-simplified} of \cref{relu-cntk-def}, respectively.
\end{lemma}
{\it Proof of \cref{thm:cntk-relu}:}
	To prove the lemma, we first show by induction on $h=1,2, \ldots$ that $N^{(h)}_{i,j}(x) \equiv \Sigma^{(h-1)}_{i,j,i,j}(x,x)$ for every $x\in \RR^{d_1\times d_2 \times c}$ and every $i \in [d_1]$ and $j \in [d_2]$, where $\Sigma^{(h-1)}(x,x)$ is defined as per \cref{eq:dp-cntk-zero} and \cref{eq:dp-cntk-covar}. The {\bf base of induction} trivially holds for $h=1$ because by definition of $N^{(1)}(x)$ and \cref{eq:dp-cntk-zero} we have,
	\[ N^{(1)}_{i,j}(x) = \sum_{a=-\frac{q-1}{2}}^{\frac{q-1}{2}} \sum_{b=-\frac{q-1}{2}}^{\frac{q-1}{2}} \sum_{l=1}^c \left| x_{i+a,j+b,l} \right|^2 \equiv \Sigma^{(0)}_{i,j,i,j}(x,x). \]
	To prove the {\bf inductive step}, suppose that the inductive hypothesis $N^{(h-1)}_{i,j}(x) = \Sigma^{(h-2)}_{i,j,i,j}(x,x)$ holds for some $h\ge2$. Now we show that conditioned on the inductive hypothesis, the inductive claim holds. By \cref{eq:dp-cntk-covar}, we have,
	\begin{align*}
		\Sigma^{(h-1)}_{i,j,i,j}(x,x) &= \sum_{a=-\frac{q-1}{2}}^{\frac{q-1}{2}} \sum_{b=-\frac{q-1}{2}}^{\frac{q-1}{2}}  \Gamma^{(h-1)}_{i+a,j+b,i+a,j+b}(x,x)\\
		&= \sum_{a=-\frac{q-1}{2}}^{\frac{q-1}{2}} \sum_{b=-\frac{q-1}{2}}^{\frac{q-1}{2}} \frac{\EE_{(u,v) \sim \mathcal{N}\left( 0, \Lambda^{(h-1)}_{i+a,j+b,i+a,j+b}(x,x) \right)} \left[ \sigma(u) \cdot \sigma(v) \right]}{q^2 \cdot \EE_{w\sim \mathcal{N}(0,1)} \left[ |\sigma(w)|^2 \right]}\\
		&= \sum_{a=-\frac{q-1}{2}}^{\frac{q-1}{2}} \sum_{b=-\frac{q-1}{2}}^{\frac{q-1}{2}} \frac{\EE_{u \sim \mathcal{N}\left( 0, \Sigma_{i+a,j+b,i+a,j+b}^{(h-2)}(x,x) \right)} \left[ |\max(0,u)|^2 \right]}{q^2 \cdot \EE_{w\sim \mathcal{N}(0,1)} \left[ |\max(0,w)|^2 \right]}\\
		&= \sum_{a=-\frac{q-1}{2}}^{\frac{q-1}{2}} \sum_{b=-\frac{q-1}{2}}^{\frac{q-1}{2}} \frac{1}{q^2} \cdot \Sigma^{(h-2)}_{i+a,j+b,i+a,j+b}(x,x)\\
		&= \sum_{a=-\frac{q-1}{2}}^{\frac{q-1}{2}} \sum_{b=-\frac{q-1}{2}}^{\frac{q-1}{2}} \frac{1}{q^2} \cdot N^{(h-1)}_{i+a,j+b}(x) \equiv N^{(h)}_{i,j}(x). \text{~~~~~~~~~~~~~~~~~~~~~~~~~~~~~~~~~~~~(by \cref{eq:dp-cntk-norm-simplified})}
	\end{align*}
	Therefore, this proves that $N^{(h)}_{i,j}(x) \equiv \Sigma^{(h-1)}_{i,j,i,j}(x,x)$ for every $x$ and every integer $h\ge 1$.
	
	Now, note that the $2 \times 2$ covariance matrix $\Lambda^{(h)}_{i,j,i',j'}(y,z)$, defined in \cref{eq:dp-cntk-covar}, can be decomposed as $\Lambda^{(h)}_{i,j,i',j'}(y,z) = \begin{pmatrix} f^\top \\ g^\top \end{pmatrix} \cdot \begin{pmatrix}
		f & g \end{pmatrix}$, where $f,g \in \RR^2$.
	Also note that $\|f\|_2^2 = \Sigma^{(h-1)}_{i,j,i,j}(y,y)$ and $\|g\|_2^2 = \Sigma^{(h-1)}_{i',j',i',j'}(z,z)$, hence, by what we proved above, we have,
	\[ \|f\|_2^2 = N^{(h)}_{i,j}(y), \text{ and } \|g\|_2^2 = N^{(h)}_{i',j'}(z). \]
	Therefore, by \cref{relu-covariance}, we can write:
	\begin{align*}
		\Gamma^{(h)}_{i,j,i',j'}(y,z) &= \frac{1}{q^2 \cdot \EE_{w\sim \mathcal{N}(0,1)} \left[ |\sigma(w)|^2 \right]} \cdot \EE_{(u,v) \sim \mathcal{N}\left( 0, \Lambda^{(h)}_{i,j,i',j'}(y,z) \right)} \left[ \sigma(u) \cdot \sigma(v) \right]\\
		&= \frac{1}{q^2 \cdot \EE_{w\sim \mathcal{N}(0,1)} \left[ |\sigma(w)|^2 \right]} \cdot \EE_{u \sim \mathcal{N}\left( 0, I_d \right)} \left[ \sigma(u^\top f) \cdot \sigma(u^\top g) \right]\\
		&= \frac{2 \cdot \|f\|_2 \cdot \|g\|_2}{q^2 \cdot \kappa_1(1)} \cdot \frac{1}{2} \cdot \kappa_1\left(\frac{\langle f, g \rangle}{\|f\|_2 \cdot \|g\|_2}\right)\\
		&= \frac{\sqrt{N^{(h)}_{i,j}(y) \cdot N^{(h)}_{i',j'}(z)}}{q^2} \cdot \kappa_1\left(\frac{\Sigma^{(h-1)}_{i,j,i',j'}(y,z)}{\sqrt{N^{(h)}_{i,j}(y) \cdot N^{(h)}_{i',j'}(z)}}\right)\\
		&= \frac{\sqrt{N^{(h)}_{i,j}(y) \cdot N^{(h)}_{i',j'}(z)}}{q^2} \cdot \kappa_1\left( \frac{\sum_{a=-\frac{q-1}{2}}^{\frac{q-1}{2}} \sum_{b=-\frac{q-1}{2}}^{\frac{q-1}{2}}  \Gamma^{(h-1)}_{i+a,j+b,i'+a,j'+b}(y,z)}{\sqrt{N^{(h)}_{i,j}(y) \cdot N^{(h)}_{i',j'}(z)}} \right),
	\end{align*}
	where the third line follows from \cref{relu-covariance} and fourth line follows because we have $\langle f, g \rangle = \Sigma^{(h-1)}_{i,j,i',j'}(y,z)$. The fifth line above follows from \cref{eq:dp-cntk-covar}. This proves the equivalence between the tensor covariance defined in \cref{eq:dp-cntk-covar} and the one defined in \cref{eq:dp-cntk-covar-simplified} of \cref{relu-cntk-def}. Similarly, by using \cref{relu-covariance}, we can prove the statement of the lemma about $\dot{\Gamma}^{(h)}_{i,j,i',j'}(y,z)$ as follows,
	\begin{align*}
		\dot{\Gamma}^{(h)}_{i,j,i',j'}(y,z) &= \frac{1}{q^2 \cdot \EE_{w\sim \mathcal{N}(0,1)} \left[ |\sigma(w)|^2 \right]} \cdot \EE_{(u,v) \sim \mathcal{N}\left( 0, \Lambda^{(h)}_{i,j,i',j'}(y,z) \right)} \left[ \dot{\sigma}(u) \cdot \dot{\sigma}(v) \right]\\
		&= \frac{1}{q^2 \cdot \EE_{w\sim \mathcal{N}(0,1)} \left[ |\sigma(w)|^2 \right]} \cdot \EE_{u \sim \mathcal{N}\left( 0, I_d \right)} \left[ \dot{\sigma}(u^\top f) \cdot \dot{\sigma}(u^\top g) \right]\\
		&= \frac{2}{q^2 \cdot \kappa_1(1)} \cdot \frac{1}{2} \cdot \kappa_0\left(\frac{\langle f, g \rangle}{\|f\|_2 \cdot \|g\|_2}\right)\\
		&= \frac{1}{q^2} \cdot \kappa_0\left(\frac{\Sigma^{(h-1)}_{i,j,i',j'}(y,z)}{\sqrt{N^{(h)}_{i,j}(y) \cdot N^{(h)}_{i',j'}(z)}}\right)\\
		&= \frac{1}{q^2} \cdot \kappa_0\left( \frac{\sum_{a=-\frac{q-1}{2}}^{\frac{q-1}{2}} \sum_{b=-\frac{q-1}{2}}^{\frac{q-1}{2}}  \Gamma^{(h-1)}_{i+a,j+b,i'+a,j'+b}(y,z)}{\sqrt{N^{(h)}_{i,j}(y) \cdot N^{(h)}_{i',j'}(z)}} \right).
	\end{align*}
This completes the proof of \cref{thm:cntk-relu}.
\qed

\begin{corr}[Consequence of \cref{thm:cntk-relu}]\label{N-to-Sigma}
	Consider the preconditions of \cref{thm:cntk-relu}. For every $x \in \RR^{d_1\times d_2 \times c}$, $N^{(h)}_{i,j}(x) \equiv \sum_{a=-\frac{q-1}{2}}^{\frac{q-1}{2}} \sum_{b=-\frac{q-1}{2}}^{\frac{q-1}{2}}  \Gamma^{(h-1)}_{i+a,j+b,i+a,j+b}(x,x)$.
\end{corr}
We describe some of the basic properties of the function $\Gamma^{(h)}(y,z)$ defined in \cref{eq:dp-cntk-covar-simplified} in the following lemma,
\begin{lemma}[Properties of $\Gamma^{(h)}(y,z)$]\label{properties-gamma}
	For every images $y,z \in \RR^{d_1 \times d_2 \times c}$, every integer $h \ge 0$ and every  $i,i' \in [d_1]$ and $j,j' \in [d_2]$ the following properties are satisfied by functions $\Gamma^{(h)}$ and $N^{(h)}$ defined in \cref{eq:dp-cntk-covar-simplified} and \cref{eq:dp-cntk-norm-simplified} of \cref{relu-cntk-def}:
	\begin{enumerate}
		\item {\bf Cauchy–Schwarz inequality:} $\left| \Gamma_{i,j,i',j'}^{(h)}(y,z) \right| \le \frac{\sqrt{N_{i,j}^{(h)}(y) \cdot N_{i',j'}^{(h)}(z)}}{q^2}$.
		\item {\bf Norm value:} $\Gamma_{i,j,i,j}^{(h)}(y,y) = \frac{N_{i,j}^{(h)}(y)}{q^2} \ge 0$.
	\end{enumerate}
\end{lemma}
{\it Proof of \cref{properties-gamma}:}
	We prove the lemma by induction on $h$. The {\bf base of induction} corresponds to $h=0$. In the base case, by \cref{eq:dp-cntk-norm-simplified} and \cref{eq:dp-cntk-covar-simplified} and Cauchy–Schwarz inequality, we have
	\begin{align*}
		\left| \Gamma_{i,j,i',j'}^{(0)}(y,z) \right| &\equiv \left| \sum_{l=1}^c y_{i,j,l} \cdot z_{i',j',l} \right|\\
		&\le \sqrt{\sum_{l=1}^c |y_{i,j,l}|^2 \cdot \sum_{l=1}^c |z_{i',j',l}|^2}\\
		&= \frac{\sqrt{N_{i,j}^{(0)}(y) \cdot N_{i',j'}^{(0)}(z)}}{q^2}.
	\end{align*}
	This proves the base for the first statement. Additionally we have, $\Gamma_{i,j,i,j}^{(0)}(y,y) = \sum_{l=1}^c y_{i,j,l}^2 = \frac{N^{(0)}_{i,j}(y)}{q^2} \ge 0$ which proves the base for the second statement of the lemma.
	Now, in order to prove the inductive step, suppose that statements of the lemma hold for $h-1$, where $h\ge 1$. Then, conditioned on this, we prove that the lemma holds for $h$. First note that by conditioning on the inductive hypothesis, applying Cauchy–Schwarz inequality, and using the definition of $N^{(h)}$ in \cref{eq:dp-cntk-norm-simplified}, we can write
	\small
	\[ \begin{split}
		\left| \frac{\sum_{a=-\frac{q-1}{2}}^{\frac{q-1}{2}} \sum_{b=-\frac{q-1}{2}}^{\frac{q-1}{2}}  \Gamma^{(h-1)}_{i+a,j+b,i'+a,j'+b}(y,z)}{\sqrt{N^{(h)}_{i,j}(y) \cdot N^{(h)}_{i',j'}(z)}} \right| &\le \frac{\sum_{a=-\frac{q-1}{2}}^{\frac{q-1}{2}} \sum_{b=-\frac{q-1}{2}}^{\frac{q-1}{2}}  \sqrt{\frac{N^{(h-1)}_{i+a,j+b}(y)}{q^2} \cdot \frac{N^{(h-1)}_{i'+a,j'+b}(z)}{q^2}}}{\sqrt{N^{(h)}_{i,j}(y) \cdot N^{(h)}_{i',j'}(z)}}\\ 
		&\le 1.
	\end{split}
	\]
	\normalsize
	Thus, by monotonicity of the function $\kappa_1:[-1,1] \to \RR$, we can write,
	\begin{align*}
		\left| \Gamma_{i,j,i',j'}^{(h)}(y,z) \right| &\equiv \frac{\sqrt{N^{(h)}_{i,j}(y) \cdot N^{(h)}_{i',j'}(z)}}{q^2} \cdot \kappa_1\left( \frac{\sum_{a=-\frac{q-1}{2}}^{\frac{q-1}{2}} \sum_{b=-\frac{q-1}{2}}^{\frac{q-1}{2}}  \Gamma^{(h-1)}_{i+a,j+b,i'+a,j'+b}(y,z)}{\sqrt{N^{(h)}_{i,j}(y) \cdot N^{(h)}_{i',j'}(z)}} \right)\\
		& \le \frac{\sqrt{N^{(h)}_{i,j}(y) \cdot N^{(h)}_{i',j'}(z)}}{q^2} \cdot \kappa_1(1)\\
		&= \frac{\sqrt{N^{(h)}_{i,j}(y) \cdot N^{(h)}_{i',j'}(z)}}{q^2},
	\end{align*}
	where the second line above follows because of the fact that $\kappa_1(\cdot)$ is a monotonically increasing function. This completes the inductive step for the first statement of lemma. Now we prove the inductive step for the second statement as follows,
	\begin{align*}
		\Gamma_{i,j,i,j}^{(h)}(y,y) &\equiv \frac{N^{(h)}_{i,j}(y)}{q^2} \cdot \kappa_1\left( \frac{\sum_{a=-\frac{q-1}{2}}^{\frac{q-1}{2}} \sum_{b=-\frac{q-1}{2}}^{\frac{q-1}{2}}  \Gamma^{(h-1)}_{i+a,j+b,i+a,j+b}(y,y)}{N^{(h)}_{i,j}(y)} \right)\\
		&= \frac{N^{(h)}_{i,j}(y)}{q^2} \cdot \kappa_1(1)\\
		&= \frac{N^{(h)}_{i,j}(y)}{q^2} \ge 0,
	\end{align*}
	where we used \cref{N-to-Sigma} to conclude that $\sum_{a=-\frac{q-1}{2}}^{\frac{q-1}{2}} \sum_{b=-\frac{q-1}{2}}^{\frac{q-1}{2}}  \Gamma^{(h-1)}_{i+a,j+b,i+a,j+b}(y,y) =N^{(h)}_{i,j}(y)$ and then used the fact that $N^{(h)}_{i,j}(y)$ is non-negative. This completes the inductive proof of the lemma. This completes the proof of \cref{properties-gamma}.
\qed

We also describe some of the main properties of the function $\dot{\Gamma}^{(h)}(y,z)$ defined in \cref{eq:dp-cntk-derivative-covar-simplified} in the following lemma,
\begin{lemma}[Properties of $\dot{\Gamma}^{(h)}(y,z)$]\label{properties-gamma-dot}
	For every images $y,z \in \RR^{d_1 \times d_2 \times c}$, every integer $h \ge 0$ and every $i,i' \in [d_1]$ and $j,j' \in [d_2]$ the following properties are satisfied by function $\dot{\Gamma}^{(h)}$ defined in \cref{eq:dp-cntk-derivative-covar-simplified} of \cref{relu-cntk-def}:
	\begin{enumerate}
		\item {\bf Cauchy–Schwarz inequality:} $\left| \dot{\Gamma}_{i,j,i',j'}^{(h)}(y,z) \right| \le \frac{1}{q^2}$.
		\item {\bf Norm value:} $\dot{\Gamma}_{i,j,i,j}^{(h)}(y,y) = \frac{1}{q^2} \ge 0$.
	\end{enumerate}
\end{lemma}
{\it Proof of \cref{properties-gamma-dot}:}
	First, note that by \cref{properties-gamma} and the definition of $N^{(h)}$ in \cref{eq:dp-cntk-norm-simplified} we have,
	\small
	\[ \begin{split}
		\left| \frac{\sum_{a=-\frac{q-1}{2}}^{\frac{q-1}{2}} \sum_{b=-\frac{q-1}{2}}^{\frac{q-1}{2}}  \Gamma^{(h-1)}_{i+a,j+b,i'+a,j'+b}(y,z)}{\sqrt{N^{(h)}_{i,j}(y) \cdot N^{(h)}_{i',j'}(z)}} \right| &\le \frac{\sum_{a=-\frac{q-1}{2}}^{\frac{q-1}{2}} \sum_{b=-\frac{q-1}{2}}^{\frac{q-1}{2}}  \sqrt{\frac{N^{(h-1)}_{i+a,j+b}(y)}{q^2} \cdot \frac{N^{(h-1)}_{i'+a,j'+b}(z)}{q^2}}}{\sqrt{N^{(h)}_{i,j}(y) \cdot N^{(h)}_{i',j'}(z)}}\\ 
		&\le 1.
	\end{split}
	\]
	\normalsize
	Thus, by monotonicity of function $\kappa_0:[-1,1] \to \RR$ and using \cref{eq:dp-cntk-derivative-covar-simplified}, we can write, $ \dot{\Gamma}_{i,j,i',j'}^{(h)}(y,z) \le \frac{1}{q^2} \cdot \kappa_0(1) = \frac{1}{q^2}$. Moreover, the equality is achieved when $y=z$ and $i=i'$ and $j=j'$. This proves both statements of the lemma.
\qed

We also need to use some properties of $\Pi^{(h)}( y , z )$ defined in \cref{eq:dp-cntk} and \cref{eq:dp-cntk-last-layer}. We present these propertied in the next lemma,
\begin{lemma}[Properties of $\Pi^{(h)}$]\label{prop-pi}
	For every images $y,z \in \RR^{d_1 \times d_2 \times c}$, every integer $h \ge 0$ and every $i \in [d_1]$ and $j \in [d_2]$ the following properties are satisfied by the function $\Pi^{(h)}$ defined in \cref{eq:dp-cntk} and \cref{eq:dp-cntk-last-layer} of \cref{relu-cntk-def}:
	\begin{enumerate}
		\item {\bf Cauchy–Schwarz inequality:} $\Pi_{i,j,i',j'}^{(h)}(y,z) \le \sqrt{\Pi_{i,j,i,j}^{(h)}(y,y) \cdot \Pi_{i',j',i',j'}^{(h)}(z,z)}$.
		\item {\bf Norm value:} $\Pi_{i,j,i,j}^{(h)}(y,y) = \begin{cases}
			h \cdot N_{i,j}^{(h+1)}(y) & \text{if } h < L\\
			\frac{L-1}{q^2} \cdot N_{i,j}^{(L)}(y) & \text{if } h = L
		\end{cases}$.
	\end{enumerate}
\end{lemma}
{\it Proof of \cref{prop-pi}:}
	The proof is by induction on $h$. The base of induction corresponds to $h=0$. By definition of $\Pi_{i,j,i,j}^{(0)} \equiv 0$ in \cref{eq:dp-cntk}, the base of induction for both statements of the lemma follow immediately. 
	
	Now we prove the inductive hypothesis. Suppose that the lemma statement holds for $h-1$. We prove that conditioned on this, the statements of the lemma hold for $h$. There are two cases. The first case corresponds to $h<L$. In this case, by definition of $\Pi_{i,j,i,j}^{(h)}(x,x)$ in \cref{eq:dp-cntk} and using \cref{properties-gamma} and \cref{properties-gamma-dot} we can write,
	\small
	\begin{align*}
		&\left| \Pi^{(h)}_{i,j,i',j'}(y,z) \right| \equiv \left| \sum_{a=-\frac{q-1}{2}}^{\frac{q-1}{2}} \sum_{b=-\frac{q-1}{2}}^{\frac{q-1}{2}}  \left[\Pi^{(h-1)}(y,z) \odot \dot{\Gamma}^{(h)}(y,z) + \Gamma^{(h)}(y,z)\right]_{i+a,j+b,i'+a,j'+b} \right| \\
		&\le \sum_{a=-\frac{q-1}{2}}^{\frac{q-1}{2}} \sum_{b=-\frac{q-1}{2}}^{\frac{q-1}{2}}  \frac{\sqrt{\Pi^{(h-1)}_{i+a,j+b,i+a,j+b}(y,y) \cdot \Pi^{(h-1)}_{i'+a,j'+b,i'+a,j'+b}(z,z)}}{q^2} + \frac{\sqrt{N^{(h)}_{i+a,j+b}(y) \cdot N^{(h)}_{i'+a,j'+b}(z)}}{q^2}\\
		&\le \sum_{a=-\frac{q-1}{2}}^{\frac{q-1}{2}} \sum_{b=-\frac{q-1}{2}}^{\frac{q-1}{2}}  \sqrt{\frac{\Pi^{(h-1)}_{i+a,j+b,i+a,j+b}(y,y) + N^{(h)}_{i+a,j+b}(y)}{q^2}} \cdot \sqrt{\frac{\Pi^{(h-1)}_{i'+a,j'+b,i'+a,j'+b}(z,z) + N^{(h)}_{i'+a,j'+b}(z)}{q^2}}\\
		&\le \sqrt{\Pi^{(h)}_{i,j,i,j}(y,y)} \cdot \sqrt{ \Pi^{(h)}_{i',j',i',j'}(z,z)},
	\end{align*}
	\normalsize
	where the second line above follows from inductive hypothesis along with \cref{properties-gamma} and \cref{properties-gamma-dot}. The third and fourth lines above follow by Cauchy–Schwarz inequality.
	The second case corresponds to $h=L$. In this case, by definition of $\Pi_{i,j,i',j'}^{(L)}(y,z)$ in \cref{eq:dp-cntk-last-layer} and using \cref{properties-gamma} and \cref{properties-gamma-dot} along with the inductive hypothesis we can write,
	\small
	\begin{align*}
		\left| \Pi^{(L)}_{i,j,i',j'}(y,z) \right| &\equiv \left| \Pi^{(L-1)}_{i,j,i',j'}(y,z) \cdot \dot{\Gamma}^{(L)}_{i,j,i',j'}(y,z) \right| \\
		&\le \frac{\sqrt{\Pi^{(L-1)}_{i,j,i,j}(y,y) \cdot \Pi^{(L-1)}_{i',j',i',j'}(z,z)}}{q^2} \\
		&= \sqrt{\Pi^{(L)}_{i,j,i,j}(y,y)} \cdot \sqrt{ \Pi^{(L)}_{i',j',i',j'}(z,z)},
	\end{align*}
	\normalsize
	where the second line above follows from inductive hypothesis along with \cref{properties-gamma-dot}.
	This completes the inductive step and in turn proves the first statement of the lemma.
	
	To prove the inductive step for the second statement of lemma we consider two cases again. The first case is $h<L$. In this case, note that by using inductive hypothesis together with \cref{properties-gamma} and \cref{properties-gamma-dot} we can write,
	\begin{align*}
		\Pi^{(h)}_{i,j,i,j}(y,y) &\equiv \sum_{a=-\frac{q-1}{2}}^{\frac{q-1}{2}} \sum_{b=-\frac{q-1}{2}}^{\frac{q-1}{2}}  \left[\Pi^{(h-1)}(y,y) \odot \dot{\Gamma}^{(h)}(y,y) + \Gamma^{(h)}(y,y)\right]_{i+a,j+b,i+a,j+b}\\
		&= \sum_{a=-\frac{q-1}{2}}^{\frac{q-1}{2}} \sum_{b=-\frac{q-1}{2}}^{\frac{q-1}{2}} \frac{(h-1) \cdot N_{i+a,j+b}^{(h)}(y)}{q^2} + \frac{ N_{i+a,j+b}^{(h)}(y)}{q^2}\\
		&= h \cdot \sum_{a=-\frac{q-1}{2}}^{\frac{q-1}{2}} \sum_{b=-\frac{q-1}{2}}^{\frac{q-1}{2}}  \frac{N^{(h)}_{i+a,j+b}(y)}{q^2}\\
		&= h \cdot N^{(h+1)}_{i,j}(y),
	\end{align*}
	where the last line above follows from definition of $N^{(h)}$ in \cref{eq:dp-cntk-norm-simplified}. 
	The second case corresponds to $h=L$. In this case, by inductive hypothesis together with \cref{properties-gamma} and \cref{properties-gamma-dot} we can write,
	\begin{align*}
		\Pi^{(L)}_{i,j,i,j}(y,y) &\equiv \Pi^{(L-1)}_{i,j,i,j}(y,y) \cdot \dot{\Gamma}^{(L)}_{i,j,i,j}(y,y)\\
		&= \frac{(L-1) \cdot N_{i,j}^{(L)}(y)}{q^2}.
	\end{align*}
	This completes the inductive step for the second statement and in turn proves the second statement of the lemma. This completes the proof of \cref{prop-pi}.
\qed

\section{CNTK Sketch: Algorithm, Claims and Invariants}\label{app-cntk-sketch}

In this section we give our sketching algorithm for the CNTK kernel and prove our main theorem for this algorithm, i.e., \cref{maintheorem-cntk}.
We start by introducing our \CNTKS algorithm in the following definition:

\begin{defn}[\CNTKS Algorithm] \label{alg-def-cntk-sketch}
	For every image $x \in \RR^{d_1 \times d_2 \times c}$, we compute the \CNTKS, $\Psi_{\tt cntk}^{(L)}(x)$, recursively as follows,
	\begin{enumerate}[wide, labelwidth=!, labelindent=0pt]
		\item[$\bullet$] Let $s = \widetilde{\bigo}\left(\frac{L^2}{\varepsilon^2}  \right)$, $r = \widetilde{\bigo}\left(\frac{L^6}{\varepsilon^4}  \right)$, $n_1=\widetilde{\bigo}\left(\frac{L^4}{\varepsilon^4} \right)$, $m=\widetilde{\bigo}\left( \frac{L^8}{\varepsilon^{16/3}} \right)$, and $s^* = \bigo(\frac{1}{\varepsilon^2} \log\frac{1}{\delta})$ and $P^{(p)}_{\tt relu}(\alpha) = \sum_{l=0}^{2p+2} c_l \cdot \alpha^l$ and $\dot{P}^{(p')}_{\tt relu}(\alpha) = \sum_{l=0}^{2p'+1} b_l \cdot \alpha^l$ be the polynomials defined in \cref{eq:poly-approx-krelu}.
		
		\item For every $i \in [d_1]$, $j \in [d_2]$, and $h = 0, 1, 2, \ldots L$ compute $N_{i,j}^{(h)}(x)$ as per \cref{eq:dp-cntk-norm-simplified} of \cref{relu-cntk-def}.
		\item Let $\S \in \RR^{r \times c}$ be an SRHT. For every $i \in [d_1]$ and $j \in [d_2]$, compute $\phi_{i,j}^{(0)}(x) \in \RR^r$ as,
		\begin{equation}\label{cntk-sketch-covar-zero}
			\phi_{i,j}^{(0)}(x) \gets \S \cdot x_{(i,j,:)}.
		\end{equation}
		
		\item
		Let $\Q^{2p+2} \in \RR^{m \times \left(q^2r\right)^{2p+2}}$ be a degree-$2p+2$ \PolyS, and $\T \in \RR^{r \times (2p+3)\cdot m}$ be an SRHT.
		For every $h \in [L]$, every $i \in [d_1]$ and $j \in [d_2]$, and $l=0,1,2, \ldots 2p+2$ compute:
		\begin{equation}\label{eq:maping-cntk-covar}
			\begin{split}
			&\mu^{(h)}_{i,j}(x) \gets   \bigoplus_{a=-\frac{q-1}{2}}^{\frac{q-1}{2}} \bigoplus_{b=-\frac{q-1}{2}}^{\frac{q-1}{2}}  \frac{\phi_{i+a,j+b}^{(h-1)}(x)}{\sqrt{N^{(h)}_{i,j}(x)}},\\ &\left[Z^{(h)}_{i,j}(x)\right]_l \gets \Q^{2p+2} \cdot \left(\left[ \mu^{(h)}_{i,j}(x) \right]^{\otimes l} \otimes {e}_1^{\otimes 2p+2-l}\right), \\	&\phi_{i,j}^{(h)}(x) \gets \frac{\sqrt{N^{(h)}_{i,j}(x)}}{q} \cdot \T \cdot \left( \bigoplus_{l=0}^{2p+2} \sqrt{c_l}  \left[Z^{(h)}_{i,j}(x)\right]_l \right).
			\end{split}
		\end{equation}
		\item 
		Let $\Q^{2p'+1} \in \RR^{n_1 \times \left(q^2r\right)^{2p'+1}}$ be a degree-$2p'+1$ \PolyS, and $\W \in \RR^{s \times (2p'+2)\cdot n_1}$ be an SRHT.
		For every $h \in [L], i \in [d_1], j \in [d_2]$, and $l=0,1, \ldots 2p'+1$ compute:
		\begin{equation}\label{eq:cntk-map-phidot}
		\begin{split}
			&\left[Y^{(h)}_{i,j}(x)\right]_l \gets \Q^{2p'+1} \left(\left[ \mu^{(h)}_{i,j}(x) \right]^{\otimes l} \otimes {e}_1^{\otimes 2p'+1-l}\right),\\ &\dot{\phi}_{i,j}^{(h)}(x) \gets \frac{1}{q} \cdot \W \left( \bigoplus_{l=0}^{2p'+1} \sqrt{b_l}  \left[Y^{(h)}_{i,j}(x)\right]_l \right).
		\end{split}
		\end{equation}

		\item Let $\Q^{2} \in \RR^{s \times s^2}$ be a degree-$2$ \PolyS, and ${\bm R} \in \RR^{s \times q^2(s+r)}$ be an SRHT. 
		Let $\psi^{(0)}_{i,j}(x) \gets 0$ and for every $h \in [L-1]$, and $i \in [d_1], j \in [d_2]$, compute $\psi^{(h)}_{i,j}(x) \in \RR^s$ as:
		\begin{equation}\label{psi-cntk}
		\begin{split}
			&\eta^{(h)}_{i,j}(x) \gets \Q^2 \left(\psi_{i,j}^{(h-1)}(x) \otimes \dot{\phi}_{i,j}^{(h)}(x)\right) \oplus \phi_{i,j}^{(h)}(x),\\
			&\psi^{(h)}_{i,j}(x) \gets {\bm R} \left(\bigoplus_{a=-\frac{q-1}{2}}^{\frac{q-1}{2}} \bigoplus_{b=-\frac{q-1}{2}}^{\frac{q-1}{2}}  \eta_{i+a,j+b}^{(h)}(x)\right).
		\end{split}
		\end{equation}
		\begin{equation}\label{psi-cntk-last}
			\text{(For $ h=L$:) } ~~~~~~~~~\psi^{(L)}_{i,j}(x) \gets \Q^2 \cdot \left(\psi^{(L-1)}_{i,j}(x) \otimes \dot{\phi}_{i,j}^{(L)}(x)\right).
		\end{equation}
		\item Let $\G \in \RR^{s^* \times s}$ be a random matrix of i.i.d. normal entries with distribution $\mathcal{N}(0,1/s^*)$. The \CNTKS is the following:
		\begin{equation}\label{Psi-cntk-def}
			\Psi_{\tt cntk}^{(L)}(y,z) := \frac{1}{d_1d_2} \cdot \G \cdot \left(\sum_{i \in [d_1]} \sum_{j \in [d_2]} \psi^{(L)}_{i,j}(x)\right).
		\end{equation}
	\end{enumerate}
\end{defn}

In the following lemma, we analyze the correctness of the \CNTKS algorithm by giving the invariants that the algorithm maintains at all times,
\begin{lemma}[Invariants of the \CNTKS]
	\label{lem:cntk-sketch-corr}
	For every positive integers $d_1, d_2, c$, and $L$, every $\varepsilon, \delta>0$, every images $y,z \in \RR^{d_1\times d_2 \times c}$ 
	, if we let $N^{(h)}: \RR^{d_1\times d_2\times c}\to \RR^{d_1\times d_2}$, $\Gamma^{(h)}(y,z) \in \RR^{d_1\times d_2 \times d_1\times d_2}$ and $\Pi^{(h)}(y,z) \in \RR^{d_1\times d_2 \times d_1\times d_2}$ be the tensor functions defined in \cref{eq:dp-cntk-norm-simplified}, \cref{eq:dp-cntk-covar-simplified}, \cref{eq:dp-cntk}, and \cref{eq:dp-cntk-last-layer} of \cref{relu-cntk-def}, respectively, then with probability at least $1-\delta$ the following invariants are maintained simultaneously for all $i,i' \in [d_1]$ and $j,j' \in [d_2]$ and every $h =0, 1, 2, \ldots L$:
	\begin{enumerate}
		\item The mapping $\phi_{i,j}^{(h)}(\cdot)$ computed by the CNTK Sketch algorithm in \cref{cntk-sketch-covar-zero} and \cref{eq:maping-cntk-covar} of \cref{alg-def-cntk-sketch} satisfy the following,
		\[ \left| \left< \phi_{i,j}^{(h)}(y), \phi_{i',j'}^{(h)}(z) \right> - \Gamma_{i,j,i',j'}^{(h)}\left( y , z \right) \right| \le ({h+1}) \cdot \frac{\varepsilon^2}{60L^3}\cdot \frac{\sqrt{N_{i,j}^{(h)}(y) \cdot N_{i',j'}^{(h)}(z)}}{q^2}. \]
		\item The mapping $\psi_{i,j}^{(h)}(\cdot)$ computed by the CNTK Sketch algorithm in \cref{psi-cntk} and \cref{psi-cntk-last} of \cref{alg-def-cntk-sketch} satisfy the following,
		\[ \left| \left< \psi_{i,j}^{(h)}(y), \psi_{i',j'}^{(h)}(z) \right> - \Pi_{i,j,i',j'}^{(h)}\left( y , z \right) \right| \le \begin{cases}
			\frac{\varepsilon}{10} \cdot \frac{h^2}{L+1} \cdot \sqrt{N^{(h+1)}_{i,j}(y) \cdot N^{(h+1)}_{i',j'}(z)} &\text{if } h<L\\
			\frac{\varepsilon}{10} \cdot \frac{L-1}{q^2} \cdot \sqrt{N^{(L)}_{i,j}(y) \cdot N^{(L)}_{i',j'}(z)} &\text{if } h=L
		\end{cases}. \]
	\end{enumerate}
\end{lemma}
{\it Proof of \cref{lem:cntk-sketch-corr}:}
	The proof is by induction on the value of $h=0,1,2, \ldots L$. 
	More formally, consider the following statements for every $h=0,1,2, \ldots L$:
	\begin{enumerate}[leftmargin=1.5cm]
		\item[${\bf P_1(h) :}$] Simultaneously for all $i,i' \in [d_1]$ and $j,j' \in [d_2]$:
		\[ \begin{split}
			&\left| \left< \phi_{i,j}^{(h)}(y), \phi_{i',j'}^{(h)}(z) \right> - \Gamma_{i,j,i',j'}^{(h)}\left( y , z \right) \right| \le ({h+1}) \cdot \frac{\varepsilon^2}{60L^3} \cdot \frac{\sqrt{N_{i,j}^{(h)}(y) \cdot N_{i',j'}^{(h)}(z)}}{q^2},\\
			&\left| \left\| \phi_{i,j}^{(h)}(y) \right\|_2^2 - \Gamma_{i,j,i,j}^{(h)}\left( y , y \right) \right| \le  \frac{({h+1}) \cdot\varepsilon^2}{60L^3} \cdot \frac{N_{i,j}^{(h)}(y)}{q^2}, \\ 
			&\left| \left\| \phi_{i',j'}^{(h)}(z) \right\|_2^2 - \Gamma_{i',j',i',j'}^{(h)}\left( z , z \right) \right| \le  \frac{({h+1}) \cdot\varepsilon^2}{60L^3} \cdot \frac{N_{i',j'}^{(h)}(z)}{q^2}.
		\end{split} \]
		\item[${\bf P_2(h) :}$] Simultaneously for all $i,i' \in [d_1]$ and $j,j' \in [d_2]$:
		\[ \begin{split}
			&\left| \left< \psi_{i,j}^{(h)}(y), \psi_{i',j'}^{(h)}(z) \right> - \Pi_{i,j,i',j'}^{(h)}\left( y , z \right) \right| \le \begin{cases}
				\frac{\varepsilon}{10} \cdot \frac{h^2}{L+1} \cdot \sqrt{N^{(h+1)}_{i,j}(y) \cdot N^{(h+1)}_{i',j'}(z)} &\text{if } h<L\\
				\frac{\varepsilon}{10} \cdot \frac{L-1}{q^2} \cdot \sqrt{N^{(L)}_{i,j}(y) \cdot N^{(L)}_{i',j'}(z)} &\text{if } h=L
			\end{cases},\\
			&{(\text{only for } h<L):} ~~~ \left| \left\| \psi_{i,j}^{(h)}(y)\right\|_2^2 - \Pi_{i,j,i,j}^{(h)}\left( y , y \right) \right| \le \frac{\varepsilon}{10} \cdot \frac{h^2}{L+1} \cdot N^{(h+1)}_{i,j}(y),\\
			&{(\text{only for } h<L):} ~~~ \left| \left\| \psi_{i',j'}^{(h)}(z)\right\|_2^2 - \Pi_{i',j',i',j'}^{(h)}\left( z , z \right) \right| \le \frac{\varepsilon}{10} \cdot \frac{h^2}{L+1} \cdot N^{(h+1)}_{i',j'}(z).
		\end{split}
		\]
		
	\end{enumerate}
	We prove that probabilities $\Pr[ P_1(0)]$ and $\Pr[ P_2(0)|P_1(0)]$ are both greater than $1 - \bigo(\delta/L)$. Additionally, for every $h = 1,2, \ldots L$, we prove that the conditional probabilities $\Pr[ P_1(h) | P_1(h-1)]$ and $\Pr[ P_2(h) | P_2(h-1), P_1(h), P_1(h-1)]$ are greater than $1 - \bigo(\delta/L)$.
	
	The {\bf base of induction} corresponds to $h=0$. By \cref{cntk-sketch-covar-zero}, $\phi_{i,j}^{(0)}(y) = \S \cdot y_{(i,j,:)}$ and $\phi_{i',j'}^{(0)}(z) = \S \cdot z_{(i',j',:)}$, thus, \cref{lem:srht} implies the following
	\[ \Pr\left[ \left| \left< \phi_{i,j}^{(0)}(y), \phi_{i',j'}^{(0)}(z)\right> - \left< y_{(i,j,:)}, z_{(i',j',:)} \right> \right| \le \bigo\left(\varepsilon^2/L^3\right)\cdot \| y_{(i,j,:)} \|_2 \| z_{(i',j',:)} \|_2 \right] \ge 1 - \bigo\left(\frac{\delta}{d_1^2d_2^2L}\right), \]
	therefore, by using \cref{eq:dp-cntk-norm-simplified} and \cref{eq:dp-cntk-covar-simplified} we have
	\small
	\[ \Pr\left[ \left| \left< \phi_{i,j}^{(0)}(y), \phi_{i',j'}^{(0)}(z)\right> - \Gamma_{i,j,i',j'}^{(0)}(y,z) \right| \le \bigo\left(\varepsilon^2/L^3\right) \cdot \frac{\sqrt{N^{(0)}_{i,j}(y) \cdot N_{i',j'}^{(0)}(z) }}{q^2} \right] \ge 1 - \bigo\left(\frac{\delta}{d_1^2d_2^2L}\right). \]
	\normalsize
	Similarly, we can prove that with probability at least $1 - \bigo\left(\frac{\delta}{d_1^2d_2^2L}\right)$, the following hold
	\small
	\[ \begin{split}
		&\left| \left\| \phi_{i,j}^{(0)}(y)\right\|_2^2 - \Gamma_{i,j,i,j}^{(0)}\left( y , y \right) \right| \le \bigo\left(\varepsilon^2/L^3\right) \cdot \frac{N^{(0)}_{i,j}(y)}{q^2}, \\ 
		&\left| \left\| \phi_{i',j'}^{(0)}(z)\right\|_2^2 - \Gamma_{i',j',i',j'}^{(0)}\left( z , z \right) \right| \le \bigo\left(\varepsilon^2/L^3\right) \cdot \frac{N^{(0)}_{i',j'}(z)}{q^2}.
	\end{split}\]
	\normalsize
	By union bounding over all $i,i' \in [d_1]$ and $j,j' \in [d_2]$, this proves the base of induction for statement $P_1(h)$, i.e., $\Pr[ P_1(0) ] \ge 1 - \bigo(\delta/L)$.
	
	Moreover, by \cref{psi-cntk}, we have that $\psi_{i,j}^{(0)}(y) = 0$ and $\psi_{i',j'}^{(0)}(z) = 0$, thus, by \cref{eq:dp-cntk}, it trivially holds that $\Pr[P_2(0)|P_1(0)] = 1 \ge 1 - \bigo(\delta/L)$. This completes the base of induction.
	
	Now, we proceed to prove the {\bf inductive step}. That is, by assuming the inductive hypothesis for $h-1$, we prove that statements $P_1(h)$ and $P_2(h)$ hold. More precisely, first we condition on the statement $P_1(h-1)$ being true for some $h \ge 1$, and then prove that $P_1(h)$ holds with probability at least $1 - \bigo(\delta / L)$. Next we show that conditioned on statements $P_2(h-1), P_1(h), P_1(h-1)$ being true, $P_2(h)$ holds with probability at least $1 - \bigo(\delta / L)$. This will complete the induction.
	
	First, note that by \cref{lem:srht}, union bound, and using \cref{eq:maping-cntk-covar}, the following holds simultaneously for all $i,i' \in [d_1]$ and all $j,j' \in [d_2]$, with probability at least $1 - \bigo\left(\frac{\delta}{L}\right)$,
	\small
	\begin{equation}\label{eq:phi-inner-prod-bound1}
		\left| \left< \phi_{i,j}^{(h)}(y), \phi_{i',j'}^{(h)}(z)\right> - \frac{\sqrt{N^{(h)}_{i,j}(y) N^{(h)}_{i',j'}(z)}}{q^2} \cdot \sum_{l=0}^{2p+2} c_l \left<\left[Z^{(h)}_{i,j}(y)\right]_l, \left[Z^{(h)}_{i',j'}(z)\right]_l\right> \right| \le \bigo\left(\frac{\varepsilon^2}{L^3}\right) \cdot A,
	\end{equation}
	\normalsize
	where $A := \frac{\sqrt{N^{(h)}_{i,j}(y) N^{(h)}_{i',j'}(z)}}{q^2} \cdot \sqrt{\sum_{l=0}^{2p+2} c_l \left\| \left[ Z^{(h)}_{i,j}(y) \right]_l \right\|_2^2} \cdot \sqrt{\sum_{l=0}^{2p+2} c_l \left\| \left[Z^{(h)}_{i',j'}(z)\right]_l \right\|_2^2}$ and the collection of vectors $\left\{\left[Z^{(h)}_{i,j}(y)\right]_l\right\}_{l=0}^{2p+2}$ and $\left\{\left[Z^{(h)}_{i',j'}(z)\right]_l\right\}_{l=0}^{2p+2}$ and coefficients $c_0, c_1, c_2, \ldots c_{2p+2}$ are defined as per \cref{eq:maping-cntk-covar} and \cref{eq:poly-approx-krelu}, respectively. Additionally, by \cref{soda-result} and union bound, the following inequalities hold, with probability at least $1 - \bigo\left( \frac{\delta}{L} \right)$, simultaneously for all $l = 0,1,2, \ldots 2p+2$, all $i,i' \in [d_1]$ and all $j,j' \in [d_2]$:
	\begin{align}
		&\left|\left<\left[Z^{(h)}_{i,j}(y)\right]_l, \left[Z^{(h)}_{i',j'}(z)\right]_l\right> - \left<\mu_{i,j}^{(h)}(y), \mu_{i',j'}^{(h)}(z)\right>^l \right| \le O\left( \frac{\varepsilon^2}{L^3} \right) \left\| \mu_{i,j}^{(h)}(y) \right\|_2^l \left\| \mu_{i',j'}^{(h)}(z)\right\|_2^l \nonumber\\
		&\left\| \left[Z^{(h)}_{i,j}(y)\right]_l \right\|_2^2 \le \frac{11}{10} \cdot \left\| \mu_{i,j}^{(h)}(y) \right\|_2^{2l} \label{eq:Zinner-prod-bound}\\
		& \left\| \left[Z^{(h)}_{i',j'}(z)\right]_l \right\|_2^2 \le \frac{11}{10} \cdot \left\| \mu_{i',j'}^{(h)}(z) \right\|_2^{2l} \nonumber
	\end{align}
	Therefore, by plugging \cref{eq:Zinner-prod-bound} back to \cref{eq:phi-inner-prod-bound1} and using union bound and triangle inequality as well as 
	Cauchy–Schwarz inequality, we find that with probability at least $1 - \bigo\left( \frac{\delta}{L} \right)$, the following holds simultaneously for all $i,i' \in [d_1]$ and $j,j' \in [d_2]$
	\small
	\begin{equation} \label{eq:phi-inner-prod-bound2}
		\left| \left< \phi_{i,j}^{(h)}(y), \phi_{i',j'}^{(h)}(z)\right> - \frac{\sqrt{N^{(h)}_{i,j}(y) N^{(h)}_{i',j'}(z)}}{q^2} \cdot P^{(p)}_{\tt relu}\left( \left<\mu_{i,j}^{(h)}(y), \mu_{i',j'}^{(h)}(z)\right> \right) \right| \le \bigo\left(\frac{\varepsilon^2}{L^3}\right) \cdot B,
	\end{equation}
	\normalsize
	where $B:= \frac{\sqrt{N^{(h)}_{i,j}(y) N^{(h)}_{i',j'}(z)}}{q^2} \cdot \sqrt{P^{(p)}_{\tt relu}\left(\|\mu^{(h)}_{i,j}(y)\|_2^2\right) \cdot P^{(p)}_{\tt relu}\left(\|\mu^{(h)}_{i',j'}(z)\|_2^2\right)}$ and $P^{(p)}_{\tt relu}(\alpha) = \sum_{l=0}^{2p+2} c_l \cdot \alpha^l$ is the polynomial defined in \cref{eq:poly-approx-krelu}. By using the definition of $\mu_{i,j}^{(h)}(\cdot)$ in \cref{eq:maping-cntk-covar} we have,
	\begin{equation}\label{mu-equality}
		\begin{split}
			&\left<\mu_{i,j}^{(h)}(y), \mu_{i',j'}^{(h)}(z)\right> = \frac{\sum_{a=-\frac{q-1}{2}}^{\frac{q-1}{2}} \sum_{b=-\frac{q-1}{2}}^{\frac{q-1}{2}}  \left< \phi_{i+a,j+b}^{(h-1)}(y), \phi_{i'+a,j'+b}^{(h-1)}(z) \right>}{\sqrt{N^{(h)}_{i,j}(y) N^{(h)}_{i',j'}(z)}},\\
			&\left\|\mu_{i,j}^{(h)}(y)\right\|_2^2 = \frac{\sum_{a=-\frac{q-1}{2}}^{\frac{q-1}{2}} \sum_{b=-\frac{q-1}{2}}^{\frac{q-1}{2}}  \left\| \phi_{i+a,j+b}^{(h-1)}(y) \right\|_2^2}{N^{(h)}_{i,j}(y)},\\
			&\left\|\mu_{i',j'}^{(h)}(z)\right\|_2^2 = \frac{\sum_{a=-\frac{q-1}{2}}^{\frac{q-1}{2}} \sum_{b=-\frac{q-1}{2}}^{\frac{q-1}{2}}  \left\| \phi_{i'+a,j'+b}^{(h-1)}(z) \right\|_2^2}{N^{(h)}_{i,j}(z)}.
		\end{split} 
	\end{equation}
	Hence, by conditioning on the inductive hypothesis $P_1(h-1)$ and using \cref{mu-equality} and \cref{N-to-Sigma} we have, 
	\[
	\left| \left\| \mu_{i,j}^{(h)}(y) \right\|_2^2 - 1 \right| \le h \cdot \frac{\varepsilon^2}{60L^3}, \text{ and } \left| \left\| \mu_{i',j'}^{(h)}(z) \right\|_2^2 - 1 \right| \le h \cdot \frac{\varepsilon^2}{60L^3}.
	\]
	Therefore, by invoking  \cref{lema:sensitivity-polynomial}, it follows that $\left| P_{\tt relu}^{(p)}\left(\|\mu^{(h)}_{i,j}(y)\|_2^2\right) - P_{\tt relu}^{(p)}(1) \right| \le h \cdot \frac{\varepsilon^2}{60L^3}$ and $\left| P_{\tt relu}^{(p)}\left(\|\mu^{(h)}_{i',j'}(z)\|_2^2\right) - P_{\tt relu}^{(p)}(1) \right| \le h \cdot \frac{\varepsilon^2}{60L^3}$. Consequently, because $P_{\tt relu}^{(p)}(1) \le P_{\tt relu}^{(+\infty)}(1) = 1$, we find that
	\[ B \le \frac{11}{10} \cdot \frac{\sqrt{N^{(h)}_{i,j}(y) N^{(h)}_{i',j'}(z)}}{q^2}.\]
	For shorthand we use the notation $\beta:=\frac{\sqrt{N^{(h)}_{i,j}(y) N^{(h)}_{i',j'}(z)}}{q^2}$. By plugging this into \cref{eq:phi-inner-prod-bound2} and using the notation $\beta$, we find that the following holds simultaneously for all $i,i' \in [d_1]$ and all $j,j' \in [d_2]$, with probability at least $1 - \bigo\left( \frac{\delta}{ L} \right)$,
	\begin{equation} \label{eq:phi-inner-prod-bound3}
		\left| \left< \phi_{i,j}^{(h)}(y), \phi_{i',j'}^{(h)}(z)\right> - \beta \cdot P^{(p)}_{\tt relu}\left( \left<\mu_{i,j}^{(h)}(y), \mu_{i',j'}^{(h)}(z)\right> \right) \right| \le \bigo\left(\frac{\varepsilon^2}{L^3}\right) \cdot \beta.
	\end{equation}
	Furthermore, by conditioning on the inductive hypothesis $P_1(h-1)$ and combining it with \cref{mu-equality} and applying Cauchy–Schwarz inequality and invoking \cref{N-to-Sigma} we find that,
	\small
	\begin{equation}\label{eq:mu-innerprod-bound}
		\begin{split}
			&\left| \left<\mu_{i,j}^{(h)}(y), \mu_{i',j'}^{(h)}(z)\right> - \frac{\sum_{a=-\frac{q-1}{2}}^{\frac{q-1}{2}} \sum_{b=-\frac{q-1}{2}}^{\frac{q-1}{2}}  \Gamma_{i+a,j+b,i'+a,j'+b}^{(h-1)}\left( y , z \right)}{\sqrt{N^{(h)}_{i,j}(y) \cdot N^{(h)}_{i',j'}(z)}} \right|\\ 
			&\qquad \le \frac{\sum_{a=-\frac{q-1}{2}}^{\frac{q-1}{2}} \sum_{b=-\frac{q-1}{2}}^{\frac{q-1}{2}}  \sqrt{N_{i+a,j+b}^{(h-1)}(y) \cdot N_{i'+a,j'+b}^{(h-1)}(z)}}{q^2 \cdot \sqrt{N^{(h)}_{i,j}(y) \cdot N^{(h)}_{i',j'}(z)}} \cdot h \cdot \frac{\varepsilon^2}{60L^3}\\
			&\qquad \le \frac{\sqrt{\sum_{a=-\frac{q-1}{2}}^{\frac{q-1}{2}} \sum_{b=-\frac{q-1}{2}}^{\frac{q-1}{2}} \frac{N_{i+a,j+b}^{(h-1)}(y)}{q^2} } \cdot \sqrt{\sum_{a=-\frac{q-1}{2}}^{\frac{q-1}{2}} \sum_{b=-\frac{q-1}{2}}^{\frac{q-1}{2}} \frac{N_{i'+a,j'+b}^{(h-1)}(z)}{ q^2}}}{ \sqrt{N^{(h)}_{i,j}(y) \cdot N^{(h)}_{i',j'}(z)}} \cdot \frac{h \cdot \varepsilon^2}{60L^3}\\
			&\qquad= h \cdot \frac{\varepsilon^2}{60L^3},
		\end{split}
	\end{equation}
	\normalsize
	where the last line follows from \cref{eq:dp-cntk-norm-simplified}.
	
	For shorthand, we use the notation $\gamma := \frac{\sum_{a=-\frac{q-1}{2}}^{\frac{q-1}{2}} \sum_{b=-\frac{q-1}{2}}^{\frac{q-1}{2}}  \Gamma_{i+a,j+b,i'+a,j'+b}^{(h-1)}\left( y , z \right)}{\sqrt{N^{(h)}_{i,j}(y) \cdot N^{(h)}_{i',j'}(z)}}$. Note that by \cref{properties-gamma} and \cref{eq:dp-cntk-norm-simplified}, $-1 \le \gamma \le 1$. Hence, we can invoke \cref{lema:sensitivity-polynomial} and use \cref{eq:mu-innerprod-bound} to find that,
	\[ \left|P^{(p)}_{\tt relu}\left( \left<\mu_{i,j}^{(h)}(y), \mu_{i',j'}^{(h)}(z)\right> \right) - P^{(p)}_{\tt relu}\left(\gamma\right) \right| \le h \cdot \frac{\varepsilon^2}{60L^3}. \]
	By incorporating the above inequality into \cref{eq:phi-inner-prod-bound3} using triangle inequality we find that, with probability at least $1 - \bigo\left( \frac{\delta}{ L} \right)$, the following holds simultaneously for all $i,i' \in [d_1]$ and all $j,j'\in [d_2]$:
	\begin{equation} \label{eq:phi-inner-prod-bound4}
		\left| \left< \phi_{i,j}^{(h)}(y), \phi_{i',j'}^{(h)}(z)\right> - \beta \cdot P^{(p)}_{\tt relu}\left(\gamma \right) \right| \le \left(\bigo\left(\frac{\varepsilon^2}{L^3}\right) + \frac{h \cdot\varepsilon^2}{60L^3}\right) \cdot \beta.
	\end{equation}
	Additionally, since $-1 \le \gamma \le 1$, we can invoke \cref{lem:polynomi-approx-krelu} and use the fact that $p = \left\lceil 2L^2/{\varepsilon}^{4/3} \right\rceil$ to conclude,
	\[ \left| P_{\tt relu}^{(p)}\left(\gamma\right) - \kappa_1(\gamma ) \right| \le \frac{\varepsilon^2}{76 L^3}. \]
	By combining the above inequality with \cref{eq:phi-inner-prod-bound4} via triangle inequality and using the fact that, by \cref{eq:dp-cntk-covar-simplified}, $\beta \cdot \kappa_1(\gamma) \equiv \Gamma_{i,j,i',j'}^{(h)}(y,z)$ we get the following inequality, with probability at least $1 - \bigo\left( \frac{\delta}{ L} \right)$,
	\[  \left| \left< \phi_{i,j}^{(h)}(y), \phi_{i',j'}^{(h)}(z)\right> - \Gamma_{i,j,i',j'}^{(h)}(y,z) \right| \le (h+1) \cdot \frac{\varepsilon^2}{60L^3} \cdot \frac{\sqrt{N^{(h)}_{i,j}(y) N^{(h)}_{i',j'}(z)}}{q^2}. \]
	Similarly, we can prove that with probability at least $1 - \bigo\left( \frac{\delta}{ L} \right)$ the following hold, simultaneously for all $i,i' \in [d_1]$ and $j,j' \in [d_2]$,
	\small
	\[ 
	\begin{split}
	&\left| \left\| \phi_{i,j}^{(h)}(y)\right\|_2^2 - \Gamma_{i,j,i,j}^{(h)}(y,y) \right| \le \frac{(h+1) \varepsilon^2}{60L^3} \cdot \frac{N^{(h)}_{i,j}(y)}{q^2},\\ &\left| \left\| \phi_{i',j'}^{(h)}(z)\right\|_2^2 - \Gamma_{i',j',i',j'}^{(h)}(z,z) \right| \le \frac{(h+1) \varepsilon^2}{60L^3} \cdot \frac{N^{(h)}_{i',j'}(z)}{q^2}. 
	\end{split}
	\]
	\normalsize
	This is sufficient to prove the inductive step for statement $P_1(h)$, i.e., $\Pr[P_1(h)|P_1(h-1)] \ge 1 - \bigo(\delta/L)$.
	
	Now we prove the inductive step for statement $P_2(h)$. That is, we prove that conditioned on $P_2(h-1), P_1(h)$, and $P_1(h-1)$, $P_2(h)$ holds with probability at least $1-\bigo(\delta/L)$.
	First, note that by \cref{lem:srht} and using \cref{eq:cntk-map-phidot} and union bound, we have the following simultaneously for all $i,i' \in [d_1]$ and all $j,j' \in [d_2]$, with probability at least $1 - O\left( \frac{\delta}{L} \right)$,
	\begin{equation}\label{cntk:phi-dot-bound1}
		\left| \left< \dot{\phi}_{i,j}^{(h)}(y), \dot{\phi}_{i',j'}^{(h)}(z)\right> - \frac{1}{q^2}  \sum_{l=0}^{2p'+1} b_l \left<\left[Y^{(h)}_{i,j}(y)\right]_l, \left[Y^{(h)}_{i',j'}(z)\right]_l\right> \right| \le \bigo\left(\frac{\varepsilon}{L}\right) \widehat{A},
	\end{equation}
	where $\widehat{A} := \frac{1}{q^2} \cdot \sqrt{\sum_{l=0}^{2p'+1} b_l \left\| \left[Y^{(h)}_{i,j}(y)\right]_l \right\|_2^2} \cdot \sqrt{\sum_{l=0}^{2p'+1} b_l \left\| \left[Y^{(h)}_{i',j'}(z)\right]_l \right\|_2^2}$ and the collection of vectors $\left\{\left[Y^{(h)}_{i,j}(y)\right]_l\right\}_{l=0}^{2p'+1}$ and $\left\{\left[Y^{(h)}_{i',j'}(z)\right]_l\right\}_{l=0}^{2p'+1}$ and coefficients $b_0, b_1, b_2, \ldots b_{2p'+1}$ are defined as per \cref{eq:cntk-map-phidot} and \cref{eq:poly-approx-krelu}, respectively. 
	By \cref{soda-result} and union bound, with probability at least $1 - \bigo\left( \frac{\delta}{L} \right)$, the following inequalities hold true simultaneously for all $l \in \{0,1,2, \ldots 2p'+1\}$, all $i,i' \in [d_1]$ and all $j,j' \in [d_2]$,
	\begin{align}
		& \left|\left<\left[Y^{(h)}_{i,j}(y)\right]_l, \left[Y^{(h)}_{i',j'}(z)\right]_l\right> - \left<\mu_{i,j}^{(h)}(y), \mu_{i',j'}^{(h)}(z)\right>^l \right| \le \bigo\left( \frac{\varepsilon}{L} \right) \cdot \left\| \mu_{i,j}^{(h)}(y) \right\|_2^l \left\| \mu_{i',j'}^{(h)}(z)\right\|_2^l \nonumber\\
		& \left\| \left[Y^{(h)}_{i,j}(y)\right]_l\right\|_2^2 \le \frac{11}{10} \cdot \left\| \mu_{i,j}^{(h)}(y) \right\|_2^{2l} \label{eq:Y-innerprod-bound}\\
		& \left\| \left[Y^{(h)}_{i',j'}(z)\right]_l \right\|_2^2 \le \frac{11}{10} \cdot \left\| \mu_{i',j'}^{(h)}(z) \right\|_2^{2l} \nonumber
	\end{align}
	Therefore, by plugging \cref{eq:Y-innerprod-bound} into \cref{cntk:phi-dot-bound1} and using union bound and triangle inequality as well as 
	Cauchy–Schwarz inequality, we find that with probability at least $1 - \bigo\left( \frac{\delta}{L} \right)$, the following holds simultaneously for all $i,i' \in [d_1]$ and $j,j' \in [d_2]$ 
	\begin{equation} \label{cntk:phi-dot-bound2}
		\left| \left< \dot{\phi}_{i,j}^{(h)}(y), \dot{\phi}_{i',j'}^{(h)}(z)\right> - \frac{1}{q^2} \cdot \dot{P}^{(p')}_{\tt relu}\left( \left<\mu_{i,j}^{(h)}(y), \mu_{i',j'}^{(h)}(z)\right> \right) \right| \le \bigo\left(\frac{\varepsilon}{L}\right) \cdot \widehat{B},
	\end{equation}
	where $\widehat{B}:= \frac{1}{q^2} \cdot \sqrt{\dot{P}^{(p')}_{\tt relu}\left(\|\mu_{i,j}^{(h)}(y)\|_2^2\right) \cdot \dot{P}^{(p')}_{\tt relu}\left(\|\mu_{i',j'}^{(h)}(z)\|_2^2\right)}$ and $\dot{P}^{(p)}_{\tt relu}(\alpha) = \sum_{l=0}^{2p'+1} b_l \cdot \alpha^l$ is the polynomial defined in \cref{eq:poly-approx-krelu}.
	By conditioning on the inductive hypothesis $P_1(h-1)$ and using \cref{mu-equality} and \cref{N-to-Sigma} we have $\left| \left\| \mu_{i,j}^{(h)}(y) \right\|_2^2 - 1 \right| \le h \cdot \frac{\varepsilon^2}{60L^3}$ and $\left| \left\| \mu_{i',j'}^{(h)}(z) \right\|_2^2 - 1 \right| \le h \cdot \frac{\varepsilon^2}{60L^3}$. Therefore, using the fact that $p' = \left\lceil 9L^2 /\varepsilon^{2} \right\rceil$ and by invoking \cref{lema:sensitivity-polynomial}, it follows that $\left| \dot{P}_{\tt relu}^{(p')}\left(\|\mu^{(h)}_{i,j}(y)\|_2^2\right) - \dot{P}_{\tt relu}^{(p')}(1) \right| \le  \frac{h \cdot\varepsilon}{20L^2}$ and $\left| \dot{P}_{\tt relu}^{(p')}\left(\|\mu^{(h)}_{i',j'}(z)\|_2^2\right) - \dot{P}_{\tt relu}^{(p')}(1) \right| \le \frac{ h \cdot \varepsilon}{20L^2}$. Consequently, because $\dot{P}_{\tt relu}^{(p')}(1) \le \dot{P}_{\tt relu}^{(+\infty)}(1) = 1$, we find that
	\[ \widehat{B} \le \frac{11}{10\cdot q^2}.\]
	By plugging this into \cref{cntk:phi-dot-bound2} we get the following, with probability at least $1 - \bigo\left( \frac{\delta}{L} \right)$,
	\begin{equation} \label{cntk:phi-dot-bound3}
		\left| \left< \dot{\phi}_{i,j}^{(h)}(y), \dot{\phi}_{i',j'}^{(h)}(z)\right> - \frac{1}{q^2} \cdot \dot{P}^{(p')}_{\tt relu}\left( \left<\mu_{i,j}^{(h)}(y), \mu_{i',j'}^{(h)}(z)\right> \right) \right| \le \bigo\left(\frac{\varepsilon}{q^2\cdot L}\right).
	\end{equation}
	Furthermore, recall the notation $\gamma = \frac{\sum_{a=-\frac{q-1}{2}}^{\frac{q-1}{2}} \sum_{b=-\frac{q-1}{2}}^{\frac{q-1}{2}}  \Gamma_{i+a,j+b,i'+a,j'+b}^{(h-1)}\left( y , z \right)}{\sqrt{N^{(h)}_{i,j}(y) \cdot N^{(h)}_{i',j'}(z)}}$ and note that by \cref{properties-gamma} and \cref{eq:dp-cntk-norm-simplified}, $-1 \le \gamma \le 1$. Hence, we can invoke \cref{lema:sensitivity-polynomial} and use the fact that $p' = \lceil 9 L^2 / \varepsilon^2 \rceil$ to find that \cref{eq:mu-innerprod-bound} implies the following,
	\[ \left|\dot{P}^{(p')}_{\tt relu}\left( \left<\mu_{i,j}^{(h)}(y), \mu_{i',j'}^{(h)}(z)\right> \right) - \dot{P}^{(p')}_{\tt relu}\left(\gamma \right) \right| \le \frac{h \cdot \varepsilon}{20L^2}. \]
	By incorporating the above inequality into \cref{cntk:phi-dot-bound3} using triangle inequality, we find that, with probability at least $1 - \bigo\left( \frac{\delta}{ L} \right)$, the following holds simultaneously for all $i,i' \in [d_1]$ and all $j,j'\in [d_2]$:
	\begin{equation} \label{cntk:phi-dot-bound4}
		\left| \left< \dot{\phi}_{i,j}^{(h)}(y), \dot{\phi}_{i',j'}^{(h)}(z)\right> - \frac{1}{q^2} \cdot \dot{P}^{(p')}_{\tt relu}\left(\gamma \right) \right| \le \bigo\left(\frac{\varepsilon}{q^2 L^2}\right) + \frac{h}{q^2} \cdot \frac{\varepsilon}{20L^2}.
	\end{equation}
	Since $-1 \le \gamma \le 1$, we can invoke \cref{lem:polynomi-approx-krelu} and use the fact that $p' = \left\lceil 9L^2 / {\varepsilon}^2 \right\rceil$ to conclude,
	\[ \left| \dot{P}_{\tt relu}^{(p')}\left(\gamma\right) - \kappa_0\left(\gamma\right) \right| \le \frac{\varepsilon}{15 L}. \]
	By combining above inequality with \cref{cntk:phi-dot-bound4} via triangle inequality and using the fact that, by \cref{eq:dp-cntk-derivative-covar-simplified}, $\frac{1}{q^2} \cdot \kappa_0(\gamma) \equiv \dot{\Gamma}_{i,j,i',j'}^{(h)}(y,z)$ we get the following bound simultaneously for all $i,i' \in [d_1]$ and all $j,j'\in [d_2]$, with probability at least $1 - \bigo\left( \frac{\delta}{ L} \right)$:
	\begin{equation}\label{cntk:phi-dot-bound-final}
		\left| \left< \dot{\phi}_{i,j}^{(h)}(y), \dot{\phi}_{i',j'}^{(h)}(z)\right> - \dot{\Gamma}_{i,j,i',j'}^{(h)}(y,z) \right| \le \frac{1}{q^2} \cdot \frac{\varepsilon}{8L}.
	\end{equation}
	Similarly we can prove that with probability at least $1 - \bigo\left( \frac{\delta}{L} \right)$, the following hold simultaneously for all $i,i' \in [d_1]$ and all $j,j'\in [d_2]$,
	\begin{equation}\label{cntk:phi-norm-bound-final}
		\left| \left\| \dot{\phi}_{i,j}^{(h)}(y)\right\|_2^2 - \dot{\Gamma}_{i,j,i,j}^{(h)}(y,y) \right| \le \frac{1}{q^2} \cdot \frac{\varepsilon}{8L}, \text{ and } \left| \left\| \dot{\phi}_{i',j'}^{(h)}(z)\right\|_2^2 - \dot{\Gamma}_{i',j',i',j'}^{(h)}(z,z) \right| \le \frac{1}{q^2} \cdot \frac{\varepsilon}{8L}.
	\end{equation}
	We will use \cref{cntk:phi-dot-bound-final} and \cref{cntk:phi-norm-bound-final} to prove the inductive step for $P_2(h)$.

	Next, we consider two cases for the value of $h$. When $h<L$, the vectors $\psi_{i,j}^{(h)}(y) , \psi_{i',j'}^{(h)}(z)$ are defined in \cref{psi-cntk} and when $h=L$, these vectors are defined differently in \cref{psi-cntk-last}. First we consider the case of $h<L$. Note that in this case, if we let $\eta_{i,j}^{(h)}(y)$ and $\eta_{i',j'}^{(h)}(z)$ be the vectors defined in \cref{psi-cntk}, then by \cref{lem:srht} and union bound, the following holds simultaneously for all $i,i'\in [d_1]$ and all $j,j' \in [d_2]$, with probability at least $1 - \bigo\left(\frac{\delta}{L}\right)$:
	\begin{equation}\label{eq:psi-bound1}
		\left| \left< \psi_{i,j}^{(h)}(y) , \psi_{i',j'}^{(h)}(z) \right> - \sum_{a=-\frac{q-1}{2}}^{\frac{q-1}{2}} \sum_{b=-\frac{q-1}{2}}^{\frac{q-1}{2}}  \left< \eta_{i+a,j+b}^{(h)}(y), \eta_{i'+a,j'+b}^{(h)}(z) \right> \right| \le \bigo\left( {\varepsilon}/{L} \right) \cdot D, 
	\end{equation}
	where $D := \sqrt{\sum_{a=-\frac{q-1}{2}}^{\frac{q-1}{2}} \sum_{b=-\frac{q-1}{2}}^{\frac{q-1}{2}} \|\eta_{i+a,j+b}^{(h)}(y)\|_2^2} \cdot \sqrt{\sum_{a=-\frac{q-1}{2}}^{\frac{q-1}{2}} \sum_{b=-\frac{q-1}{2}}^{\frac{q-1}{2}} \|\eta_{i'+a,j'+b}^{(h)}(z)\|_2^2}$. 
	Now, if we let $f_{i,j} := \psi^{(h-1)}_{i,j}(y) \otimes \dot{\phi}_{i,j}^{(h)}(y)$ and $g_{i',j'} := \psi^{(h-1)}_{i',j'}(z) \otimes \dot{\phi}_{i',j'}^{(h)}(z)$, then by \cref{psi-cntk}, $\eta_{i,j}^{(h)}(y) = \left(\Q^2\cdot f_{i,j}\right) \oplus \phi_{i,j}^{(h)}(y)$ and $\eta_{i',j'}^{(h)}(z) = \left(\Q^2\cdot g_{i',j'}\right) \oplus \phi_{i',j'}^{(h)}(z)$. Thus by \cref{soda-result} and union bound, with probability at least $1 - \bigo\left( \frac{\delta}{L} \right)$, we have the following inequalities simultaneously for all $i,i' \in [d_1]$ and $j,j' \in [d_2]$:
	\begin{align}
		& \left|\left<\eta^{(h)}_{i,j}(y), \eta^{(h)}_{i',j'}(z)\right> - \langle f_{i,j},g_{i',j'} \rangle - \left<\phi_{i,j}^{(h)}(y), \phi_{i',j'}^{(h)}(z)\right> \right| \le \bigo\left( \frac{\varepsilon}{L} \right) \cdot \left\| f_{i,j} \right\|_2 \left\| g_{i',j'} \right\|_2  \nonumber\\
		& \left\| \eta^{(h)}_{i,j}(y)\right\|_2^2 \le \frac{11}{10} \cdot  \|f_{i,j}\|_2^2 + \left\| \phi_{i,j}^{(h)}(y) \right\|_2^2 \label{eta-innerprod-bound}\\
		& \left\| \eta^{(h)}_{i',j'}(z)\right\|_2^2 \le \frac{11}{10} \cdot \|g_{i',j'}\|_2^2 + \left\| \phi_{i',j'}^{(h)}(z) \right\|_2^2 \nonumber
	\end{align}
	Therefore, if we condition on inductive hypotheses $P_1(h)$ and $P_2(h-1)$, then by using \cref{N-to-Sigma}, \cref{prop-pi}, the inequality \cref{cntk:phi-norm-bound-final} and \cref{properties-gamma-dot} along with the fact that $\|f_{i,j}\|_2^2 = \|\psi^{(h-1)}_{i,j}(y)\|_2^2 \cdot \|\dot{\phi}_{i,j}^{(h)}(y)\|_2^2$, we have:
	\small
	\begin{align*}
		&\sum_{a=-\frac{q-1}{2}}^{\frac{q-1}{2}} \sum_{b=-\frac{q-1}{2}}^{\frac{q-1}{2}} \|\eta_{i+a,j+b}^{(h)}(y)\|_2^2\\
		&\qquad\le \sum_{a=-\frac{q-1}{2}}^{\frac{q-1}{2}} \sum_{b=-\frac{q-1}{2}}^{\frac{q-1}{2}} \frac{11}{10} \|f_{i+a,j+b}\|_2^2 + \Gamma_{i+a,j+b,i+a,j+b}^{(h)}( y , y) + \frac{N_{i+a,j+b}^{(h)}(y)}{10q^2}\\
		&\qquad= \frac{11}{10} \cdot \sum_{a=-\frac{q-1}{2}}^{\frac{q-1}{2}} \sum_{b=-\frac{q-1}{2}}^{\frac{q-1}{2}} \|\psi^{(h-1)}_{i+a,j+b}(y)\|_2^2 \cdot \|\dot{\phi}_{i+a,j+b}^{(h)}(y)\|_2^2 + \Gamma_{i+a,j+b,i+a,j+b}^{(h)}( y , y)\\
		&\qquad\le \frac{12}{10} \sum_{a=-\frac{q-1}{2}}^{\frac{q-1}{2}} \sum_{b=-\frac{q-1}{2}}^{\frac{q-1}{2}} \Pi^{(h-1)}_{i+a,j+b, i+a, j+b}(y,y) \cdot \dot{\Gamma}_{i+a,j+b,i+a,j+b}^{(h)}(y,y) + \Gamma_{i+a,j+b,i+a,j+b}^{(h)}( y , y)\\
		&\qquad = \frac{12}{10} \cdot \Pi^{(h)}_{i,j, i, j}(y,y) = \frac{12}{10} \cdot h \cdot N^{(h+1)}_{i,j}(y),
	\end{align*}
	\normalsize
	where the fourth line above follows from the inductive hypothesis $P_2(h-1)$ along with \cref{cntk:phi-norm-bound-final} and \cref{properties-gamma-dot} and \cref{prop-pi}. The last line above follows from \cref{eq:dp-cntk} and \cref{prop-pi}.
	Similarly we can prove, $\sum_{a=-\frac{q-1}{2}}^{\frac{q-1}{2}} \sum_{b=-\frac{q-1}{2}}^{\frac{q-1}{2}} \|\eta_{i'+a,j'+b}^{(h)}(z)\|_2^2 \le \frac{12}{10} \cdot h \cdot N^{(h+1)}_{i',j'}(z)$, thus conditioned on $P_2(h-1), P_1(h), P_1(h-1)$, with probability at least $1 - \bigo\left(\frac{\delta}{L}\right)$:
	\[ D \le \frac{12}{10} \cdot h \cdot \sqrt{ N^{(h+1)}_{i,j}(y) \cdot N^{(h+1)}_{i',j'}(z)}. \]
	By incorporating this into \cref{eq:psi-bound1} it follows that if we condition on $P_2(h-1), P_1(h), P_1(h-1)$, then, with probability at least $1 - \bigo\left(\frac{\delta}{L}\right)$, the following holds simultaneously for all $i,i' \in [d_1]$ and all $j,j' \in [d_2]$,
	\begin{equation}\label{eq:psi-bound2}
		\begin{split}
			&\left| \left< \psi_{i,j}^{(h)}(y) , \psi_{i',j'}^{(h)}(z) \right> - \sum_{a=-\frac{q-1}{2}}^{\frac{q-1}{2}} \sum_{b=-\frac{q-1}{2}}^{\frac{q-1}{2}}  \left< \eta_{i+a,j+b}^{(h)}(y), \eta_{i'+a,j'+b}^{(h)}(z) \right> \right|\\ 
			&\qquad\qquad\le \bigo\left( \varepsilon h/{L} \right) \cdot \sqrt{ N^{(h+1)}_{i,j}(y) \cdot N^{(h+1)}_{i',j'}(z)}.
		\end{split}
	\end{equation}
	
	Now we bound the term $\left|\left<\eta^{(h)}_{i,j}(y), \eta^{(h)}_{i',j'}(z)\right> - \langle f_{i,j},g_{i',j'} \rangle - \left<\phi_{i,j}^{(h)}(y), \phi_{i',j'}^{(h)}(z)\right> \right|$ using \cref{eta-innerprod-bound}, \cref{cntk:phi-norm-bound-final}, and \cref{properties-gamma-dot} along with inductive hypotheses $P_2(h-1)$ and \cref{prop-pi}. With probability at least $1 - \bigo\left(\frac{\delta}{L}\right)$ the following holds simultaneously for all $i,i' \in [d_1]$ and all $j,j' \in [d_2]$:
	\small
	\[
	\begin{split}
		&\left|\left<\eta^{(h)}_{i,j}(y), \eta^{(h)}_{i',j'}(z)\right> - \langle f_{i,j},g_{i',j'} \rangle - \left<\phi_{i,j}^{(h)}(y), \phi_{i',j'}^{(h)}(z)\right> \right| \\
		&\qquad \le \bigo\left( \frac{\varepsilon}{L} \right) \cdot \sqrt{\Pi_{i,j,i,j}^{(h-1)}(y,y) \cdot \dot{\Gamma}_{i,j,i,j}^{(h)}(y,y)\cdot \Pi_{i',j',i',j'}^{(h-1)}(z,z) \cdot \dot{\Gamma}_{i',j',i',j'}^{(h)}(z,z)}\\
		&\qquad = \bigo\left( \frac{\varepsilon \cdot h}{L} \right) \cdot \frac{ \sqrt{N_{i,j}^{(h)}(y) \cdot N_{i',j'}^{(h)}(z)}}{q^2},
	\end{split}
	\]
	\normalsize
	where the last line above follows from \cref{prop-pi} together with the fact that $\dot{\Gamma}_{i,j,i,j}^{(h)}(y,y) = \dot{\Gamma}_{i',j',i',j'}^{(h)}(z,z) = \frac{1}{q^2}$.
	
	By combining the above with inductive hypotheses $P_1(h), P_2(h-1)$ and \cref{cntk:phi-dot-bound-final} via triangle inequality and invoking \cref{prop-pi} we get that the following holds simultaneously for all $i,i' \in [d_1]$ and all $j,j' \in [d_2]$, with probability at least $1 - \bigo\left(\frac{\delta}{L}\right)$,
	\small
	\begin{align}\label{eta-innerprod-bound2}
		&\left|\left<\eta^{(h)}_{i,j}(y), \eta^{(h)}_{i',j'}(z)\right> - \Pi_{i,j,i',j'}^{(h-1)}(y,z)\cdot \dot{\Gamma}_{i,j,i',j'}^{(h)}(y,z) - \Gamma_{i,j,i',j'}^{(h)}(y,z) \right| \nonumber\\
		& \le \frac{\varepsilon}{10} \cdot \frac{(h-1)^2}{L+1} \cdot \sqrt{N_{i,j}^{(h)}(y) \cdot N_{i',j'}^{(h)}(z)} \cdot \left(\left| \dot{\Gamma}_{i,j,i',j'}^{(h)}(y,z) \right|+ \frac{1}{q^2} \cdot \frac{\varepsilon}{8L} \right) + \frac{1}{q^2} \cdot \frac{\varepsilon}{8L} \cdot \left| \Pi_{i,j,i',j'}^{(h-1)}(y,z)\right| \nonumber\\ 
		& + \frac{(h+1) \cdot \varepsilon^2}{60L^3} \cdot \frac{\sqrt{N_{i,j}^{(h)}(y) \cdot N_{i',j'}^{(h)}(z)}}{q^2} + \bigo\left( \frac{\varepsilon \cdot h}{L} \right) \cdot \frac{ \sqrt{N_{i,j}^{(h)}(y) \cdot N_{i',j'}^{(h)}(z)}}{q^2}\nonumber\\
		& \le \frac{\varepsilon}{10} \cdot \frac{(h-1)^2}{L+1} \cdot \frac{\sqrt{N_{i,j}^{(h)}(y) \cdot N_{i',j'}^{(h)}(z)}}{q^2} \cdot \left(1 + \frac{\varepsilon}{8L} \right) + \frac{h-1}{q^2} \cdot \frac{\varepsilon}{8L} \cdot \sqrt{N_{i,j}^{(h)}(y) \cdot N_{i',j'}^{(h)}(z)}\nonumber\\
		&+ \left( \frac{(h+1) \cdot \varepsilon^2}{60L^3} + \bigo\left( \frac{\varepsilon \cdot h}{L} \right) \right) \cdot \frac{\sqrt{N_{i,j}^{(h)}(y) \cdot N_{i',j'}^{(h)}(z)}}{q^2} \nonumber\\
		&\le \frac{\varepsilon}{10} \cdot \frac{h^2-h/2}{L+1} \cdot \frac{\sqrt{N_{i,j}^{(h)}(y) \cdot N_{i',j'}^{(h)}(z)}}{q^2}.\nonumber
	\end{align}
	\normalsize
	By plugging the above bound into \cref{eq:psi-bound2} using triangle inequality and using \cref{eq:dp-cntk} we get the following, with probability at least $1 - \bigo\left(\frac{\delta}{L}\right)$:
	\begin{equation}\label{eq:psi-bound3}
		\begin{split}
			&\left| \left< \psi_{i,j}^{(h)}(y) , \psi_{i',j'}^{(h)}(z) \right> - \Pi_{i,j,i',j'}^{(h)}(y,z) \right|\\ 
			&\le \bigo\left( \varepsilon h/{L} \right) \cdot \sqrt{ N^{(h+1)}_{i,j}(y) \cdot N^{(h+1)}_{i',j'}(z)} \\ 
			&\qquad+ \frac{\varepsilon}{10} \cdot \frac{h^2-h/2}{L+1} \cdot \sum_{a=-\frac{q-1}{2}}^{\frac{q-1}{2}} \sum_{b=-\frac{q-1}{2}}^{\frac{q-1}{2}}  \frac{\sqrt{N_{i+a,j+b}^{(h)}(y) \cdot N_{i'+a,j'+b}^{(h)}(z)}}{q^2}\\
			&\le \bigo\left( \varepsilon h/{L} \right) \cdot \sqrt{ N^{(h+1)}_{i,j}(y) \cdot N^{(h+1)}_{i',j'}(z)} \\ 
			&\qquad+ \frac{\varepsilon}{10} \cdot \frac{h^2-h/2}{L+1} \cdot \sqrt{\sum_{a=-\frac{q-1}{2}}^{\frac{q-1}{2}} \sum_{b=-\frac{q-1}{2}}^{\frac{q-1}{2}} \frac{N_{i+a,j+b}^{(h)}(y)}{q^2}} \cdot \sqrt{ \sum_{a=-\frac{q-1}{2}}^{\frac{q-1}{2}} \sum_{b=-\frac{q-1}{2}}^{\frac{q-1}{2}} \frac{N_{i'+a,j'+b}^{(h)}(z)}{q^2}}\\
			&\le \frac{\varepsilon}{10} \cdot \frac{h^2}{L+1} \cdot\sqrt{ N^{(h+1)}_{i,j}(y) \cdot N^{(h+1)}_{i',j'}(z)}.
		\end{split}
	\end{equation}
	Similarly, we can prove that with probability at least $1 - \bigo\left( \frac{\delta}{L} \right)$ the following hold simultaneously for all $i,i' \in [d_1]$ and all $j,j' \in [d_2]$,
	\[ \begin{split}
		&\left| \left\| \psi_{i,j}^{(h)}(y)\right\|_2^2 - \Pi_{i,j,i,j}^{(h)}(y,y) \right| \le \frac{\varepsilon}{10} \cdot \frac{h^2}{L+1} \cdot N^{(h+1)}_{i,j}(y),\\ 
		&\left| \left\| \psi_{i',j'}^{(h)}(z)\right\|_2^2 - \Pi_{i',j',i',j'}^{(h)}(z,z) \right| \le \frac{\varepsilon}{10} \cdot \frac{h^2}{L+1} \cdot N^{(h+1)}_{i',j'}(z).
	\end{split} \]
	This is sufficient to prove the inductive step for statement $P_2(h)$, in the case of $h<L$, i.e., $\Pr[P_2(h)|P_2(h-1), P_1(h), P_1(h-1)] \ge 1 - \bigo(\delta/L)$.
	
	Now we prove the inductive step for $P_2(h)$ in the case of $h=L$. Similar to before, if we let $f_{i,j} := \psi^{(L-1)}_{i,j}(y) \otimes \dot{\phi}_{i,j}^{(L)}(y)$ and $g_{i',j'} := \psi^{(L-1)}_{i',j'}(z) \otimes \dot{\phi}_{i',j'}^{(L)}(z)$, then by \cref{psi-cntk-last}, we have $\psi_{i,j}^{(L)}(y) = \Q^2\cdot f_{i,j}$ and $\psi_{i',j'}^{(L)}(z) = \Q^2\cdot g_{i',j'}$. Thus by \cref{soda-result} and union bound, we find that, with probability at least $1 - \bigo\left( \frac{\delta}{ L} \right)$, the following inequality holds simultaneously for all $i,i' \in [d_1]$ and $j,j' \in [d_2]$:
	\[ \left|\left<\psi^{(L)}_{i,j}(y), \psi^{(L)}_{i',j'}(z)\right> - \langle f_{i,j},g_{i',j'} \rangle \right| \le \bigo\left( \frac{\varepsilon}{L} \right) \cdot \left\| f_{i,j} \right\|_2 \left\| g_{i',j'} \right\|_2.\]
	Therefore, using \cref{cntk:phi-norm-bound-final} and \cref{properties-gamma-dot} along with inductive hypotheses $P_2(L-1)$ and \cref{prop-pi}, with probability at least $1 - \bigo\left(\frac{\delta}{L}\right)$, the following holds simultaneously for all $i,i' \in [d_1]$ and $j,j' \in [d_2]$,
	\small
	\[
	\begin{split}
		\left|\left<\psi^{(L)}_{i,j}(y), \psi^{(L)}_{i',j'}(z)\right> - \langle f_{i,j},g_{i',j'} \rangle \right| & \le \bigo\left( \frac{\varepsilon}{L} \right)  \sqrt{\Pi_{i,j,i,j}^{(L-1)}(y,y) \cdot \dot{\Gamma}_{i,j,i,j}^{(L)}(y,y)\cdot \Pi_{i',j',i',j'}^{(L-1)}(z,z) \cdot \dot{\Gamma}_{i',j',i',j'}^{(L)}(z,z)}\\
		& = \bigo\left( {\varepsilon} \right) \cdot \frac{\sqrt{N_{i,j}^{(L)}(y) \cdot N_{i',j'}^{(L)}(z)}}{q^2}.
	\end{split}
	\]
	\normalsize
	By combining the above with inductive hypotheses $P_1(L), P_2(L-1)$ and \cref{cntk:phi-dot-bound-final} via triangle inequality and invoking \cref{prop-pi} and also using the definition of $\Pi^{(L)}(y,z)$ given in \cref{eq:dp-cntk-last-layer}, we get that the following holds, simultaneously for all $i,i' \in [d_1]$ and $j,j' \in [d_2]$, with probability at least $1 - \bigo\left(\frac{\delta}{L}\right)$,
	\small
	\begin{align}
		&\left|\left<\psi^{(L)}_{i,j}(y), \psi^{(L)}_{i',j'}(z)\right> - \Pi_{i,j,i',j'}^{(L)}(y,z) \right| \nonumber\\
		& \le \frac{\varepsilon}{10} \cdot \frac{(L-1)^2}{L+1} \cdot \sqrt{N_{i,j}^{(L)}(y) \cdot N_{i',j'}^{(L)}(z)} \cdot \left(\left| \dot{\Gamma}_{i,j,i',j'}^{(L)}(y,z) \right|+ \frac{1}{q^2} \cdot \frac{\varepsilon}{8L} \right) + \frac{1}{q^2} \cdot \frac{\varepsilon}{8L} \cdot \left| \Pi_{i,j,i',j'}^{(L-1)}(y,z)\right| \nonumber\\ 
		& + \frac{(L+1) \cdot \varepsilon^2}{60L^3} \cdot \frac{\sqrt{N_{i,j}^{(L)}(y) \cdot N_{i',j'}^{(L)}(z)}}{q^2} + \bigo\left( {\varepsilon} \right) \cdot \frac{ \sqrt{N_{i,j}^{(L)}(y) \cdot N_{i',j'}^{(L)}(z)}}{q^2}\nonumber\\
		& \le \frac{\varepsilon}{10} \cdot \frac{(L-1)^2}{L+1} \cdot \frac{\sqrt{N_{i,j}^{(L)}(y) \cdot N_{i',j'}^{(L)}(z)}}{q^2} \cdot \left(1 + \frac{\varepsilon}{8L} \right) +  \frac{\varepsilon}{8q^2} \cdot \sqrt{N_{i,j}^{(L)}(y) \cdot N_{i',j'}^{(L)}(z)}\nonumber\\
		&+ \left( \frac{(L+1) \cdot \varepsilon^2}{60L^3} + \bigo\left( {\varepsilon} \right) \right) \cdot \frac{\sqrt{N_{i,j}^{(L)}(y) \cdot N_{i',j'}^{(L)}(z)}}{q^2} \nonumber\\
		&\le \frac{\varepsilon \cdot (L-1)}{10} \cdot \frac{\sqrt{N_{i,j}^{(L)}(y) \cdot N_{i',j'}^{(L)}(z)}}{q^2}.\nonumber
	\end{align}
	\normalsize
	This proves the inductive step for statement $P_2(h)$, in the case of $h=L$, i.e., $\Pr[P_2(L)|P_2(L-1), P_1(L), P_1(L-1)] \ge 1 - \bigo(\delta/L)$.
	The induction is complete and hence the statements of lemma are proved by union bounding over all $h = 0,1,2, \ldots L$.
    This completes the proof of \cref{lem:cntk-sketch-corr}.
\qed

In the following lemma we analyze the runtime of the CNTK Sketch algorithm,
\begin{lemma}[Runtime of the CNTK Sketch]
	\label{thm:cntk-sketch-runtime}
	For every positive integers $d_1,d_2,c$, and $L$, every $\varepsilon, \delta>0$, every image $x \in \RR^{d_1\times d_2\times c}$, the time to compute the CNTK Sketch $\Psi_{\tt cntk}^{(L)}(x) \in \RR^{s^*}$, for $s^*=\bigo\left( \frac{1}{\varepsilon^2} \cdot \log \frac{1}{\delta} \right)$, using the procedure given in \cref{alg-def-cntk-sketch} is bounded by $\bigo\left( \frac{L^{11}}{\varepsilon^{6.7}} \cdot (d_1d_2) \cdot \log^3 \frac{d_1d_2L}{\varepsilon\delta}\right)$.
\end{lemma}
{\it Proof of \cref{thm:cntk-sketch-runtime}:}
	First note that the total time to compute $N_{i,j}^{(h)}(x)$ for all $i \in [d_1]$ and $j \in [d_2]$ and $h=0,1, \ldots L$ as per \cref{eq:dp-cntk-norm-simplified} is bounded by $\bigo\left(q^2 L \cdot d_1d_2\right)$.
	Besides the time to compute $N_{i,j}^{(h)}(x)$, there are two other main components to the runtime of this procedure. The first heavy operation corresponds to computing vectors $\left[Z_{i,j}^{(h)}(x)\right]_l = \Q^{2p+2} \cdot \left(\left[ \mu_{i,j}^{(h)}(x) \right]^{\otimes l} \otimes {e}_1^{\otimes 2p+2-l}\right)$ for $l=0,1,2, \ldots 2p+2$ and $h=1,2, \ldots L$ and all indices $i \in[d_1]$ and $j \in [d_2]$, in \cref{eq:maping-cntk-covar}. By \cref{soda-result}, the time to compute $\left[Z_{i,j}^{(h)}(x)\right]_l$ for a fixed $h$, fixed $i \in[d_1]$ and $j \in [d_2]$, and all $l=0,1,2, \ldots 2p+2$ is bounded by,
	\[ \bigo\left( \frac{L^{10}}{\varepsilon^{20/3}} \cdot \log^2\frac{L}{\varepsilon} \cdot \log^3 \frac{d_1d_2L}{\varepsilon\delta} + q^2 \cdot \frac{L^8}{\varepsilon^{16/3}} \cdot \log^3 \frac{d_1d_2L}{\varepsilon\delta} \right) = \bigo\left( \frac{ L^{10}}{\varepsilon^{6.7}} \cdot \log^3 \frac{d_1d_2L}{\varepsilon\delta} \right). \]
	The total time to compute vectors $\left[Z_{i,j}^{(h)}(x)\right]_l$ for all $h=1,2, \ldots L$ and all $l=0,1,2, \ldots 2p+2$ and all indices $i \in [d_1]$ and $j \in [d_2]$ is thus bounded by $\bigo\left( \frac{L^{11}}{\varepsilon^{6.7}} \cdot (d_1d_2) \cdot \log^3 \frac{d_1d_2L}{\varepsilon\delta} \right)$. 
	The next computationally expensive operation is computing vectors $\left[Y_{i,j}^{(h)}(x)\right]_l$ for $l=0,1,2, \ldots 2p'+1$ and $h=1,2, \ldots L$, and all indices $i \in [d_1]$ and $j \in [d_2]$, in \cref{eq:cntk-map-phidot}.
	By \cref{soda-result}, the runtime of computing $\left[Y_{i,j}^{(h)}(x)\right]_l$ for a fixed $h$, fixed $i \in [d_1]$ and $j \in [d_2]$, and all $l=0,1,2, \ldots 2p'+1$ is bounded by,
	\[ \bigo\left( \frac{L^{6}}{\varepsilon^6} \cdot \log^2\frac{L}{\varepsilon} \log^3 \frac{d_1d_2L}{\varepsilon\delta} + \frac{q^2 \cdot L^{8}}{\varepsilon^6} \cdot \log^3 \frac{d_1d_2L}{\varepsilon\delta} \right) = \bigo\left( \frac{ L^{8}}{\varepsilon^6} \log^2\frac{L}{\varepsilon} \cdot \log^3 \frac{d_1d_2L}{\varepsilon\delta} \right). \]
	Hence, the total time to compute vectors $\left[Y_{i,j}^{(h)}(x)\right]_l$ for all $h=1,2, \ldots L$ and $l=0,1,2, \ldots 2p'+1$ and all indices $i \in [d_1]$ and $j \in [d_2]$ is $\bigo\left( \frac{ L^{9}}{\varepsilon^6} \log^2\frac{L}{\varepsilon} \cdot (d_1d_2) \cdot \log^3 \frac{d_1d_2L}{\varepsilon\delta} \right)$. The total runtime bound is obtained by summing up these three contributions.
	This completes the proof of \cref{thm:cntk-sketch-runtime}.
\qed

Now we are ready to prove \cref{maintheorem-cntk}. 
\maintheoremcntk*
{\it Proof of \cref{maintheorem-cntk}:}
	Let $\psi^{(L)}:\RR^{d_1\times d_2\times c} \to \RR^{d_1 \times d_2 \times s}$ for $s=\bigo\left( \frac{L^4}{\varepsilon^2} \cdot \log^3 \frac{d_1d_2L}{\varepsilon\delta} \right)$ be the mapping defined in \cref{psi-cntk-last} of \cref{alg-def-cntk-sketch}. By \cref{Psi-cntk-def}, the CNTK Sketch $\Psi_{\tt cntk}^{(L)}(x)$ is defined as 
	\[\Psi_{ntk}^{(L)}(x):=\frac{1}{d_1d_2} \cdot \G \cdot \left(\sum_{i \in [d_1]} \sum_{j \in [d_2]} \psi^{(L)}_{i,j}(x)\right).\]
	The matrix $\G$ is defined in \cref{Psi-cntk-def} to be a matrix of i.i.d. normal entries with $s^{*} = C \cdot \frac{1}{\varepsilon^2} \cdot \log\frac{1}{\delta}$ rows for large enough constant $C$. \cite{dasgupta2003elementary} shows that $\G$ is a JL transform and hence $\Psi_{\tt cntk}^{(L)}$ satisfies the following,
	\[ \Pr \left[ \left| \left< \Psi_{\tt cntk}^{(L)}(y) , \Psi_{\tt cntk}^{(L)}(z) \right> - \frac{1}{d_1^2 d_2^2} \cdot \sum_{i,i'\in[d_1]}\sum_{j ,j'\in [d_2]} \left< \psi_{i,j}^{(L)}( y),  \psi_{i',j'}^{(L)}( z) \right> \right| \le \bigo(\varepsilon) \cdot A \right] \ge 1 - \bigo(\delta), \]
	where $A := \frac{1}{d_1^2 d_2^2} \cdot \left\| \sum_{i\in[d_1]}\sum_{j \in [d_2]}  \psi_{i,j}^{(L)}( y)\right\|_2 \cdot \left\|\sum_{i\in[d_1]}\sum_{j \in [d_2]} \psi_{i,j}^{(L)}(z)\right\|_2$. By triangle inequality together with \cref{lem:cntk-sketch-corr} and \cref{prop-pi}, the following bounds hold with probability at least $1 - \bigo(\delta)$:
	\[ \begin{split}
		&\left\| \sum_{i\in[d_1]}\sum_{j \in [d_2]}  \psi_{i,j}^{(L)}( y)\right\|_2 \le \frac{11}{10} \cdot \frac{\sqrt{L-1}}{q} \cdot \sum_{i\in[d_1]}\sum_{j \in [d_2]} \sqrt{N^{(L)}_{i,j}(y)},\\
		&\left\| \sum_{i\in[d_1]}\sum_{j \in [d_2]} \psi_{i,j}^{(L)}( z)\right\|_2 \le \frac{11}{10} \cdot \frac{\sqrt{L-1}}{q} \cdot \sum_{i\in[d_1]}\sum_{j \in [d_2]} \sqrt{N^{(L)}_{i,j}(z)},
	\end{split} \]
	Therefore, by union bound we find that, with probability at least $1 - \bigo(\delta)$:
	\[\begin{split}
		&\left| \left< \Psi_{\tt cntk}^{(L)}(y) , \Psi_{\tt cntk}^{(L)}(z) \right> - \frac1{d_1^2 d_2^2} \cdot {\sum_{i,i' \in [d_1]}\sum_{j,j' \in [d_2]} \left< \psi_{i,j}^{(L)}( y),  \psi_{i',j'}^{(L)}( z) \right>} \right|\\ 
		&\qquad \le \bigo\left(\frac{\varepsilon L}{q^2 \cdot d_1^2 d_2^2}\right) \cdot \sum_{i, i' \in[d_1]}\sum_{j, j' \in [d_2]} \sqrt{N^{(L)}_{i,j}(y)\cdot  N^{(L)}_{i',j'}(z)}.
	\end{split} \]
	Be combining the above with \cref{lem:cntk-sketch-corr} using triangle inequality and union bound and also using \cref{eq:dp-cntk-finalkernel}, the following holds with probability at least $1 - \bigo(\delta)$:
	\small
	\begin{equation}\label{Psi-error-bound1}
		\left| \left< \Psi_{\tt cntk}^{(L)}(y) , \Psi_{\tt cntk}^{(L)}(z) \right> - \Theta_{\tt cntk}^{(L)}(y,z) \right| \le \frac{\varepsilon \cdot (L-1)}{9 q^2 \cdot d_1^2 d_2^2} \cdot \sum_{i, i' \in[d_1]}\sum_{j, j' \in [d_2]} \sqrt{N^{(L)}_{i,j}(y)\cdot  N^{(L)}_{i',j'}(z)}. 
	\end{equation}
	\normalsize
	
	Now we prove that $\Theta_{\tt cntk}^{(L)}(y,z) \ge \frac{L-1}{9q^2 d_1^2 d_2^2} \cdot \sum_{i,i'\in[d_1]}\sum_{j,j' \in [d_2]} \sqrt{N^{(L)}_{i,j}(y) \cdot N^{(L)}_{i',j'}(z)}$ for every $L \ge 2$.
	First note that, it follows from \cref{eq:dp-cntk-covar-simplified} that $\Gamma_{i,j,i',j'}^{(1)}(y,z) \ge 0$ for any $i,i',j,j'$ because the function $\kappa_1$ is non-negative everywhere on $[-1,1]$. This also implies that $\Gamma_{i,j,i',j'}^{(2)}(y,z) \ge \frac{\sqrt{N^{(2)}_{i,j}(y) \cdot N^{(2)}_{i',j'}(z)}}{\pi \cdot q^2}$ because $\kappa_1(\alpha) \ge \frac{1}{\pi}$ for every $\alpha \in [0,1]$. Since, $\kappa_1(\cdot)$ is a monotone increasing function, by recursively using \cref{eq:dp-cntk-norm-simplified} and \cref{eq:dp-cntk-covar-simplified} along with \cref{properties-gamma}, we can show that for every $h \ge 1$, the value of $\Gamma_{i,j,i',j'}^{(h)}(y,z)$ is lower bounded by $\frac{\sqrt{N^{(h)}_{i,j}(y) \cdot N^{(h)}_{i',j'}(z)}}{q^2} \cdot \Sigma_{\tt relu}^{(h)}(-1)$, where $\Sigma_{\tt relu}^{(h)}:[-1,1] \to \RR$ is the function defined in \cref{eq:dp-covar-relu}.

	Furthermore, it follows from \cref{eq:dp-cntk-derivative-covar-simplified} that $\dot{\Gamma}_{i,j,i',j'}^{(1)}(y,z) \ge 0$ for any $i,i',j,j'$ because the function $\kappa_0$ is non-negative everywhere on $[-1,1]$. Additionally, $\dot{\Gamma}_{i,j,i',j'}^{(2)}(y,z) \ge \frac{1}{2 q^2}$ because $\kappa_0(\alpha) \ge \frac{1}{2}$ for every $\alpha \in [0,1]$. By using the inequality $\Gamma_{i,j,i',j'}^{(h)}(y,z) \ge \frac{\sqrt{N^{(h)}_{i,j}(y) \cdot N^{(h)}_{i',j'}(z)}}{q^2} \cdot \Sigma_{\tt relu}^{(h)}(-1)$ that we proved above along with the fact that $\kappa_0(\cdot)$ is a monotone increasing function and recursively using  \cref{eq:dp-cntk-derivative-covar-simplified} and \cref{properties-gamma}, it follows that for every $h \ge 1$, we have $\dot{\Gamma}_{i,j,i',j'}^{(h)}(y,z) \ge \frac{1}{q^2} \cdot \dot{\Sigma}_{\tt relu}^{(h)}(-1)$. 
	
	By using these inequalities and Definition of $\Pi^{(h)}$ in \cref{eq:dp-cntk} together with \cref{eq:dp-cntk-norm-simplified}, recursively, it follows that, for every $i,i',j,j'$ and $h = 2,\ldots L-1$:
	\begin{align*}
		\Pi_{i,j,i',j'}^{(h)}(y,z) \ge \frac{h}{4} \cdot \sqrt{N_{i,j}^{(h+1)}(y) \cdot N_{i',j'}^{(h+1)}(z)},
	\end{align*}
	Therefore, using this inequality and \cref{eq:dp-cntk-last-layer} we have that for every $L\ge2$:
	\begin{align*}
		\Pi_{i,j,i',j'}^{(L)}(y,z) &\ge \frac{L-1}{4} \cdot \sqrt{N_{i,j}^{(L)}(y) \cdot N_{i',j'}^{(L)}(z)} \cdot \frac{\dot{\Sigma}_{\tt relu}^{(L)}(-1)}{q^2}\\
		&\ge \frac{L-1}{9q^2} \cdot \sqrt{N_{i,j}^{(L)}(y) \cdot N_{i',j'}^{(L)}(z)}.
	\end{align*}
	Now using this inequality and \cref{eq:dp-cntk-finalkernel}, the following holds for every $L\ge2$:
	\[ \Theta_{\tt cntk}^{(L)}(y,z) \ge \frac{L-1}{9q^2 d_1^2 d_2^2} \cdot \sum_{i,i'\in[d_1]}\sum_{j,j' \in [d_2]} \sqrt{N^{(L)}_{i,j}(y) \cdot N^{(L)}_{i',j'}(z)}. \]
	Therefore, by incorporating the above into \cref{Psi-error-bound1} we get that,
	\[ \Pr \left[ \left| \left< \Psi_{\tt cntk}^{(L)}(y) , \Psi_{\tt cntk}^{(L)}(z) \right> - \Theta_{\tt cntk}^{(L)}(y,z) \right| \le \varepsilon \cdot \Theta_{\tt cntk}^{(L)}(y,z) \right] \ge 1 - \delta. \]
	
	{\bf Runtime analysis:} By \cref{thm:cntk-sketch-runtime}, time to compute the CNTK Sketch is $\bigo\left( \frac{ L^{11}}{\varepsilon^{6.7}} \cdot (d_1d_2) \cdot \log^3 \frac{d_1d_2L}{\varepsilon\delta} \right)$.

	This completes the proof of \cref{maintheorem-cntk}.
\qed


\end{document}